\documentclass[11pt]{article}

\usepackage[final]{acl}

\usepackage{times}
\usepackage{latexsym}

\usepackage[T1]{fontenc}

\usepackage[utf8]{inputenc}

\usepackage{microtype}

\usepackage{inconsolata}

\usepackage{graphicx}

\usepackage{enumitem}
\usepackage{algorithm}
\usepackage{algorithmic}
\usepackage{tcolorbox}
\usepackage{subcaption} 
\usepackage{amsmath}
\usepackage{amssymb}
\usepackage{booktabs} 
\usepackage{multirow}
\usepackage{tabularx} 
\newcolumntype{Y}{>{\centering\arraybackslash}X} 
\usepackage{makecell}
\usepackage{xcolor}
\usepackage{pifont} 
\newcommand{\cmark}{\textcolor{green!60!black}{\ding{51}}} 
\newcommand{\xmark}{\textcolor{red!70!black}{\ding{55}}}   
\tcbuselibrary{listings, skins, breakable} 

\title{Beyond Itinerary Planning—A Real-World Benchmark for \\ Multi-Turn and Tool-Using Travel Tasks}
\author{
Xiang Cheng\textsuperscript{1,4},
Yulan Hu\textsuperscript{4}\thanks{Corresponding authors.},
Xiangwen Zhang\textsuperscript{4},\\
{\bf Lu Xu\textsuperscript{2,4},
Lide Tan\textsuperscript{3,4},
Zheng Pan\textsuperscript{4},
Xin Li\textsuperscript{4},
Yong Liu\textsuperscript{1}}\footnotemark[1] \\
\textsuperscript{1}Gaoling School of Artificial Intelligence, Renmin University of China\\
\textsuperscript{2}National University of Singapore \quad
\textsuperscript{3}Beihang University\\
\textsuperscript{4}AMAP, Alibaba Group\\
\texttt{\{chengxiang1,liuyonggsai\}@ruc.edu.cn} \quad \texttt{zy2406221@buaa.edu.cn}\\
\texttt{\{huyulan,zhangxiangwen.zxw,yuxi.xl,panzheng.pan,beilai.bl\}@alibaba-inc.com}
}

\begin{document}
\maketitle
\begin{abstract}
Travel planning is a natural real-world task to test large language models' (LLMs) planning and tool-use abilities. Although prior work has studied LLM performance on travel planning, existing settings still differ from real-world needs, mainly due to limited domain coverage, insufficient modeling of users' implicit preferences in multi-turn conversations, and a lack of evaluation of agents' capability boundaries. To mitigate these gaps, we propose \mbox{\textbf{TravelBench}}, a benchmark for \textit{truly real-world} travel planning. We collect user queries, user preferences, and tools from real scenarios, and construct three subtasks---\textit{Single-Turn}, \textit{Multi-Turn}, and \textit{Unsolvable}---to evaluate agents' three core capabilities in real settings: (1) solving problems independently, (2) interacting with users to elicit implicit preferences, and (3) recognizing the capability boundaries. To enable stable tool invocation and reproducible evaluation, we cache real tool-call results and build a sandbox environment which integrates ten travel-related tools, enabling agents to combine these tools to solve most practical travel planning problems. We evaluate multiple LLMs on TravelBench and find that even advanced models exhibit imbalanced performance across different capabilities. Our further systematic verification demonstrates the stability of the proposed benchmark. TravelBench provides a practical and reproducible benchmark to advance research on LLM agents for real-world travel planning\footnote{Code and data are publicly available at \url{https://github.com/small-xiangcheng/TravelBench}}.

\end{abstract}

\section{Introduction}
Recently, LLM-based agents have made significant progress in planning and tool use \cite{team2025kimi_k2,zeng2025glm_45,qian2025toolrl,liu2025vaiage}, enabling them to autonomously invoke external tools during reasoning to solve problems. Travel planning is a complex real-world task that naturally involves multiple subtasks, including point-of-interest (POI\footnote{POI refers to real-world places such as restaurants, tourist attractions, hospitals, subway stations, etc.}) exploration, weather-aware decisions, route planning, and itinerary design for both short and long trips. These subtasks require an agent to coordinate constraints, decompose goals, and iteratively refine plans with tool support \cite{xie2024human-like,zheng2024natural}. Therefore, travel planning serves as a suitable testbed for evaluating an agent’s multi-step reasoning, tool use, and ability in multi-turn interactions with users.

To evaluate model capability in this domain, \citet{xie2024travelplanner} introduced the first travel-planning benchmark. However, because the tasks and constraints were relatively simple, it was soon solved by solver-based methods \cite{solve-travelplanner}. ChinaTravel \cite{shao2024chinatravel} uses real user queries and imposes stricter constraints with a scalable evaluation protocol. TripScore \cite{qu2025tripscore} further proposes a finer-grained evaluation scheme and provides a single reward signal as the scoring criterion. Other work extends these benchmarks to multi-turn user interaction \cite{qin2025compass,oh2025flex-travelplanner,deng2025retail}, enlarges the dataset scale \cite{wang-etal-2025-triptailor}, or leverages external data sources during planning \cite{TP-RAG}.  

Despite these advances, several key limitations remain. (1) User preferences and constraints are typically pre-defined, either injected directly into the instruction or revealed step by step by a user simulator, and thus do not support \textit{dynamically eliciting users’ implicit preferences\footnote{\emph{Implicit preferences} refer to preferences that are known to the user but are not directly observable to the assistant unless they are actively elicited during interaction.}} through multi-turn interactions. (2) Many benchmarks cover only short-trip or long-trip planning, \textit{overlooking other diverse travel-planning tasks} in real-world settings. (3) They either do not support \textit{tool use}, or rely on \textit{synthetic queries and preferences}, which cannot faithfully reflect real-world data and preference distributions. As a result, \textit{existing benchmarks still fall short of fully assessing an agent’s ability to handle realistic travel scenarios.}

\begin{table*}
\centering
\caption{Comparison with prior travel-planning benchmarks along several key dimensions. \textit{Implicit pref.} denotes implicit user preferences that are not explicitly stated but are inherent in the user profile, and \textit{Broad dom.} indicates coverage of broader data domains. \cmark{} and \xmark{} indicate the presence and absence of each capability, respectively.}
\small
\setlength{\tabcolsep}{4.5pt}
\renewcommand{\arraystretch}{1.15}
\begin{tabular}{lcccccc}
\toprule
\textbf{Work} &
\textbf{Sandbox} &
\textbf{Real-queries} &
\textbf{Multi-turn} &
\textbf{Unsolved} &
\textbf{Implicit pref.} &
\textbf{Broad dom.} \\
\midrule
TravelPlanner \cite{xie2024travelplanner}
& \cmark & \xmark & \xmark & \xmark & \xmark & \xmark \\
Flex-TravelPlanner\cite{oh2025flex-travelplanner}
& \xmark & \xmark & \cmark & \xmark & \xmark & \xmark \\
ChinaTravel \cite{shao2024chinatravel}
& \cmark & \cmark & \xmark & \xmark & \xmark & \xmark \\
TripScore \cite{qu2025tripscore}
& \xmark & \cmark & \xmark & \xmark & \xmark & \xmark \\
TP-RAG \cite{TP-RAG}
& \xmark & \cmark & \xmark & \xmark & \xmark & \xmark \\
TripTailor\cite{wang-etal-2025-triptailor}
& \cmark & \xmark & \xmark & \xmark & \xmark & \xmark \\
COMPASS\cite{qin2025compass}
& \cmark & \xmark & \cmark & \xmark & \xmark & \xmark \\
\textbf{TravelBench (Ours)}
& \cmark & \cmark & \cmark & \cmark & \cmark & \cmark \\
\bottomrule
\end{tabular}
\label{tab:comparison_dimensions}
\vspace{-10pt}
\end{table*}

To fill this gap, we propose \textbf{TravelBench}, a new travel-planning benchmark for comprehensive evaluation of agent capabilities in realistic travel tasks. We collect a large set of real user queries and user preferences covering diverse travel needs (e.g., POI exploration, route planning, solution comparison, and itinerary design). To evaluate three core abilities: \textit{whether an agent can handle a request on its own, ask the user when key information is missing, and admit when it cannot do the task}, we include three subsets in our benchmark: 500 single-turn queries, 500 multi-turn queries, and 100 unsolvable queries. We further curate a toolkit of 10 real-world travel-planning tools (e.g., POI search, flight search, train search, route planning, and weather forecast; see Table~\ref{tab:tool_library}). Based on these 1,100 instances, we build a tool-call cache with approximately 200,000 real tool responses to provide stable and consistent tool outputs during evaluation. All agent reasoning is executed in an isolated sandbox environment, ensuring fully reproducible evaluation. For scoring, we adopt an LLM-as-a-judge protocol \cite{pathak2025rubric-is-all-you-need} to assess response quality and task completion under a unified rubric.

The difference between TravelBench and prior benchmarks is summarized in Table~\ref{tab:comparison_dimensions}.
Our main contributions are:\par
\noindent(1) \textbf{TravelBench}, a truly real-world benchmark built from \textbf{real} user queries, preferences, and tools, covering a \textbf{broader} and more practical task scope.\par
\noindent(2) The first travel-planning benchmark that incorporates \textbf{implicit preferences} and allows \textbf{multi-turn user--agent interaction} to elicit them, while also explicitly including \textbf{unsolvable queries}. These settings enable a focused evaluation of the three core agents' abilities. \par
\noindent(3) A \textbf{reproducible sandbox} with cached real tool calls, integrating a broader and more comprehensive set of \textbf{travel-related tools} to enable stable tool-augmented evaluation with consistent outputs.\par
\section{Related Work}
\subsection{LLM Agents and Agentic RL in Travel}
LLM agents remain brittle in practical travel planning: agents may lose long-horizon focus, choose inappropriate tools, or fail under multiple interacting constraints~\cite{xie2024travelplanner,deng2025retail}. To improve robustness, prior work enhances LLM-based travel planners with retrieval (e.g., trajectory/POI databases)~\cite{TP-RAG}, optimization modules (e.g., numerical solvers for enforcing constraints)~\cite{shao-etal-personal_travel_solver,solve-travelplanner,ju-etal-2024-TTG}, and symbolic components (e.g., DSL-based feasibility checking)~\cite{shao2024chinatravel}. In parallel, recent studies explore stronger agent structures beyond a single planner, such as multi-agent collaboration~\cite{choi2025atlas,zhang-planning-with-multi-constraints,chen2024travelagent,liu2025vaiage}. Building on this trajectory, a newer line of work emphasizes \emph{tool-centric} training via agentic reinforcement learning: inspired by ReAct-style tool use~\cite{yao2022react}, methods such as DeepTravel~\cite{ning2025deeptravel} and TripScore~\cite{qu2025tripscore} train agents with reward signals to improve feasibility and consistency. These developments make tool-using travel agents more practical, but most systems are still evaluated in relatively constrained settings, often tied to a specific domain. Therefore, we need a unified, reproducible benchmark that jointly supports \textit{realistic tool use, multi-turn interaction, and explicit handling of unsolvable cases}.
\subsection{Travel Planning Benchmarks}

Several benchmarks have been proposed for natural-language based travel planning. TravelPlanner\cite{xie2024travelplanner} as the first benchmark in this line of work, centers on multi-day trip itinerary construction and provides a sandbox for tool calls. However, it was soon solved by solver-based methods. Subsequent studies extend it from several perspectives. ChinaTravel~\cite{shao2024chinatravel} introduces stricter constraints and uses DSL-based formulations that make constraints explicitly checkable. TripScore~\cite{qu2025tripscore} proposes a unified scoring metric for plan quality beyond binary feasibility. Beyond hard-constraint satisfaction, TripTailor~\cite{wang-etal-2025-triptailor} further evaluates the overall reasonableness and personalization of itineraries, while Compass~\cite{qin2025compass} focuses on optimizing soft preferences to search for better solutions. Other work explores retrieval-augmented planning with trajectory references~\cite{TP-RAG} and studies how language choices and output formats affect travel-planning performance\cite{jung-etal-2025-can-llm-Truly-plan}. There is also growing interest in robustness under dynamic changes to travel plans~\cite{deng2025retail,karmakar2025triptide}, as well as interactive clarification when user instructions are underspecified~\cite{zhang-ask-before-plan,deng2025retail,qin2025compass}.

Overall, existing benchmarks often focus primarily on trip itinerary planning as a single subtask. Moreover, many of them do not jointly include: (i) diverse real user queries, (ii) multi-turn dialogues that enable dynamic refinement and implicit preference elicitation, and (iii) a reproducible sandbox with real tool outputs for controlled evaluation. As a result, the gap to real-world usage remains, and \textit{performance on these benchmarks may not faithfully reflect how agents behave in practical travel-planning environments}.
\section{TravelBench}

\begin{figure*}[t]
  \centering
  \includegraphics[width=0.99\textwidth]{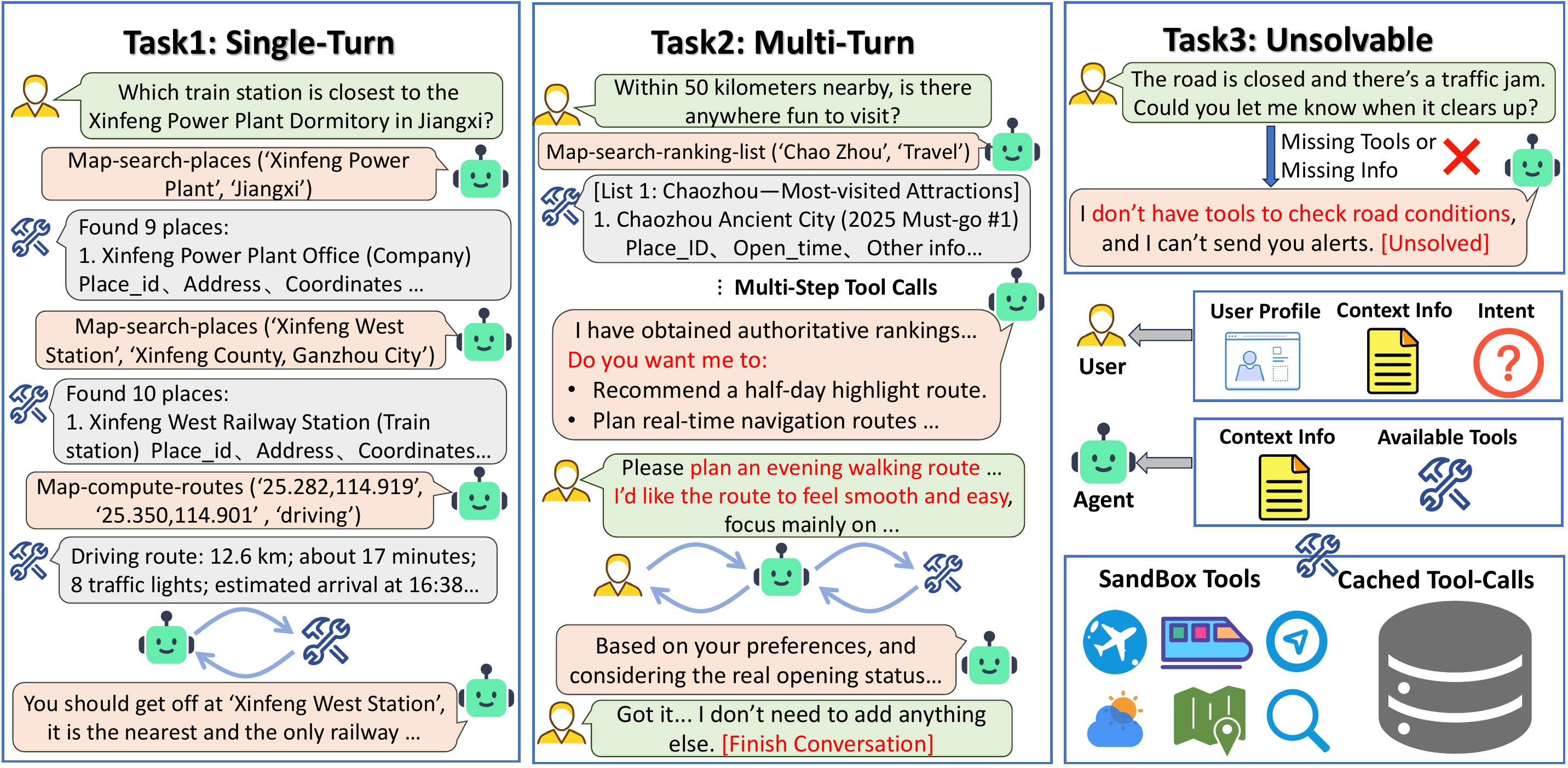}
  \caption{\textbf{Overview of TravelBench.} The user is simulated by an LLM with user profile and contextual information, while the agent is given the same context and access to external tools through an offline \textbf{sandbox} backed by a cached tool-response pool, enabling stable and accurate tool execution. We define three settings: \textbf{Single-turn}, where the agent may perform multi-step tool use without interacting with the user; \textbf{Multi-turn}, where the agent may both use tools and conduct multi-round dialogue to request missing information; and \textbf{Unsolvable}, where the agent must recognize its capability boundaries and abstain when required tools or information are unavailable.}
  \label{fig:main-tasks}
  \vspace{-10pt}
\end{figure*}

This section describes the construction of TravelBench, including query collection and filtering, subtask decomposition, data distribution statistics, dataset quality validation, and the sandbox environment. An overview of the three subtasks in TravelBench is shown in Figure~\ref{fig:main-tasks}.

\subsection{Data Collection and Process}
\label{sec:data-collection}
\textbf{Data Collection.} All initial queries and user preferences in TravelBench are collected from \textbf{real-world data}\footnote{Sourced from Alibaba Amap's online platform under applicable regulatory constraints.}.
We gather four months of logs (Aug--Nov 2025), resulting in about 5,000 curated queries (including noisy or semantically ambiguous ones) with associated context and user preferences after an initial filtering and deduplication pass. The data reflects a broad and realistic spectrum of travel needs, including POI exploration, navigation, route planning, solution comparison, and itinerary design, which is a main focus of prior work. 

\textbf{User Profile.} 
To use real user preferences while protecting privacy, we anonymize all preference information. Specifically, we use an instruction-following LLM (Qwen-Plus) to remove any potentially personally identifiable information (PII), such as company addresses, exact addresses of frequently visited POIs, and license plate numbers. At the same time, we preserve preference signals as much as possible---for example, by abstracting preferences for specific POI addresses into preferences for broader POI categories. \textit{We treat these preference descriptions as implicit constraints that guide multi-turn interactions and better reflect real-world scenarios.} We further leverage the LLM to synthesize basic user attributes, such as gender, family structure, and lifestyle, and combine them with the extracted real preference information to construct user profiles for our benchmark. As illustrated in Figure~\ref{fig:main-tasks}, these user profiles are maintained only by the user simulator and are not provided to the agent as part of the interaction context. The detailed structure of the user profiles is provided in Appendix~\ref{app:user_profile clarification}; for more details on the use of real-world data and the associated considerations, see \nameref{sec:ethics}.

\textbf{Diversity Annotation.}
To reduce the semantic redundancy in real-world user logs, we employ the K-Center-Greedy algorithm \cite{sener2018k-center-greedy} to rank queries by diversity. This approach iteratively selects samples that maximize the distance from the existing set in the embedding space, effectively capturing the long-tail distribution of user intents. We construct subtasks by sampling the top-$k$ most diverse queries, ensuring broad semantic coverage while avoiding mode collapse.

\textbf{Unsolvable Task.}
To determine whether a query is executable, we use three models, GPT-5.1\cite{openai_gpt51_2025}, Qwen3-235B-Th-2507 and Qwen-Plus~\cite{yang2025qwen3} to label each query as \emph{solvable} or \emph{unsolvable}. For unsolvable queries, we further assign one of three causes: (i) lack of tool support, (ii) missing necessary context, or (iii) no clear executable intent. We treat queries that are judged unsolvable by all three models as the \textbf{unsolvable} subtasks to evaluate the assistant's ability to recognize these queries. Meanwhile, we remove from the raw query pool any query that is marked as unsolvable by at least one model. The prompt used for executability annotation is shown in Figure~\ref{app:prompt:unsolvability determination}.

\subsection{Multi-turn vs.\ Single-turn Tasks}
For solvable queries, we further categorize them by whether the agent can complete the task in a single turn or requires multi-turn interaction.

Our key premise is that \textit{whether a query is single-turn or multi-turn should be determined by the agent's actual interaction behavior}, rather than predefined heuristics. Accordingly, we run every solvable query in our interactive framework as multi-turn subtasks and label it based on whether the assistant asks clarification questions: queries with follow-up questions are \textbf{multi-turn}, and the rest are \textbf{single-turn}. In practice, follow-up questions typically arise from missing constraints or preference-dependent choices.

Concretely, we instantiate the assistant with two models, Gemini-3-Flash-preview~\cite{google_gemini3_2025} and GPT-5.1, using a temperature of 0.7. For each model, we conduct three independent trials, resulting in six runs per query in total. We record the number of interaction turns in each run, where an interaction count of 1 means the assistant completes the task in a single turn without any follow-up questions. We then assign labels based on how many of the six runs have a turn count of 1. If at least four runs have a turn count of 1, we label the query as \textit{single-turn}, meaning the task can be completed without interaction with high probability. If none of the runs has a turn count of 1, we label the query as \textit{multi-turn}, indicating that user input is consistently required. Finally, following the diversity ranking described above, we sample 500 single-turn queries and 500 multi-turn queries to construct subtasks. The prompts used for multi-turn interaction and the user simulator are shown in Figures~\ref{app:prompt:assistant-multi-turn} and~\ref{app:prompt:user-simulator}.

\subsection{Query Types and Distribution}
\label{sec:query_distribution}

A key contribution of \textit{TravelBench} is its broader and more practical task scope compared to existing trip-planning benchmarks. As shown in Figure~\ref{fig:data_distribution}, queries in the dataset are organized according to a diverse taxonomy derived from the inherent \texttt{primary\_intent} field of the raw user data. These categories range from \textit{Discovery} and \textit{Planning} to \textit{Rules and Policies}, reflecting a real-world distribution where traditional itinerary planning is merely a subset of user needs. 

Beyond intent variety, TravelBench offers extensive coverage along multiple axes:
\begin{itemize}[leftmargin=*, noitemsep, topsep=0pt]
    \item \textbf{Geographic \& Temporal:} The dataset spans 32 province-level regions and 243 cities across China (Figure~\ref{fig:city-distribution}). The collection covers weekdays, weekends, and major holidays across all time windows (Figure \ref{fig:time-distribution}).
    \item \textbf{Demographics \& Preferences:} Queries represent diverse user groups aged 18--60+ and cover a wide range of occupations (Figure~\ref{fig:age-distribution}). User preferences are implicitly reflected in profiles covering various aspects of daily life---such as mobility, dining, and cultural interests---requiring models to infer user needs through interaction.

\end{itemize}
Detailed statistics and the full intent hierarchy are provided in Appendix~\ref{app:sec:dataset_coverage}.

\subsection{Dataset Quality and Human Validation}
\label{sec:dataset_validation}

To ensure the quality and reliability of TravelBench, we conducted systematic human verification across key stages of dataset construction:
\begin{itemize}[leftmargin=*, noitemsep, topsep=0pt]
    \item \textbf{De-identification:} All 500 multi-turn user preferences underwent \textbf{100\% manual verification} to remove sensitive information.
    \item \textbf{Label Consistency:} Human evaluation on 100 sampled queries per task showed high agreement with our annotations: \textbf{97.0\%} for unsolvable cases, \textbf{97.0\%} for the single-turn split, and \textbf{95.0\%} for the multi-turn split.

\end{itemize}
Human validation results demonstrate the high quality of our dataset and the soundness of our processing pipeline; see Appendix~\ref{appendix:dataset_quality} for details.

\subsection{Sandbox Environment}
To cover the core needs in travel scenarios, we integrate 10 real tools into TravelBench (Table~\ref{tab:tool_library}), such as flight/train search, map-based POI search, and route planning. These tools can be composed as needed to solve most real travel inquiries, e.g., “find the highest-rated hotel near a location and plan a route from the station to the hotel.”

To ensure stable and reproducible evaluation, following prior work~\cite{qin2023toolllm,guo2024stabletoolbench}, we build a sandbox environment. During evaluation, we do not query external tool endpoints directly; instead, we return tool outputs from a pre-built cache whenever available. To construct this real-response cache, we set the temperature to 0.7 and run multiple models (e.g., GPT-4.1, GPT-5.1, Qwen-Plus, Qwen3-4B, and Qwen3-30B) multiple times with access to the real tool APIs. We cache each tool call from the model traces together with its real response, yielding approximately 200{,}000 tool-call traces with real outputs. The distribution of cached tool calls is shown in Figure~\ref{fig:cached_tool_distribution}.

Although the cache covers most tool calls, exact-match misses are still unavoidable due to minor variations in tool arguments (e.g., “Beijing City” vs.\ “Beijing”). To keep the evaluation deterministic and to match the real tool-response distribution under cache misses, we adopt an \textbf{embedding-based retrieval + ICL simulation} strategy~\cite{brown2020gpt3,guo2024stabletoolbench}. Specifically, we precompute embeddings of cached tool inputs using Qwen3-Embedding-8B~\cite{yang2025qwen3}. When a miss occurs, we embed the current tool arguments and use Faiss to retrieve the top-8 most similar cached calls. We then feed these retrieved examples to an LLM as in-context demonstrations to generate a simulated tool response consistent with the cached results, ensuring stable and reproducible outputs. In addition, in our offline execution environment we perform strict argument validation for every tool call (e.g., required-field checks, type and range constraints, and completeness). Invalid or inconsistent calls are recorded as tool-call errors. We also report the tool-call error rate as an indicator of model reliability in tool use. More details are provided in Appendix~\ref{app:sec:sandbox-tools}.

\section{Evaluation Protocol}
This section presents the evaluation protocol for TravelBench. For the three subsets, we use rule-based scoring for the unsolvable subset, and an LLM-as-a-judge protocol with additional penalty terms for the single-turn and multi-turn subsets. The detailed prompts are provided in Appendix~\ref{app:prompt}.

\textbf{Unsolvable subset.} For the unsolvable subset, we use the same reasoning framework in the single-turn setting, but explicitly instruct the agent to output the special tag \texttt{[Unsolved]} immediately once it determines that the request cannot be completed due to missing information or missing tool support. Therefore, evaluation reduces to checking whether the assistant's first response contains \texttt{[Unsolved]}. Formally, for an instance $j$, we define
\begin{equation}
y_j =
\begin{cases}
1, & \text{if \texttt{[Unsolved]} in first response}, \\
0, & \text{otherwise}.
\end{cases}
\end{equation}
The unsolvable score is then reported as accuracy over the subset:
\begin{equation}
S_{\text{unsolved}} = \frac{1}{N_{\text{unsolved}}}\sum_{j=1}^{N_{\text{unsolved}}} y_j.
\end{equation}

\textbf{Single-turn \& Multi-turn.}
For the solvable single-turn and multi-turn subsets, rule-based rewards used in prior work are unreliable, as our benchmark covers broader domains and includes interactive dialogues. We therefore adopt an LLM-as-a-judge framework with (i) rubric-based scoring, (ii) instance-level meta-judge for score calibration, and (iii) instance-level tool-call error penalties.

For each instance $j$ in subset $t\in\{\text{single},\text{multi}\}$, we score the trajectory on a 1--5 scale across $d_t$ rubric dimensions. For single-turn, $d_{\text{single}}=3$, including \textit{reasoning\_planning}, \textit{summarization\_extraction}, and \textit{presentation}. For multi-turn, $d_{\text{multi}}=4$, with an additional \textit{user\_interaction} dimension. Specifically, \textbf{\textit{reasoning\_planning}} evaluates whether the model correctly understands the user goal, plans tool usage appropriately, and avoids redundant or incorrect tool calls; \textbf{\textit{summarization\_extraction}} evaluates whether the model accurately extracts and integrates relevant information from tool outputs and dialogue context without hallucination; \textbf{\textit{presentation}} evaluates whether the final response is clear, accurate, concise, and helpful to the user; and \textbf{\textit{user\_interaction}} evaluates whether the model asks necessary and high-value follow-up questions when information is insufficient or tool execution fails, while minimizing unnecessary user burden. Detailed rubric definitions are provided in Appendix~\ref{sec:appendix_rubric_details}.

Let $r_{t,j,i}\in\{1,2,3,4,5\}$ denote the score on dimension $i$. We compute the normalized score of instance $j$ as
\vspace{-5pt}
\begin{equation}
S_{t,j} = \frac{\left(\frac{1}{d_t}\sum_{i=1}^{d_t} r_{t,j,i}\right)-1}{4}.
\end{equation}

\textbf{Meta-evaluation calibration.}
We further introduce a meta-judge to verify whether the judge score is reasonable. Its purpose is to calibrate a small number of likely over-estimated raw scores (e.g., due to missed hallucinations or clear rubric violations), rather than to broadly downweight all instances. In practice, such calibration is only triggered for a small fraction of cases (about $3\%$). The meta-judge outputs a calibration score $s_{t,j}\in\{1,2,3,4,5\}$, which we convert into a multiplicative factor:
\begin{equation}
w^{\text{meta}}_{t,j} = \frac{s_{t,j}}{5}\in(0,1].
\end{equation}
We interpret the calibrated score $S^{\text{cal}}_{t,j} = S_{t,j}\cdot w^{\text{meta}}_{t,j}$ as a corrected estimate of trajectory quality.

\textbf{Tool-use penalty.}
Tool-call errors can degrade efficiency and user experience. For each instance $j$, we compute a tool-call error rate from its execution trace. A tool call is considered erroneous if it violates the tool specification (e.g., invalid tool name, missing required arguments, type mismatch, or other schema violations). Let $N^{\text{err}}_{t,j}$ be the number of erroneous tool calls and $N^{\text{all}}_{t,j}$ be the total number of tool calls for this instance. We define the tool-use penalty coefficient as
\begin{equation}
w^{\text{tool}}_{t,j} = 1 - \frac{N^{\text{err}}_{t,j}}{N^{\text{all}}_{t,j}}, \quad w^{\text{tool}}_{t,j}\in[0,1].
\end{equation}

We apply the same penalty scheme to both single-turn and multi-turn subsets and compute all rubric scores and penalties \emph{per instance}. Here, $w^{\text{meta}}_{t,j}$ calibrates occasional overestimation in the raw judge score, while $w^{\text{tool}}_{t,j}$ penalizes tool-use errors observed in the execution trace. The penalized instance score is computed by applying the tool-use penalty to the calibrated score:
\begin{equation}
S^{\text{pen}}_{t,j} = S^{\text{cal}}_{t,j}\cdot w^{\text{tool}}_{t,j}.
\end{equation}

We report the subset score by averaging over instances, i.e., 
$S^{\text{pen}}_{t} = \sum_{j\in D_t} S^{\text{pen}}_{t,j} / |D_t|$, 
where $D_t$ denotes the set of instances in subset $t$. 
The overall benchmark score is then 
$S_{\text{avg}} = (S_{\text{single}}^{\text{pen}} + S_{\text{multi}}^{\text{pen}} + S_{\text{unsolved}})/3$.

\begin{table*}[t]
\small
\centering
\setlength{\tabcolsep}{5pt}
\renewcommand{\arraystretch}{0.9}
\caption{Main results on \textit{TravelBench} across models and the three subtasks. \textit{Raw} denotes the unpenalized score, \textit{Error} denotes the tool-call error rate, and \textit{Pen} denotes the penalized score. \textit{Th} denotes thinking models, and \textit{It} denotes instruction-following models. }
\label{tab:main-results}
\begin{tabularx}{0.9\textwidth}{l *{6}{Y} >{\centering\arraybackslash}m{1.4cm} >{\centering\arraybackslash}m{1.0cm}}
\toprule
\multirow{2}{*}{\textbf{Model}} &
\multicolumn{3}{c}{\textbf{Multi-turn}} &
\multicolumn{3}{c}{\textbf{Single-turn}} &
\multicolumn{1}{c}{\textbf{Unsolvable}} &
\multicolumn{1}{c}{\textbf{Overall}} \\
\cmidrule(lr){2-4}\cmidrule(lr){5-7}\cmidrule(lr){8-8}\cmidrule(lr){9-9}
& \textbf{Raw} & \textbf{Error} & \textbf{Pen.}
& \textbf{Raw} & \textbf{Error} & \textbf{Pen.}
& \textbf{Acc.} & \textbf{Score} \\
\midrule
\multicolumn{9}{l}{\textbf{Frontier Models}} \\
\midrule
GPT-5.1 & 75.17 & 3.97 & 71.31 & 75.98 & 1.76 & 73.81 & 80.00 & \underline{75.04} \\
GPT-4.1 & 70.20 & 4.46 & 65.58 & 73.58 & 1.97 & 70.87 & 35.00 & 57.15 \\
Kimi-K2-0925 & 54.23 & 5.26 & 49.99 & 65.89 & 4.28 & 61.62 & \textbf{94.00} & 68.54 \\
Kimi-K2-Th & 75.48 & 3.93 & \underline{71.83} & 78.40 & 1.09 & 77.31 & 73.67 & 74.27 \\
MiniMax-M2 & \underline{78.44} & 15.37 & 66.79 & 81.46 & 17.99 & 68.12 & 52.67 & 62.53 \\
DeepSeek-V3.2 & \textbf{86.36} & 1.55 & \textbf{82.80} & \textbf{87.01} & 0.57 & \textbf{83.29} & 51.33 & 72.47 \\
DeepSeek-R1 & 38.07 & 4.80 & 35.67 & 79.33 & 2.03 & 76.93 & \underline{83.67} & 65.42 \\
Qwen-Plus & 66.21 & 4.75 & 62.56 & \underline{84.86} & 2.16 & \underline{82.64} & \underline{83.67} & \textbf{76.29} \\
\midrule
\multicolumn{9}{l}{\textbf{Lightweight Models}} \\
\midrule
Qwen3-235B-Th & 59.80 & 2.54 & 57.52 & \textbf{74.55} & 0.81 & \textbf{73.51} & 51.67 & 60.90 \\
Qwen3-235B-It & \textbf{66.12} & 5.69 & \textbf{61.78} & \underline{73.82} & 3.53 & \underline{70.74} & \textbf{80.00} & \textbf{70.84} \\
Qwen3-14B & 51.35 & 4.80 & 48.41 & 59.38 & 1.93 & 57.60 & 54.00 & 53.34 \\
Qwen3-30B-It & 51.98 & 5.51 & 48.16 & 51.92 & 1.37 & 50.47 & 67.33 & 55.32 \\
Qwen3-30B-Th & \underline{62.37} & 2.02 & \underline{60.30} & 71.27 & 0.83 & 69.85 & 56.33 & \underline{62.16} \\
Qwen3-4B-It & 46.29 & 4.83 & 43.43 & 43.53 & 1.16 & 42.53 & \underline{73.00} & 52.99 \\
Qwen3-4B-Th & 58.44 & 1.80 & 56.60 & 69.55 & 1.19 & 68.32 & 58.67 & 61.20 \\
\bottomrule
\end{tabularx}
\vspace{-10pt}
\end{table*}

\section{Experiments}

In this section, we first evaluate \textit{TravelBench} on a wide range of advanced LLMs and provide detailed analyses. We then conduct additional experiments to demonstrate the stability and validity of each module in our benchmark. More results are provided in Appendix~\ref{sec:additional_experiments}.

\subsection{Models}
To comprehensively evaluate \textit{TravelBench}, we test a broad set of LLMs, covering proprietary and open-source models, instruction-following models, and reasoning-oriented models. This includes the GPT family~\cite{openai_gpt51_2025}, the DeepSeek family~\cite{liu2025deepseek,guo2025deepseekr1}, the Qwen family ranging from 4B to 235B parameters~\cite{yang2025qwen3}, as well as the latest models from MiniMax~\cite{minimax_2025} and Kimi~\cite{team2025kimi_k2}. This model suite allows us to benchmark and compare the agentic capabilities of frontier models under the same travel-planning setting. Notably, we do not include ReAct-style prompting methods~\cite{yao2022react}, because most frontier models, especially reasoning models, already internalize similar reasoning and tool-use patterns; our focus is on evaluating their agentic performance.

\subsection{Evaluation Details}
For multi-turn tasks, we use GPT-4.1 as the user simulator to generate user replies and as the tool simulator for cache misses in the sandbox. To reduce randomness and improve reproducibility, we set the simulator temperature to 0. For each evaluated agent, we set the sampling temperature to 0.7 and run three trials per instance, reporting the average score across trials. We use Gemini-3-Flash-preview as the judge model for both rubric-based scoring and meta-judging, with temperature set to 0 to further reduce evaluation variance. We cap each assistant response at 8{,}192 tokens. For open-source models, we deploy them with vLLM and set the maximum sequence length to 128k tokens to support long-context multi-turn interactions and tool-call traces. We also deploy Qwen3-Embedding-8B with vLLM for cache-miss retrieval in the sandbox, using its default embedding dimension of 4{,}096.

\subsection{Main Results}
\textbf{Reflection of Real-World Performance.}
Table~\ref{tab:main-results} reports the performance of different models on \textit{TravelBench}. Overall, the strongest proprietary models achieve around 75 points, with Qwen-Plus obtaining the best overall score of 76.29. This suggests that frontier models can already use tools to handle many real travel-planning requests, but still struggle with more complex cases. In contrast, lightweight models cluster around 50--60 points, indicating substantial room for improvement. We argue that \textit{TravelBench} not only reflects real-world performance but also places strong requirements on practical travel planning.

\textbf{Imbalanced Capabilities Across Models.}
We observe that many models show imbalanced strengths across subtasks. For example, Kimi-K2-0925 achieves a very high score on the unsolvable subset (94), but lags behind on both single-turn and multi-turn tasks. In contrast, DeepSeek-V3.2 performs best on single-turn and multi-turn tasks (around 83), but performs poorly on the unsolvable subset (51.33). We also find that thinking models tend to score lower than instruction-following models on the unsolvable subset, which may indicate that instruction-following models are better at recognizing unsolvable requests. Overall, these results suggest that (i) accurately identifying unsolvability and (ii) successfully completing complex interactive planning are two distinct capabilities. Future work may need to study how to better balance them to improve practical usability.

\textbf{Effectiveness of the Tool-Use Penalty.}
Table~\ref{tab:main-results} also reports tool-call error rates. For example, for MiniMax-M2, the rubric judge assigns relatively high scores to the interaction traces (the model may partially recover after making tool-call mistakes), but the final score drops substantially after applying the tool-use penalty (multi-turn: 78.44 $\rightarrow$ 66.79), which better reflects its practical usability. We further observe that multi-turn tasks generally incur higher tool-call error rates than single-turn tasks. In addition, reasoning-oriented models typically exhibit lower error rates than instruction-following models, suggesting that multi-turn interactions introduce additional difficulty and that test-time scaling\footnote{Here, test-time scaling means that, compared with instruction-following models, reasoning models involve additional reasoning during inference.} can improve tool-call accuracy. We believe the tool-use penalty better aligns benchmark scores with real-world performance and encourages more careful tool use during planning.

\begin{table}
  \centering
  \caption{Results across three evals (std over three runs).}
  \label{tab:3-evals}
  \small
  \setlength{\tabcolsep}{6pt}
  \renewcommand{\arraystretch}{1.15}
  \begin{tabular}{lcccc}
    \toprule
    \textbf{Model} & \textbf{E1} & \textbf{E2} & \textbf{E3} & \textbf{Std} \\
    \midrule
    \textbf{GPT-5.1}       & 75.04 & 75.04 & 75.05 & \textbf{0.01} \\
    \textbf{Qwen-Plus}    & 76.29 & 76.32 & 76.32 & \textbf{0.02} \\
    \textbf{Qwen3-30B-Th} & 62.16 & 62.15 & 62.18 & \textbf{0.02} \\
    \textbf{Qwen3-30B-It} & 55.32 & 55.34 & 55.32 & \textbf{0.01} \\
    \bottomrule
  \end{tabular}
  \vspace{-10pt}
\end{table}

\subsection{Stability and Human Validation of Benchmark Components}
\label{sec:stability-of-benchmark-components}

\begin{table*}[t]
  \small
  \centering
  \caption{Offline/Online scores and cache statistics, where Hit (Tools) denotes the hit rate (total number of tool calls).}
  \label{tab:offline-online-cache-wide-paren}
  \setlength{\tabcolsep}{5pt}
  \renewcommand{\arraystretch}{1.1}
  \begin{tabular}{lcccccc}
    \toprule
    \multirow{2}{*}{\textbf{Model}} &
    \multicolumn{2}{c}{\textbf{Score}} &
    \multicolumn{2}{c}{\textbf{Multi-turn -Hit (Tools)}} &
    \multicolumn{2}{c}{\textbf{Single-turn -Hit (Tools)}} \\
    \cmidrule(lr){2-3}\cmidrule(lr){4-5}\cmidrule(lr){6-7}
    & \textbf{Offline} & \textbf{Online} & \textbf{Offline} & \textbf{Online} & \textbf{Offline} & \textbf{Online} \\
    \midrule
    \textbf{GPT-5.1} &
    75.04 & 74.93 &
    23.23 (4503) & 25.82 (4543) &
    41.01 (3819) & 42.89 (3842) \\
    \textbf{Qwen-Plus} &
    76.29 & 76.53 &
    28.98 (8771) & 28.51 (9962) &
    58.88 (5720) & 62.10 (5675) \\
    \textbf{Qwen3-30B-Th} &
    62.16 & 61.88 &
    46.72 (4675) & 47.28 (4873) &
    61.45 (3883) & 62.57 (3911) \\
    \textbf{Qwen3-30B-It} &
    55.32 & 55.01 &
    23.44 (7547) & 26.54 (7982) &
    55.66 (4294) & 57.32 (4494) \\
    \bottomrule
  \end{tabular}
  \vspace{-10pt}
\end{table*}

In this section, we empirically demonstrate that our benchmark provides a reliable evaluation of model performance in real-world settings. We focus on two key components: the LLM-as-a-judge module and the sandbox tool caching module.

\textbf{Stable and Reasonable Scoring.}
LLM-as-a-judge has been shown to be effective and is widely used for evaluating open-ended tasks~\cite{hashemi2024llm-rubric,pathak2025rubric-is-all-you-need}. To examine the stability of our judge module, we score the same trajectory multiple times and measure the variance across trials. Specifically, for trajectories produced by four models (GPT-5.1, Qwen-Plus, Qwen3-30B-Th, and Qwen3-30B-It), we run the judge three times. As shown in Table~\ref{tab:3-evals}, the scores are highly consistent across runs, with standard deviations close to 0.01. This indicates that our LLM judge is stable in practical use.

In addition to this repeated-run stability, we also conduct direct human validation of both the raw judge and the meta-judge. In the multi-turn setting, the raw judge achieves an average MAE (mean absolute error) of 0.52 against human annotators, which is close to the human-human disagreement level of 0.48, indicating good alignment with human judgment. We also verify that the meta-judge mainly acts as a lightweight calibration step: on 1,500 sampled trajectories, only 3.3\% receive any penalty, and manual inspection shows that these cases are primarily trajectories with hallucination-related issues missed by the raw judge. Full details are provided in Appendix~\ref{appendix:judge_validation}.

\begin{figure}
    \centering
    \includegraphics[width=1.0\columnwidth]{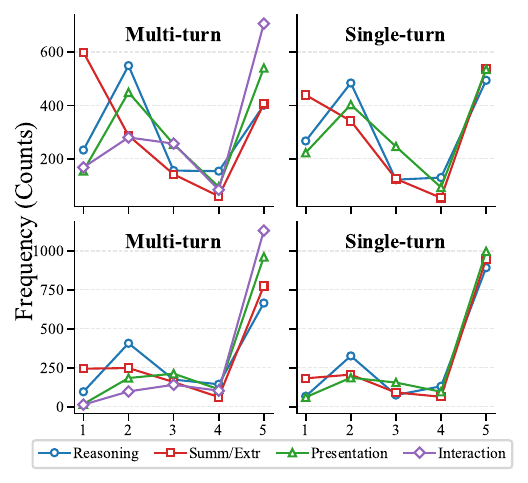}
    \caption{Distribution of judged scores across different dimensions for Qwen3-30B-It(Top) and GPT-5.1(Low). The plots compare multi-turn and single-turn subtasks, showing the frequency of scores from 1 to 5.}
    \label{fig:score-distribution}
    \vspace{-10pt}
\end{figure}

We further visualize score distributions over different trajectories in Figure~\ref{fig:score-distribution}. The upper panel (Qwen3-30B) shows that scores are well distributed across dimensions in both the single-turn and multi-turn settings. Based on a lightweight manual audit via random sampling, we observe that high-scoring trajectories typically correspond to simpler queries, while low-scoring trajectories often contain hallucinations\footnote{Such hallucinations are typically reflected in lower scores on the summarization\_extraction dimension.} or incorrect tool-call parameters, suggesting good alignment between the judge and human judgment. The lower panel shows the distribution of GPT-5.1 trajectories. Compared with Qwen3-30B, the overall scores are clearly higher, consistent with its stronger capability.

\textbf{Stability of the Sandbox Caching Module.}
To demonstrate the stability and reliability of our tool caching module, Table~\ref{tab:offline-online-cache-wide-paren} reports model scores and cache-hit rates under two settings: offline evaluation (using the sandbox environment) and online evaluation (calling real tools). The results show that the final scores are very close between offline and online setups, with only small differences (0.11\textasciitilde0.31). Moreover, for both single-turn and multi-turn tasks, the cache hit rates and the total number of tool calls are also similar across the two settings. These observations suggest that the simulated tool cache does not substantially affect subsequent reasoning or tool usage, and further confirm that our sandbox environment is stable and can provide consistent and accurate tool outputs.

\textbf{Overall Stability.}
After verifying the stability of two modules, we further evaluate the stability of the entire evaluation system. We run three independent trials for four models, following the same evaluation protocol described above. As shown in Table~\ref{tab:3-trials}, the standard deviation across runs ranges is 0.19\textasciitilde0.81, which is still acceptable. We attribute this variance mainly to mild stochasticity introduced by setting the sampling temperature to 0.7, and it can be further reduced by increasing the number of retries.
 
\subsection{Difficulty Distribution}
\label{difficulty-distribution}
\begin{figure}[t]
    \centering
    \begin{minipage}[t]{0.494\columnwidth}
        \centering
        \includegraphics[width=\linewidth]{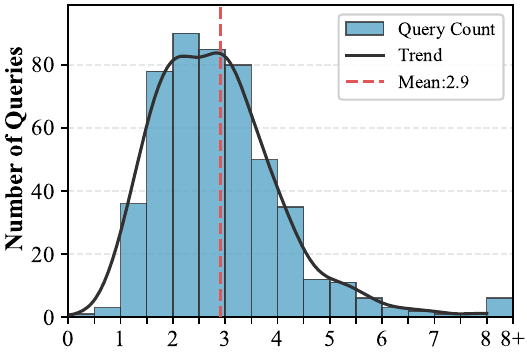}
    \end{minipage}\hfill
    \begin{minipage}[t]{0.494\columnwidth}
        \centering
        \includegraphics[width=\linewidth]{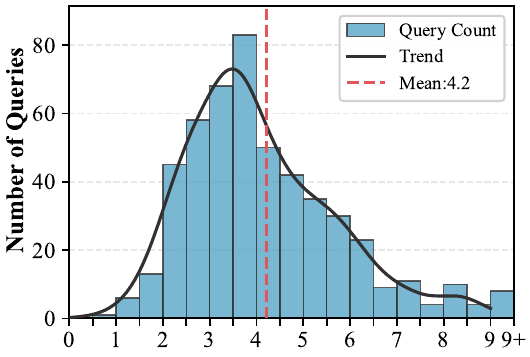}
    \end{minipage}
    \caption{Distribution plot of the average number of tool calls across four models mentioned above, with single-turn on the left and multi-turn on the right.}
    \label{fig:tool_call_distribution}
    \vspace{-10pt}
\end{figure}

To characterize the difficulty distribution of our subtasks, we use the \textit{average number of tool calls as a proxy: tasks that require more tool calls across models are likely more difficult} (e.g., more open-ended and requiring more information gathering). The distribution of tool calls therefore serves, to some extent, as a rough proxy for task difficulty. We provide additional evidence supporting this assumption in Appendix~\ref{sec:appendix_toolcall_difficulty}. As shown in Figure~\ref{fig:tool_call_distribution}, single-turn tasks generally involve fewer tool calls than multi-turn tasks, and multi-turn tasks more frequently exceed six tool calls, suggesting higher difficulty. Notably, most queries in our benchmark appear to concentrate around medium difficulty, while very easy and very hard queries form long tails. This pattern matches the difficulty distribution of real-world user requests, further supporting the realism of our benchmark. In future work, tool-call counts could also be used to select harder queries for more challenging evaluation.

\begin{table}[t]
  \centering
  \caption{Results across three trials (std over three runs).}
  \label{tab:3-trials}
  \small
  \setlength{\tabcolsep}{6pt}
  \renewcommand{\arraystretch}{1.15}
  \begin{tabular}{lcccc}
    \toprule
    \textbf{Model} & \textbf{T1} & \textbf{T2} & \textbf{T3} & \textbf{Std} \\
    \midrule
    \textbf{GPT-5.1}       & 75.04 & 74.46 & 73.76 & \textbf{0.64} \\
    \textbf{Qwen-Plus}    & 76.29 & 76.75 & 76.19 & \textbf{0.30} \\
    \textbf{Qwen3-30B-Th} & 62.16 & 63.29 & 61.71 & \textbf{0.81} \\
    \textbf{Qwen3-30B-It} & 55.32 & 55.11 & 55.49 & \textbf{0.19} \\
    \bottomrule
  \end{tabular}
  \vspace{-5pt}
\end{table}

\section{Conclusion}
In this paper, we introduce \textbf{TravelBench}, a high-fidelity benchmark for tool-augmented travel planning. It is grounded in real user queries and preferences, and covers realistic tools. TravelBench targets three core agent capabilities through three subsets: \textbf{single-turn} (completing a request when requirements are fully specified), \textbf{multi-turn} (asking clarifying questions and eliciting preferences when key information is missing), and \textbf{unsolved} (identifying and appropriately handling infeasible requests). To enable stable and reproducible evaluation, we execute all runs in an offline \textbf{sandbox} that simulates online tools using a cache of $\sim$200K real tool responses. For cache misses, we use a retrieval-augmented LLM simulator to generate responses while preserving the distribution of real tool outputs. We also adopt an \textbf{LLM-as-a-judge} rubric with explicit penalties for tool-call errors, better reflecting deployment quality. Our stability checks show that judge scores and sandbox tool outputs are consistent across repeated runs, and that offline sandbox results closely match executions with online tools. Experiments with both proprietary and open-source models show highly imbalanced performance across the three subsets, suggesting that robust real-world travel planning remains far from solved. Additional discussions on dataset scalability and practical utility are provided in Appendix~\ref{app:sec:more_discussion}.
\section*{Limitations}
Although TravelBench is designed to reflect realistic travel planning, the current sandbox includes a limited set of tools. As a result, some common subtasks (e.g., checking real-time traffic conditions or setting reminders) are marked as unsolved in our evaluation. In future work, we will incorporate additional tools into the sandbox to broaden the task coverage.

Our evaluation also relies on an LLM-as-a-judge scorer. Therefore, scores are directly comparable only when the same judge model is used. Future work may explore alternatives such as pairwise preference evaluation or trajectory-level metrics to improve robustness and comparability.

\section*{Ethical Considerations}
\label{sec:ethics}

Our benchmark involves real-world user queries and user preference data, which may be sensitive. The underlying data are collected from Alibaba Amap's online platform and include raw user queries as well as user preference information related only to domains such as hotels, attractions, dining, and travel. No other types of personal information are involved. All data collection procedures were conducted under internal supervision and in compliance with the platform’s terms of service and data usage policies.

To protect user privacy, we apply strict de-identification to preference data, such as removing company addresses, exact addresses of frequently visited POIs, and license plate numbers, followed by manual inspection. In particular, to reduce the risk of re-identification from user preferences, we use Qwen-Plus to abstract all preference information. For example, specific POI addresses are generalized into POI categories. This process preserves realistic preference patterns while preventing reverse identification of individuals.

In addition, some user profile attributes, such as gender, family structure, and lifestyle, are synthetically generated by Qwen-Plus rather than derived from real users. We include such information because we consider these attributes important components of user profiles for evaluating personalization. These synthetic profile attributes do not correspond to any real individual and are used solely to provide realistic profile context. As a result, in the final released user profile data, only the preference-related component is derived from abstractions of real-world preferences, while all other profile fields are LLM-generated. We therefore believe that the resulting user profiles can reflect realistic preference distributions while providing strong privacy protection.

The released benchmark is intended only for research on personalization and tool-use evaluation. It should not be used to infer real-user characteristics or to support high-stakes decision-making.

\section*{Acknowledgements}
This research was supported by Alibaba AMap, National Key Research and Development Program of China (NO. 2024YFE0203200), National Natural Science Foundation of China (No.62476277), CCF-ALIMAMA TECH Kangaroo Fund(No.CCF-ALIMAMA OF 2024008), and Huawei-Renmin University joint program on Information Retrieval. We also acknowledge the support provided by the fund for building worldclass universities (disciplines) of Renmin University of China and by the funds from Beijing Key Laboratory of Big Data Management and Analysis Methods, Gaoling School of Artificial Intelligence, Renmin University of China, from Engineering Research Center of Next-Generation Intelligent Search and Recommendation, Ministry of Education, from Intelligent Social Governance Interdisciplinary Platform, Major Innovation \& Planning Interdisciplinary Platform for the “DoubleFirst Class” Initiative, Renmin University of China, from Public Policy and Decision-making Research Lab of Renmin University of China, and from Public Computing Cloud, Renmin University of China.


\bibliography{custom}

@inproceedings{yao2022react,
  title={React: Synergizing reasoning and acting in language models},
  author={Yao, Shunyu and Zhao, Jeffrey and Yu, Dian and Du, Nan and Shafran, Izhak and Narasimhan, Karthik R and Cao, Yuan},
  booktitle={The eleventh international conference on learning representations},
  year={2022}
}

@inproceedings{xie2024travelplanner,
title={TravelPlanner: A Benchmark for Real-World Planning with Language Agents},
author={Jian Xie and Kai Zhang and Jiangjie Chen and Tinghui Zhu and Renze Lou and Yuandong Tian and Yanghua Xiao and Yu Su},
booktitle={Forty-first International Conference on Machine Learning},
year={2024},
url={https://openreview.net/forum?id=l5XQzNkAOe}
}

@inproceedings{solve-travelplanner,
    title = "Large Language Models Can Solve Real-World Planning Rigorously with Formal Verification Tools",
    author = "Hao, Yilun  and
      Chen, Yongchao  and
      Zhang, Yang  and
      Fan, Chuchu",
    editor = "Chiruzzo, Luis  and
      Ritter, Alan  and
      Wang, Lu",
    booktitle = "Proceedings of the 2025 Conference of the Nations of the Americas Chapter of the Association for Computational Linguistics: Human Language Technologies (Volume 1: Long Papers)",
    month = apr,
    year = "2025",
    address = "Albuquerque, New Mexico",
    publisher = "Association for Computational Linguistics",
    url = "https://aclanthology.org/2025.naacl-long.176/",
    doi = "10.18653/v1/2025.naacl-long.176",
    pages = "3434--3483",
    ISBN = "979-8-89176-189-6"
}

@article{qu2025tripscore,
  title={TripScore: Benchmarking and rewarding real-world travel planning with fine-grained evaluation},
  author={Qu, Yincen and Xiao, Huan and Li, Feng and Li, Gregory and Zhou, Hui and Dai, Xiangying and Dai, Xiaoru},
  journal={arXiv preprint arXiv:2510.09011},
  year={2025}
}

@inproceedings{TP-RAG,
    title = "{TP}-{RAG}: Benchmarking Retrieval-Augmented Large Language Model Agents for Spatiotemporal-Aware Travel Planning",
    author = "Ni, Hang  and
      Liu, Fan  and
      Ma, Xinyu  and
      Su, Lixin  and
      Wang, Shuaiqiang  and
      Yin, Dawei  and
      Xiong, Hui  and
      Liu, Hao",
    editor = "Christodoulopoulos, Christos  and
      Chakraborty, Tanmoy  and
      Rose, Carolyn  and
      Peng, Violet",
    booktitle = "Proceedings of the 2025 Conference on Empirical Methods in Natural Language Processing",
    month = nov,
    year = "2025",
    address = "Suzhou, China",
    publisher = "Association for Computational Linguistics",
    url = "https://aclanthology.org/2025.emnlp-main.626/",
    doi = "10.18653/v1/2025.emnlp-main.626",
    pages = "12403--12429",
    ISBN = "979-8-89176-332-6",
}

@inproceedings{shao2024chinatravel,
title={ChinaTravel: An Open-Ended Benchmark for Language Agents in Chinese Travel Planning},
author={Jie-Jing Shao and Bo-Wen Zhang and Xiao-Wen Yang and Baizhi Chen and Siyu Han and Wen-Da Wei and Guohao Cai and Zhenhua Dong and Lan-Zhe Guo and Yu-Feng Li},
booktitle={Workshop on Scaling Environments for Agents},
year={2025},
url={https://openreview.net/forum?id=dPMlVo3rNy}
}

@inproceedings{deng2025retail,
  title={Retail: Towards real-world travel planning for large language models},
  author={Deng, Bin and Feng, Yizhe and Liu, Zeming and Wei, Qing and Zhu, Xiangrong and Chen, Shuai and Guo, Yuanfang and Wang, Yunhong},
  booktitle={Proceedings of the 2025 Conference on Empirical Methods in Natural Language Processing},
  pages={14881--14913},
  year={2025}
}

@inproceedings{shao-etal-personal_travel_solver,
    title = "Personal Travel Solver: A Preference-Driven {LLM}-Solver System for Travel Planning",
    author = "Shao, Zijian  and
      Wu, Jiancan  and
      Chen, Weijian  and
      Wang, Xiang",
    editor = "Che, Wanxiang  and
      Nabende, Joyce  and
      Shutova, Ekaterina  and
      Pilehvar, Mohammad Taher",
    booktitle = "Proceedings of the 63rd Annual Meeting of the Association for Computational Linguistics (Volume 1: Long Papers)",
    month = jul,
    year = "2025",
    address = "Vienna, Austria",
    publisher = "Association for Computational Linguistics",
    url = "https://aclanthology.org/2025.acl-long.1339/",
    doi = "10.18653/v1/2025.acl-long.1339",
    pages = "27622--27642",
    ISBN = "979-8-89176-251-0",
}

@article{ning2025deeptravel,
  title={DeepTravel: An End-to-End Agentic Reinforcement Learning Framework for Autonomous Travel Planning Agents},
  author={Ning, Yansong and Liu, Rui and Wang, Jun and Chen, Kai and Li, Wei and Fang, Jun and Zheng, Kan and Tan, Naiqiang and Liu, Hao},
  journal={arXiv preprint arXiv:2509.21842},
  year={2025}
}

@inproceedings{team2025kimi_k2,
  title={Kimi K2: Open Agentic Intelligence},
  author={Kimi Team Yifan Bai and Yiping Bao and Guanduo Chen and Jiahao Chen and Ningxin Chen and Ruijue Chen and Yanru Chen and Yuankun Chen and Yutian Chen and Zhuofu Chen and Jialei Cui and Haochen Ding and Meng-xiao Dong and Angang Du and Chenzhuang Du and Dikang Du and Yulun Du and Yu Fan and Yichen Feng and Kelin Fu and Bofei Gao and Hongcheng Gao and Peizhong Gao and Tong Gao and Xinran Gu and Longyu Guan and Haiqing Guo and Jia-Xing Guo and Hao-Xing Hu and Xiaoru Hao and Tian He and Weiran He and Wen He and Chao Hong and Yan-Ni Hu and Zhenxing Hu and Weixiao Huang and Zhiqi Huang and Zihao Huang and Tao Jiang and Zhejun Jiang and Xinyi Jin and Yongsheng Kang and Guokun Lai and Cheng Li and Fang Li and Haoyang Li and Ming Li and Wentao Li and Yanhao Li and Yiwei Li and Zhaowei Li and Zheming Li and Hong-Li Lin and Xiaohan Lin and Zongyu Lin and Chengyi Liu and Chenyu Liu and Hongzhang Liu and Jingyuan Liu and Junqi Liu and Liang Liu and Shaowei Liu and T. Y. Liu and Tian-Bo Liu and Weizhou Liu and Yangyang Liu and Yibo Liu and Yiping Liu and Yue Liu and Zhengying Liu and Enzhe Lu and Li Lu and Shen Ma and Xinyu Ma and Yi-Xuan Ma and Shaoguang Mao and Jie Mei and Xin Men and Yibo Miao and Siyuan Pan and Yebo Peng and Ruoyu Qin and Bowen Qu and Zeyu Shang and Li-Na Shi and Sheng-Rong Shi and Feifan Song and Jian-Fei Su and Zhen-Xin Su and Xinjie Sun and Flood Sung and Heyi Tang and Ji-Hua Tao and Qi Teng and Chensi Wang and Dinglu Wang and Feng Wang and Haiming Wang and Jianzhou Wang and Jiaxing Wang and Jinhong Wang and Shengjie Wang and Shuyi Wang and Yao Wang and Yejie Wang and Yiqin Wang and Yuxin Wang and Yuzhi Wang and Zhaoji Wang and Zhengtao Wang and Zhexu Wang and Chu Wei and Qi-Feng Wei and Wenhao Wu and Xingzhe Wu and Yuxin Wu and Chenjun Xiao and Xiao-Ming Xie and Weiming Xiong and Boyu Xu and Jing Xu and Jinjing Xu and L. H. Xu and Lin Xu and Suting Xu and Weixin Xu and Xinran Xu and Yangchuan Xu and Zi-Yang Xu and Junjie Yan and Yuzi Yan and Xiaofei Yang and Ying Yang and Zhengqi Yang and Zhilin Yang and Zonghan Yang and Haotian Yao and Xingcheng Yao and Wen-guang Ye and Zhuorui Ye and Bohong Yin and Long Yu and Enming Yuan and Hongbang Yuan and Mengjie Yuan and Haobing Zhan and Dehao Zhang and Hao Zhang and Wanlu Zhang and Xiaobin Zhang and Yangkun Zhang and Yizhi Zhang and Yongting Zhang and Yu Zhang and Yutao Zhang and Yutong Zhang and Zheng Zhang and Hao-Dong Zhao and Yikai Zhao and Huabin Zheng and Shao Jian Zheng and Jianren Zhou and Xinyu Zhou and Zaida Zhou and Zhengxin Zhu and Weiyu Zhuang and Xinxing Zu},
  year={2025},
  journal={arXiv preprint arXiv:2507.20534},
  url={https://api.semanticscholar.org/CorpusID:280323540}
}

@article{zeng2025glm_45,
  title={GLM-4.5: Agentic, Reasoning, and Coding (ARC) Foundation Models},
  author={GLM-4.5 Team Aohan Zeng and Xin Lv and Qinkai Zheng and Zhenyu Hou and Bin Chen and Chengxing Xie and Cunxiang Wang and Da Yin and Hao Zeng and Jiajie Zhang and Kedong Wang and Lucen Zhong and Mingdao Liu and Rui Lu and Shulin Cao and Xiaohan Zhang and Xuancheng Huang and Yao Wei and Yean Cheng and Yifang An and Yilin Niu and Yuan Wen and Yu Bai and Zhengxiao Du and Zihan Wang and Zilin Zhu and Bohan Zhang and Bosi Wen and Bowen Wu and Bowen Xu and Can Huang and Casey Zhao and Changpeng Cai and Chao Yu and Chen Li and Chendi Ge and Chenghuan Huang and Chenhui Zhang and Chenxi Xu and Chenzheng Zhu and Chuang Li and Congfeng Yin and Daoyan Lin and Da-Wei Yang and Da-Peng Jiang and Ding Ai and Erle Zhu and Fei-Yue Wang and Gengzheng Pan and Guo Wang and Hai Lan Sun and Haitao Li and Haiyang Li and Haiyi Hu and Hanyu Zhang and Hao Peng and Hao Tai and Haoke Zhang and Haoran Wang and Haoyu Yang and He Liu and He Zhao and Hongwei Liu and Hong Yan and Huan Liu and Huilong Chen and Ji Li and Jiajing Zhao and Jiaming Ren and Jian Jiao and Jiani Zhao and Jia-Xin Yan and Jiaqi Wang and Jiayi Gui and Jiayue Zhao and Jie Liu and Jijie Li and Jing Li and Jin Lu and Jingsen Wang and Jing Yuan and Jingxuan Li and Jin-Cheng Du and Jinhua Du and Jinxin Liu and Junkai Zhi and Jun Gao and Kedong Wang and Lekang Yang and Liang Xu and Linxi Jim Fan and Lindong Wu and Lintao Ding and Lu Wang and Man Zhang and Minghao Li and Ming-wei Xu and Mingming Zhao and Mingshu Zhai and Pengfan Du and Qian Dong and Shangde Lei and Shangqing Tu and Shangtong Yang and Shaoyou Lu and Shijie Li and Shuang Li and Shuang-li and Shuxun Yang and Sibo Yi and Tianshu Yu and Wei Tian and Weihan Wang and Wenbo Yu and Weng Lam Tam and Wenjie Liang and Wentao Liu and Xiao Wang and Xiao-Zhou Jia and Xia Gu and Xiao Ling and Xin Wang and Xing Fan and Xingru Pan and Xinyuan Zhang and Xinze Zhang and Xiu-hua Fu and Xunkai Zhang and Yabo Xu and Ya-nan Wu and Yida Lu and Yidong Wang and Yilin Zhou and Yi-Ji Pan and Yiming Pan and Ying Zhang and Yingli Wang and Yingru Li and Yinpei Su and Yi Geng and Yitong Zhu and Yongkun Yang and Yuhang Li and Yuhao Wu and Yujiang Li and Yun-Hao Liu and Yunqing Wang and Yuntao Li and Yuxuan Zhang and Ze-Xian Liu and Zhen Yang and Zhen Yu Zhou and Zhongpei Qiao and Zhuoer Feng and Zhuo-Gang Liu and Zichen Zhang and Zijun Yao and Zikang Wang and Ziqiang Liu and Ziwei Chai and Zixuan Li and Zuodong Zhao and Wenguang Chen and Jidong Zhai and Bin Xu and Minlie Huang and Hongning Wang and Juanzi Li and Yu-ying Dong and Jie Tang},
  journal={ArXiv},
  year={2025},
  volume={abs/2508.06471},
  url={https://api.semanticscholar.org/CorpusID:280561359}
}

@article{liu2025deepseek,
  title={DeepSeek-V3.2: Pushing the Frontier of Open Large Language Models},
  author={DeepSeek-AI and Aixin Liu and Aoxue Mei and Ban Lin and Bing Xue and Bing-Li Wang and Bin Xu and Bochao Wu and Bowei Zhang and Chaofan Lin and Chen Dong and Chengda Lu and Chenggang Zhao and Chengqi Deng and Chenhao Xu and Chong Ruan and Damai Dai and Daya Guo and Dejian Yang and Deli Chen and Erhang Li and Fangqi Zhou and Fangyun Lin and Fucong Dai and Guangbo Hao and Guanting Chen and Guowei Li and H. Zhang and Hanwei Xu and Hao Li and Hao Liang and Haoran Wei and Haowei Zhang and Hao-sheng Luo and Haozhe Ji and Honghui Ding and Hongxuan Tang and Huan Cao and Huazuo Gao and Huixian Qu and Hui Zeng and Jialiang Huang and Jiashi Li and Jiaxin Xu and Jiewen Hu and JingChang Chen and Ji Xiang and Jingyang Yuan and Jing Cheng and Jinhua Zhu and Jun Ran and Junguang Jiang and Junjie Qiu and Junlong Li and Jun-Mei Song and Kai Dong and Kaige Gao and Kang Guan and Kexin Huang and Kexin Zhou and Ke-wei Huang and Kuai Yu and Lean Wang and Lecong Zhang and Lei Wang and Liang Zhao and Liangsheng Yin and Lihua Guo and Ling-Li Luo and Lin Ma and Litong Wang and Liyue Zhang and M. S. Di and M. Y. Xu and Mingchuan Zhang and Minghua Zhang and Min Tang and Mingxu Zhou and Pan Huang and Peixin Cong and Peiyi Wang and Qiancheng Wang and Qihao Zhu and Qingyang Li and Qinyu Chen and Qiushi Du and Ruiling Xu and Ruiqi Ge and Ruisong Zhang and Ruizhe Pan and Runji Wang and Runqiu Yin and Runxin Xu and Ru Shen and Ruoyu Zhang and S. H. Liu and Shanghao Lu and Shangyan Zhou and Shanhuang Chen and Shaofei Cai and Shaoyuan Chen and Shengding Hu and Shengyu Liu and Shiqiang Hu and Shirong Ma and Shiyu Wang and Shuiping Yu and Shunfeng Zhou and Shuting Pan and Songyang Zhou and Tao Ni and Tao Yun and Tian Pei and Tian Ye and Tianyuan Yue and Wangding Zeng and Wen Liu and Wenfeng Liang and Wenjie Pang and Wenjing Luo and Wenjun Gao and Wentao Zhang and Xi Gao and Xiangwen Wang and Xiaoling Bi and Xiaodong Liu and Xiaohan Wang and Xiaokang Chen and Xiaokang Zhang and Xiaotao Nie and Xin Cheng and Xin Liu and Xin Xie and Xingchao Liu and Xingkai Yu and Xingyou Li and Xinyu Yang and Xinyuan Li and Xu Chen and Xuecheng Su and Xuehai Pan and Xuheng Lin and Xu Fu and Y. Q. Wang and Yang Zhang and Yanhong Xu and Yanru Ma and Yao Li and Yao Zhao and Yaofeng Sun and Yaohui Wang and Yi Qian and Yingpu Yu and Yichao Zhang and Yifan Ding and Yifan Shi and Yi Xiong and Ying He and Ying Zhou and Yinmin Zhong and Yishi Piao and Yisong Wang and Yixiao Chen and Yixuan Tan and Yixuan Wei and Yiyang Ma and Yiyuan Liu and Yong Yang and Yongqiang Guo and Yongtong Wu and Yu Wu and Yuan Cheng and Yuan Ou and Yuanfang Xu and Yuduan Wang and Yue Gong and Yuhan Wu and Yu-Hui Zou and Yukun Li and Yunfan Xiong and Yu-Wei Luo and Yu-mei You and Yuxuan Liu and Yuyang Zhou and Z. F. Wu and Zehui Ren and Zehua Zhao and Zehui Ren and Zhangli Sha and Zhe Fu and Zhean Xu and Zhenda Xie and Zhen-guo Zhang and Zhewen Hao and Zhibin Gou and Zhicheng Ma and Zhigang Yan and Zhihong Shao and Zhixian Huang and Zhiyu Wu and Zhuoshu Li and Zhuping Zhang and Zian Xu and Zihao Wang and Zihui Gu and Zijia Zhu and Zi-Rui Li and Zipeng Zhang and Ziwei Xie and Ziyi Gao and Zizheng Pan and Z. P. Yao and Bei Feng and Hui Li and J. L. Cai and Jiaqi Ni and Lei Xu and Meng Li and Ning Tian and R. J. Chen and Ruiqi Jin and S. S. Li and Shuang Zhou and Tianyu Sun and X. Q. Li and Xiangyu Jin and Xiaojin Shen and Xiaosha Chen and Xinnan Song and Xinyi Zhou and Y. X. Zhu and Yanping Huang and Yao Li and Yi Zheng and Yuchen Zhu and Yunxiang Ma and Zhen Huang and Zhipeng Xu and Zhongyu Zhang and Dong-Li Ji and Jian Liang and Jianzhong Guo and Jin Chen and Leyi Xia and Miaojun Wang and Mingming Li and Peng Zhang and Ruyi Chen and Shangmian Sun and Shao-Kang Wu and Sheng-Ying Ye and T.Wang and Weizhen Xiao and Wei An and Xianzu Wang and Xiaowen Sun and Xiaoxiang Wang and Ying Tang and Yukun Zha and Ze-Na Zhang and Zhenghua Ju and Zhen Zhang and Zihua Qu},
  journal={ArXiv},
  year={2025},
  volume={abs/2512.02556},
  url={https://api.semanticscholar.org/CorpusID:283448719}
}

@inproceedings{yang2025qwen3,
  title={Qwen3 Technical Report},
  author={An Yang and Anfeng Li and Baosong Yang and Beichen Zhang and Binyuan Hui and Bo Zheng and Bowen Yu and Chang Gao and Chengen Huang and Chenxu Lv and Chujie Zheng and Dayiheng Liu and Fan Zhou and Fei Huang and Feng Hu and Hao Ge and Haoran Wei and Huan Lin and Jialong Tang and Jian Yang and Jianhong Tu and Jianwei Zhang and Jianxin Yang and Jiaxin Yang and Jingren Zhou and Jingren Zhou and Junyan Lin and Kai Dang and Keqin Bao and Ke‐Pei Yang and Le Yu and Li-Chun Deng and Mei Li and Min Xue and Mingze Li and Pei Zhang and Peng Wang and Qin Zhu and Rui Men and Ruize Gao and Shi-Qiang Liu and Shuang Luo and Tianhao Li and Tianyi Tang and Wenbiao Yin and Xingzhang Ren and Xinyu Wang and Xinyu Zhang and Xuancheng Ren and Yang Fan and Yang Su and Yi-Chao Zhang and Yinger Zhang and Yu Wan and Yuqiong Liu and Zekun Wang and Zeyu Cui and Zhenru Zhang and Zhipeng Zhou and Zihan Qiu},
  year={2025},
  journal={arXiv preprint arXiv:2505.09388},
  url={https://api.semanticscholar.org/CorpusID:278602855}
}

@inproceedings{guo2024stabletoolbench,
    title = "{S}table{T}ool{B}ench: Towards Stable Large-Scale Benchmarking on Tool Learning of Large Language Models",
    author = "Guo, Zhicheng  and
      Cheng, Sijie  and
      Wang, Hao  and
      Liang, Shihao  and
      Qin, Yujia  and
      Li, Peng  and
      Liu, Zhiyuan  and
      Sun, Maosong  and
      Liu, Yang",
    editor = "Ku, Lun-Wei  and
      Martins, Andre  and
      Srikumar, Vivek",
    booktitle = "Findings of the Association for Computational Linguistics: ACL 2024",
    month = aug,
    year = "2024",
    address = "Bangkok, Thailand",
    publisher = "Association for Computational Linguistics",
    url = "https://aclanthology.org/2024.findings-acl.664/",
    doi = "10.18653/v1/2024.findings-acl.664",
    pages = "11143--11156"
}

@inproceedings{brown2020gpt3,
 author = {Brown, Tom and Mann, Benjamin and Ryder, Nick and Subbiah, Melanie and Kaplan, Jared D and Dhariwal, Prafulla and Neelakantan, Arvind and Shyam, Pranav and Sastry, Girish and Askell, Amanda and Agarwal, Sandhini and Herbert-Voss, Ariel and Krueger, Gretchen and Henighan, Tom and Child, Rewon and Ramesh, Aditya and Ziegler, Daniel and Wu, Jeffrey and Winter, Clemens and Hesse, Chris and Chen, Mark and Sigler, Eric and Litwin, Mateusz and Gray, Scott and Chess, Benjamin and Clark, Jack and Berner, Christopher and McCandlish, Sam and Radford, Alec and Sutskever, Ilya and Amodei, Dario},
 booktitle = {Advances in Neural Information Processing Systems},
 editor = {H. Larochelle and M. Ranzato and R. Hadsell and M.F. Balcan and H. Lin},
 pages = {1877--1901},
 publisher = {Curran Associates, Inc.},
 title = {Language Models are Few-Shot Learners},
 url = {https://proceedings.neurips.cc/paper_files/paper/2020/file/1457c0d6bfcb4967418bfb8ac142f64a-Paper.pdf},
 volume = {33},
 year = {2020}
}

@article{karmakar2025triptide,
  title={TripTide: A Benchmark for Adaptive Travel Planning under Disruptions},
  author={Karmakar, Priyanshu and Chaudhuri, Soumyabrata and Mallick, Shubhojit and Gupta, Manish and Jana, Abhik and Ghosh, Shreya},
  journal={arXiv preprint arXiv:2510.21329},
  year={2025}
}

@inproceedings{zhang-ask-before-plan,
    title = "Ask-before-Plan: Proactive Language Agents for Real-World Planning",
    author = "Zhang, Xuan  and
      Deng, Yang  and
      Ren, Zifeng  and
      Ng, See-Kiong  and
      Chua, Tat-Seng",
    editor = "Al-Onaizan, Yaser  and
      Bansal, Mohit  and
      Chen, Yun-Nung",
    booktitle = "Findings of the Association for Computational Linguistics: EMNLP 2024",
    month = nov,
    year = "2024",
    address = "Miami, Florida, USA",
    publisher = "Association for Computational Linguistics",
    url = "https://aclanthology.org/2024.findings-emnlp.636/",
    doi = "10.18653/v1/2024.findings-emnlp.636",
    pages = "10836--10863"
}

@inproceedings{wang-etal-2025-triptailor,
    title = "{T}rip{T}ailor: A Real-World Benchmark for Personalized Travel Planning",
    author = "Wang, Kaimin  and
      Shen, Yuanzhe  and
      Lv, Changze  and
      Zheng, Xiaoqing  and
      Huang, Xuanjing",
    editor = "Che, Wanxiang  and
      Nabende, Joyce  and
      Shutova, Ekaterina  and
      Pilehvar, Mohammad Taher",
    booktitle = "Findings of the Association for Computational Linguistics: ACL 2025",
    month = jul,
    year = "2025",
    address = "Vienna, Austria",
    publisher = "Association for Computational Linguistics",
    url = "https://aclanthology.org/2025.findings-acl.503/",
    doi = "10.18653/v1/2025.findings-acl.503",
    pages = "9705--9723",
    ISBN = "979-8-89176-256-5",
}

@article{qin2025compass,
  title={COMPASS: A Multi-Turn Benchmark for Tool-Mediated Planning \& Preference Optimization},
  author={Tian Qin and Felix Bai and Ting-Yao Hu and Raviteja Vemulapalli and Hema Swetha Koppula and Zhiyang Xu and Bowen Jin and Mert Cemri and Jiarui Lu and Zirui Wang and Meng Cao},
  journal={ArXiv},
  year={2025},
  volume={abs/2510.07043},
  url={https://api.semanticscholar.org/CorpusID:281892403}
}

@inproceedings{oh2025flex-travelplanner,
title={{FLEX}-{TRAVELPLANNER}: A {BENCHMARK} {FOR} {FLEXIBLE} {PLANNING} {WITH} {LANGUAGE} {AGENTS}},
author={Juhyun Oh and Eunsu Kim and Alice Oh},
booktitle={Workshop on Reasoning and Planning for Large Language Models},
year={2025},
url={https://openreview.net/forum?id=a7unQ5jMx7}
}

@inproceedings{jung-etal-2025-can-llm-Truly-plan,
    title = "Can {LLM}s Truly Plan? A Comprehensive Evaluation of Planning Capabilities",
    author = "Jung, Gayeon  and
      Lim, HyeonSeok  and
      Kim, Minjun  and
      Lim, Joon-ho  and
      Lim, KyungTae  and
      Kim, Hansaem",
    editor = "Christodoulopoulos, Christos  and
      Chakraborty, Tanmoy  and
      Rose, Carolyn  and
      Peng, Violet",
    booktitle = "Findings of the Association for Computational Linguistics: EMNLP 2025",
    month = nov,
    year = "2025",
    address = "Suzhou, China",
    publisher = "Association for Computational Linguistics",
    url = "https://aclanthology.org/2025.findings-emnlp.702/",
    doi = "10.18653/v1/2025.findings-emnlp.702",
    pages = "13069--13084",
    ISBN = "979-8-89176-335-7"
}

@article{choi2025atlas,
  title={ATLAS: Constraints-Aware Multi-Agent Collaboration for Real-World Travel Planning},
  author={Choi, Jihye and Yoon, Jinsung and Chen, Jiefeng and Jha, Somesh and Pfister, Tomas},
  journal={arXiv preprint arXiv:2509.25586},
  year={2025}
}

@inproceedings{qian2025toolrl,
title={Tool{RL}: Reward is All Tool Learning Needs},
author={Cheng Qian and Emre Can Acikgoz and Qi He and Hongru WANG and Xiusi Chen and Dilek Hakkani-T{\"u}r and Gokhan Tur and Heng Ji},
booktitle={The Thirty-ninth Annual Conference on Neural Information Processing Systems},
year={2025},
url={https://openreview.net/forum?id=eOLdGbXT6t}
}

@inproceedings{zhang-planning-with-multi-constraints,
    title = "Planning with Multi-Constraints via Collaborative Language Agents",
    author = "Zhang, Cong  and
      Goh, Xin Deik  and
      Li, Dexun  and
      Zhang, Hao  and
      Liu, Yong",
    editor = "Rambow, Owen  and
      Wanner, Leo  and
      Apidianaki, Marianna  and
      Al-Khalifa, Hend  and
      Eugenio, Barbara Di  and
      Schockaert, Steven",
    booktitle = "Proceedings of the 31st International Conference on Computational Linguistics",
    month = jan,
    year = "2025",
    address = "Abu Dhabi, UAE",
    publisher = "Association for Computational Linguistics",
    url = "https://aclanthology.org/2025.coling-main.672/",
    pages = "10054--10082",
}

@misc{openai_gpt51_2025,
  author       = {{OpenAI}},
  title        = {{GPT-5.1 is Now Available: A Smarter and More Conversational ChatGPT}},
  howpublished = {\url{https://openai.com/index/gpt-5-1/}},
  year         = {2025},
  month        = {nov},
  note         = {Accessed: 2025-11-12}
}

@misc{google_gemini3_2025,
  author       = {{Google}},
  title        = {{A New Era of Intelligence with Gemini 3}},
  howpublished = {\url{https://blog.google/products/gemini/gemini-3/}},
  year         = {2025},
  month        = {dec},
  note         = {Accessed: 2025-12-18}
}

@misc{minimax_2025,
  author       = {{MiniMax}},
  title        = {{MiniMax M2 \& Agent: Great Skill Concealed in Simplicity}},
  howpublished = {\url{https://www.minimaxi.com/en/news/minimax-m2}},
  year         = {2025},
  month        = {oct},
  note         = {Accessed: 2025-10-27}
}

@article{pathak2025rubric-is-all-you-need,
  title={Rubric Is All You Need: Improving LLM-Based Code Evaluation With Question-Specific Rubrics},
  author={Aditya Pathak and Rachit Gandhi and Vaibhav Uttam and Devansh and Yashwanth Kumar Nakka and Aaryan Raj Jindal and Pratyush Ghosh and Arnav Ramamoorthy and Shreyash Verma and Aditya Mittal and Aashna Ased and Chirag Khatri and Jagat Sesh Challa and Dhruv Kumar},
  journal={Proceedings of the 2025 ACM Conference on International Computing Education Research V.1},
  year={2025},
  url={https://api.semanticscholar.org/CorpusID:277452105}
}

@inproceedings{hashemi2024llm-rubric,
    title = "{LLM}-Rubric: A Multidimensional, Calibrated Approach to Automated Evaluation of Natural Language Texts",
    author = "Hashemi, Helia  and
      Eisner, Jason  and
      Rosset, Corby  and
      Van Durme, Benjamin  and
      Kedzie, Chris",
    editor = "Ku, Lun-Wei  and
      Martins, Andre  and
      Srikumar, Vivek",
    booktitle = "Proceedings of the 62nd Annual Meeting of the Association for Computational Linguistics (Volume 1: Long Papers)",
    month = aug,
    year = "2024",
    address = "Bangkok, Thailand",
    publisher = "Association for Computational Linguistics",
    url = "https://aclanthology.org/2024.acl-long.745/",
    doi = "10.18653/v1/2024.acl-long.745",
    pages = "13806--13834"
}

@inproceedings{sener2018k-center-greedy,
title={Active Learning for Convolutional Neural Networks: A Core-Set Approach},
author={Ozan Sener and Silvio Savarese},
booktitle={International Conference on Learning Representations},
year={2018},
url={https://openreview.net/forum?id=H1aIuk-RW},
}

@article{xie2024human-like,
  title={A human-like reasoning framework for multi-phases planning task with large language models},
  author={Xie, Chengxing and Zou, Difan},
  journal={arXiv preprint arXiv:2405.18208},
  year={2024}
}

@article{liu2025vaiage,
  title={Vaiage: A Multi-Agent Solution to Personalized Travel Planning},
  author={Liu, Binwen and Ge, Jiexi and Wang, Jiamin},
  journal={arXiv preprint arXiv:2505.10922},
  year={2025}
}

@inproceedings{ju-etal-2024-TTG,
    title = "To the Globe ({TTG}): Towards Language-Driven Guaranteed Travel Planning",
    author = "Ju, Da  and
      Jiang, Song  and
      Cohen, Andrew  and
      Foss, Aaron  and
      Mitts, Sasha  and
      Zharmagambetov, Arman  and
      Amos, Brandon  and
      Li, Xian  and
      Kao, Justine T  and
      Fazel-Zarandi, Maryam  and
      Tian, Yuandong",
    editor = "Hernandez Farias, Delia Irazu  and
      Hope, Tom  and
      Li, Manling",
    booktitle = "Proceedings of the 2024 Conference on Empirical Methods in Natural Language Processing: System Demonstrations",
    month = nov,
    year = "2024",
    address = "Miami, Florida, USA",
    publisher = "Association for Computational Linguistics",
    url = "https://aclanthology.org/2024.emnlp-demo.25/",
    doi = "10.18653/v1/2024.emnlp-demo.25",
    pages = "240--249"
}

@inproceedings{qin2023toolllm,
title={Tool{LLM}: Facilitating Large Language Models to Master 16000+ Real-world {API}s},
author={Yujia Qin and Shihao Liang and Yining Ye and Kunlun Zhu and Lan Yan and Yaxi Lu and Yankai Lin and Xin Cong and Xiangru Tang and Bill Qian and Sihan Zhao and Lauren Hong and Runchu Tian and Ruobing Xie and Jie Zhou and Mark Gerstein and dahai li and Zhiyuan Liu and Maosong Sun},
booktitle={The Twelfth International Conference on Learning Representations},
year={2024},
url={https://openreview.net/forum?id=dHng2O0Jjr}
}

@article{guo2025deepseekr1,
  title={DeepSeek-R1 incentivizes reasoning in LLMs through reinforcement learning},
  author={DeepSeek-AI and Daya Guo and Dejian Yang and Haowei Zhang and Jun-Mei Song and Ruoyu Zhang and Runxin Xu and Qihao Zhu and Shirong Ma and Peiyi Wang and Xiaoling Bi and Xiaokang Zhang and Xingkai Yu and Yu Wu and Z. F. Wu and Zhibin Gou and Zhihong Shao and Zhuoshu Li and Ziyi Gao and Aixin Liu and Bing Xue and Bing-Li Wang and Bochao Wu and Bei Feng and Chengda Lu and Chenggang Zhao and Chengqi Deng and Chenyu Zhang and Chong Ruan and Damai Dai and Deli Chen and Dong-Li Ji and Erhang Li and Fangyun Lin and Fucong Dai and Fuli Luo and Guangbo Hao and Guanting Chen and Guowei Li and H. Zhang and Han Bao and Hanwei Xu and Haocheng Wang and Honghui Ding and Huajian Xin and Huazuo Gao and Hui Qu and Hui Li and Jianzhong Guo and Jiashi Li and Jiawei Wang and JingChang Chen and Jingyang Yuan and Junjie Qiu and Junlong Li and J. L. Cai and Jiaqi Ni and Jian Liang and Jin Chen and Kai Dong and Kai Hu and Kaige Gao and Kang Guan and Kexin Huang and Kuai Yu and Lean Wang and Lecong Zhang and Liang Zhao and Litong Wang and Liyue Zhang and Lei Xu and Leyi Xia and Mingchuan Zhang and Minghua Zhang and M. Tang and Meng Li and Miaojun Wang and Mingming Li and Ning Tian and Panpan Huang and Peng Zhang and Qiancheng Wang and Qinyu Chen and Qiushi Du and Ruiqi Ge and Ruisong Zhang and Ruizhe Pan and Runji Wang and R. J. Chen and Ruiqi Jin and Ruyi Chen and Shanghao Lu and Shangyan Zhou and Shanhuang Chen and Shengfeng Ye and Shiyu Wang and Shuiping Yu and Shunfeng Zhou and Shuting Pan and S. S. Li and Shuang Zhou and Shao-Kang Wu and Tao Yun and Tian Pei and Tianyu Sun and T. Wang and Wangding Zeng and Wanjia Zhao and Wen Liu and Wenfeng Liang and Wenjun Gao and Wen-Xia Yu and Wentao Zhang and Wangding Xiao and Wei An and Xiaodong Liu and Xiaohan Wang and Xiaokang Chen and Xiaotao Nie and Xin Cheng and Xin Liu and Xin Xie and Xingchao Liu and Xinyu Yang and Xinyuan Li and Xuecheng Su and Xuheng Lin and X. Q. Li and Xiangyu Jin and Xi-Cheng Shen and Xiaosha Chen and Xiaowen Sun and Xiaoxiang Wang and Xinnan Song and Xinyi Zhou and Xianzu Wang and Xinxia Shan and Y. K. Li and Y. Q. Wang and Y. X. Wei and Yang Zhang and Yanhong Xu and Yao Li and Yao Zhao and Yaofeng Sun and Yaohui Wang and Yi Yu and Yichao Zhang and Yifan Shi and Yi Xiong and Ying He and Yishi Piao and Yisong Wang and Yixuan Tan and Yiyang Ma and Yiyuan Liu and Yongqiang Guo and Yuan Ou and Yuduan Wang and Yue Gong and Yu-Jing Zou and Yujia He and Yunfan Xiong and Yu-Wei Luo and Yu-mei You and Yuxuan Liu and Yuyang Zhou and Y. X. Zhu and Yanping Huang and Yao Li and Yi Zheng and Yuchen Zhu and Yunxiang Ma and Ying Tang and Yukun Zha and Yuting Yan and Zehui Ren and Zehui Ren and Zhangli Sha and Zhe Fu and Zhean Xu and Zhenda Xie and Zhen-guo Zhang and Zhewen Hao and Zhicheng Ma and Zhigang Yan and Zhiyu Wu and Zihui Gu and Zijia Zhu and Zijun Liu and Zi-An Li and Ziwei Xie and Ziyang Song and Zizheng Pan and Zhen Huang and Zhipeng Xu and Zhongyu Zhang and Zhen Zhang},
  journal={Nature},
  year={2025},
  volume={645},
  pages={633 - 638},
  url={https://api.semanticscholar.org/CorpusID:275789950}
}

@article{zheng2024natural,
  title={NATURAL PLAN: Benchmarking LLMs on Natural Language Planning},
  author={Huaixiu Steven Zheng and Swaroop Mishra and Hugh Zhang and Xinyun Chen and Minmin Chen and Azade Nova and Le Hou and Heng-tze Cheng and Quoc V. Le and Ed Huai-hsin Chi and Denny Zhou},
  journal={ArXiv},
  year={2024},
  volume={abs/2406.04520},
  url={https://api.semanticscholar.org/CorpusID:270357954}
}

@article{chen2024travelagent,
  title={Travelagent: An ai assistant for personalized travel planning},
  author={Chen, Aili and Ge, Xuyang and Fu, Ziquan and Xiao, Yanghua and Chen, Jiangjie},
  journal={arXiv preprint arXiv:2409.08069},
  year={2024}
}

\appendix
\clearpage
\section{More Discussion}
\label{app:sec:more_discussion}

\textbf{Benchmark scalability.}
TravelBench not only provides a stable evaluation platform for tool-augmented travel planning, but also releases additional resources, including full real user queries, user preferences, and the sandbox cache that covers these queries. These resources can support future dataset expansion for both training and testing. For example, in the current release, each query is paired with a single user preference. By recombining queries with different preferences, one can construct substantially more multi-turn scenarios. Since user simulation is dynamic, different preferences can lead to different dialogue trajectories even for the same query. We therefore view TravelBench as a valuable research resource that can be extended into larger datasets.

\textbf{Benchmark practicality.}
As shown in Table~\ref{tab:main-results}, advanced models (e.g., DeepSeek-V3.2) already achieve strong performance on both single-turn and multi-turn settings (above 80). This may raise concerns about the long-term usefulness of the benchmark. Our goal, however, is to reflect model behavior on real-world travel requests, which are typically of medium-to-easy difficulty; the observed score distribution is consistent with this reality. Meanwhile, as models improve, TravelBench can remain challenging by sampling harder instances according to the difficulty distribution, and scalability allows us to increase the benchmark size when needed. Moreover, strong performance on difficult cases does not necessarily imply robust performance across all cases, especially in real-world travel planning. Therefore, a benchmark with a realistic and well-calibrated difficulty distribution is practically important. In future work, we will explore difficulty-aware sampling to provide test subsets suitable for models with different capability levels.

\section{Benchmark Details}
\label{app:sec:benchmark_details}

\begin{table}[t]
\centering
\caption{Composition of TravelBench and basic statistics. \textit{User turns} denotes the number of interaction rounds in multi-turn instances, and \textit{Tool steps} denotes the number of reasoning steps (tool calls) in single-turn instances.}
\small
\setlength{\tabcolsep}{7pt} 
\renewcommand{\arraystretch}{1.1} 
\begin{tabular}{@{}l r c@{}}
\toprule
\textbf{Subset} & \textbf{\#Inst.} & \textbf{Statistics (Mean / Min / Max)} \\ 
\midrule
Single-Turn & 500 & \textbf{Tool steps}: 3.81 / 1.33 / 11.25 \\
Multi-Turn  & 500 & \textbf{User turns}: 2.51 / 1.25 / 4.50 \\
Unsolvable  & 100 & --- \\ 
\midrule
\textbf{Total} & \textbf{1,100} &  --- \\
\bottomrule
\end{tabular}
\label{tab:dataset_composition}
\end{table}

\begin{table*}[t]
\caption{Field definitions for a TravelBench instance.}
\label{tab:dataset_fields}
\centering
\small
\renewcommand{\arraystretch}{1.1} 
\begin{tabularx}{0.95\textwidth}{@{} >{\raggedright\arraybackslash\texttt}p{0.20\textwidth} X @{}} 
\toprule
\textbf{\textrm{Field}} & \textbf{Description} \\
\midrule
trace\_id & A unique identifier for each instance. \\
\midrule
time & The timestamp of the user query. It spans from Aug.\ 2025 to Nov.\ 2025 and covers all times of day. \\
\midrule
query & The user’s raw query without any post-processing. It may contain noise, unclear expressions, or disfluencies, and is used as the assistant input. \\
\midrule
intent & The inferred underlying user intent, used as information for the user simulator. We assume the user knows their true intent. \\
\midrule
primary\_intent & The intent category label from the original data, produced by an internal model annotator. \\
\midrule
user\_profile & A de-identified user profile that preserves real preference information, used as information for the user simulator. \\
\midrule
missing\_info & Our annotation. \texttt{true} indicates the instance is infeasible due to missing required information. \\
\midrule
missing\_tool & Our annotation. \texttt{true} indicates the instance is infeasible due to missing required tools. \\
\midrule
no\_actionable & Our annotation. \texttt{true} indicates the instance is infeasible because the user intent is unclear or not actionable. \\
\midrule
context & Contextual information available at query time, including the user location and navigation-related information, used as information for both assistant and user simulator.\\
\midrule
avg\_tool\_calls &  The average number of tool calls across the four models can to some extent serve as a proxy for the difficulty of a query, as discussed in Section~\ref{difficulty-distribution}. \\
\bottomrule
\end{tabularx}
\end{table*}

\begin{figure*}[t]
    \centering
    \includegraphics[width=1.0\textwidth]{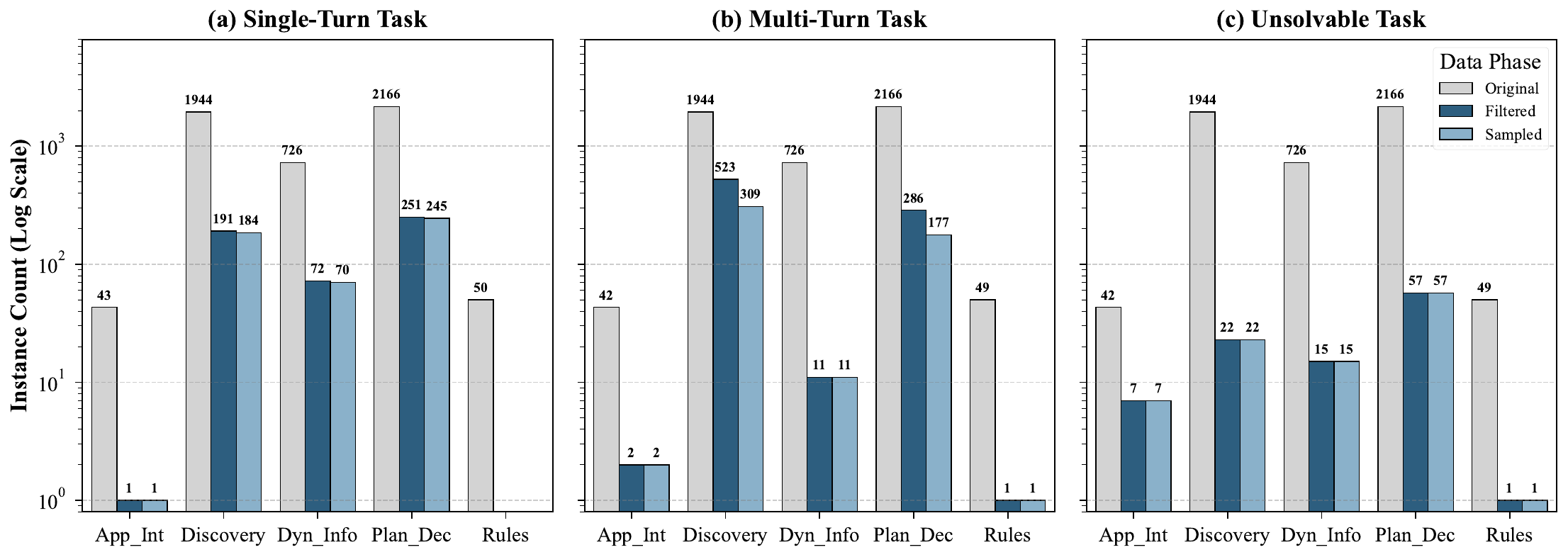}
    \caption{Data distribution across three sub-tasks: (a) Single-turn, (b) Multi-turn, and (c) Unsolvable tasks. Each category illustrates the data flow from the \textit{Original} pool to the \textit{Filtered} subset and the final \textit{Sampled} set used in our experiments. The y-axis is on a log scale with exact instance counts annotated to ensure visibility for low-frequency categories. Category abbreviations: 
    \textbf{App\_Int}: Application Interaction; 
    \textbf{Discovery}: Discovery; 
    \textbf{Dyn\_Info}: Dynamic Information; 
    \textbf{Plan\_Dec}: Planning and Decision; 
    \textbf{Rules}: Rules and Policies.}
    \label{fig:data_distribution}
\end{figure*}

\begin{figure*}[t]
    \centering
    \includegraphics[width=1.0\textwidth]{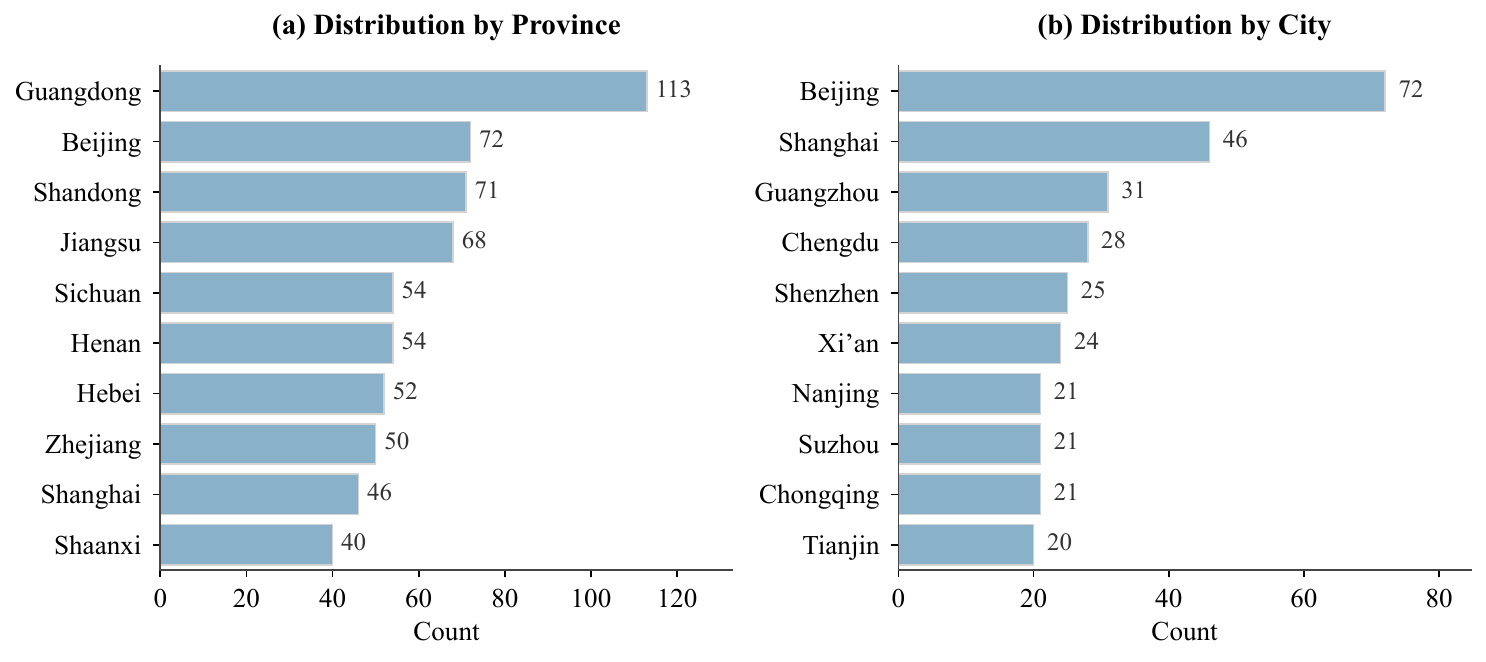}
    \caption{Query distribution by geographic regions and cities.}
    \label{fig:city-distribution}
\end{figure*}

\subsection{Dataset Composition}

Table~\ref{tab:dataset_composition} summarizes the composition of TravelBench across subtasks. Table~\ref{tab:dataset_fields} lists all fields in each instance with detailed descriptions, and Figure~\ref{app:data_example} provides a concrete example. Notably, we keep the raw user \texttt{query} without normalization. As a result, it may contain unclear wording, noise, or disfluent expressions, and is directly used as the assistant input. In contrast, the user simulator is provided with an explicit \texttt{intent}, since real users typically have a clear underlying goal driven by their preferences. This design approximates real-world interaction settings as closely as possible.

\subsection{Dataset Coverage and Distribution Analysis}
\label{app:sec:dataset_coverage}
To provide a comprehensive view of TravelBench, we analyze the dataset from both static and dynamic perspectives. We first examine its intrinsic composition and coverage, including the distributions of task categories, geographic regions, temporal span, user groups, and preference patterns. We then analyze the reasoning-step and interaction-turn distributions, which are estimated by averaging model trajectories over the evaluated systems, to characterize the practical reasoning and interaction complexity of benchmark instances. Together, these analyses show that TravelBench not only covers diverse real-world travel scenarios and user backgrounds, but also poses varied demands on multi-step reasoning and multi-turn interaction.

\subsubsection{Category Distribution}
Figure~\ref{fig:data_distribution} shows the category distribution of TravelBench, including the \textit{original} distribution, the \textit{filtered} distribution, and the \textit{sampled} distribution. The original distribution is obtained from about 5,000 instances after initial screening and deduplication, and is already diverse and representative. Filtering refers to task-specific filtering for each subset (e.g., removing infeasible instances or selecting single-turn queries; see Section~\ref{sec:data-collection} for details). Sampling is then performed to balance class frequencies while preserving diversity, where categories are defined by the \texttt{primary\_intent} field.

We observe that \textit{Discovery} and \textit{Planning and Decision} account for the largest portion of the original data. After filtering, the multi-turn subset contains more queries, suggesting that many real-world requests require iterative clarification with users. In our sampling step, we largely preserve the original distribution to reflect real user intent distributions in the benchmark. Meanwhile, since we do not provide tools for certain types of user interaction (e.g., interacting with on-device applications) or for querying real-time road rules/policies, most instances in the \textit{Application Interaction} and \textit{Rules and Policies} categories are labeled as infeasible.

Finally, prior work on travel planning typically focuses on trip itinerary planning, which corresponds to only a subcategory under our \textit{Planning and Decision} class. TravelBench substantially broadens the covered task space beyond this setting.

\subsubsection{Geographic Distribution}
We analyze geographic coverage over all 1,000 queries, including both single-turn and multi-turn instances, by extracting the provinces, municipalities, autonomous regions, special administrative regions (SARs), and cities explicitly mentioned in the queries. In total, the dataset covers 32 provinces-level administrative regions and 243 cities, indicating broad geographic coverage across China.

Figure~\ref{fig:city-distribution} presents the overall geographic distribution. The most frequently mentioned provinces include Guangdong (113, 11.31\%), Beijing (72, 7.21\%), Shandong (71, 7.11\%), Jiangsu (68, 6.81\%), and Sichuan (54, 5.41\%). At the city level, Beijing (72, 7.21\%), Shanghai (46, 4.60\%), Guangzhou (31, 3.10\%), Chengdu (28, 2.80\%), and Shenzhen (25, 2.50\%) are the most common. These results suggest that the dataset not only covers major metropolitan areas, but also includes a wide range of geographically diverse regions.

\subsubsection{Temporal Distribution}
The data collection period spans from 2025-08-14 to 2025-11-27, covering both regular days and major holiday periods such as the National Day holiday. This time span helps diversify user intents and query scenarios.

\begin{figure}[t]
    \centering
    \includegraphics[width=1.0\columnwidth]{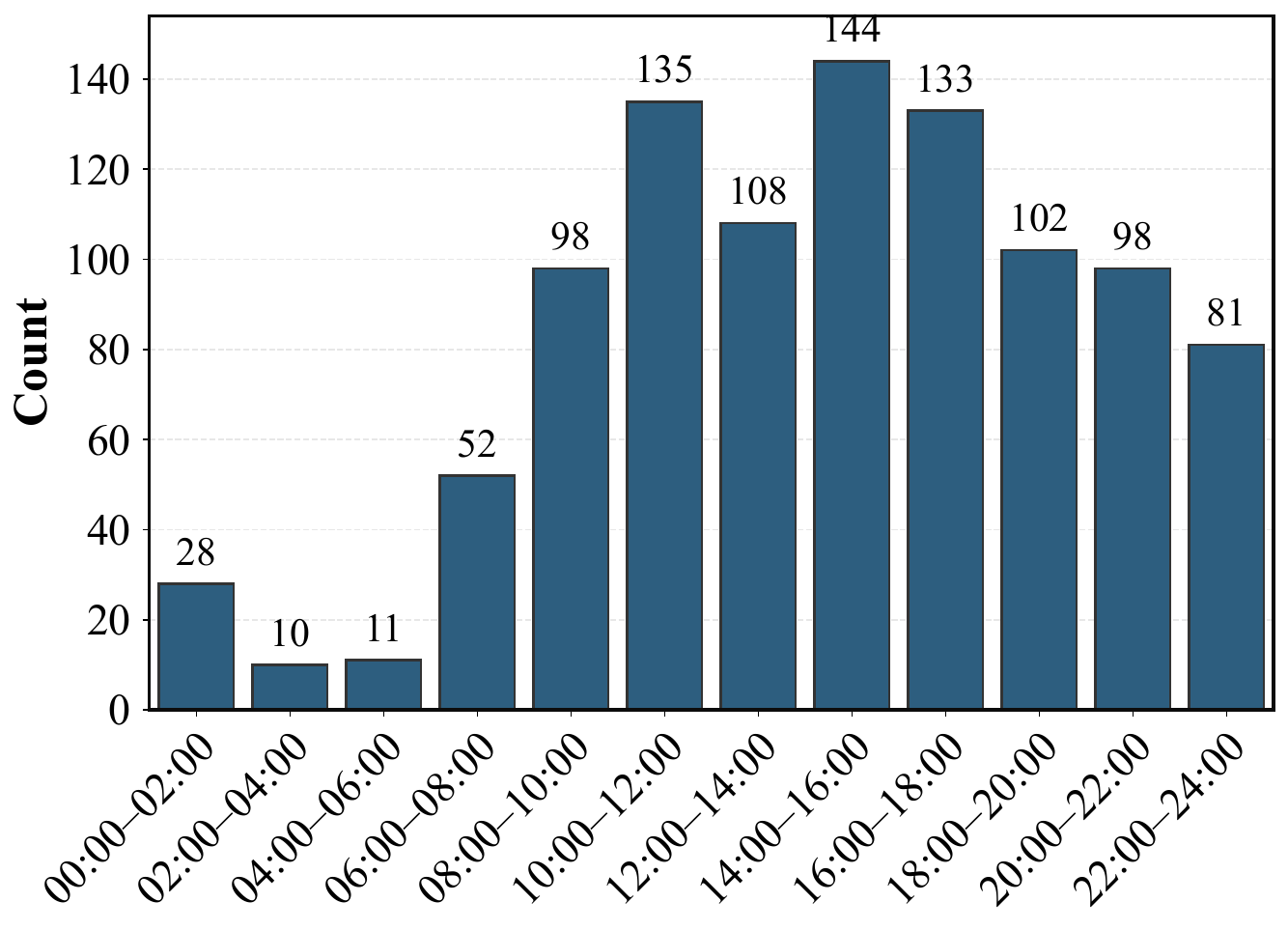}
    \caption{Query distribution by time windows throughout the day.}
    \label{fig:time-distribution}
\end{figure}

From the perspective of weekly distribution, 72.20\% of the queries were collected on weekdays and 27.80\% on weekends. Figure~\ref{fig:time-distribution} further shows the distribution across different time windows throughout the day. The data covers nearly all time periods, with relatively high frequencies observed during 10:00--12:00 (13.50\%), 14:00--16:00 (14.40\%), and 16:00--18:00 (13.30\%). Queries are also present during late-night and early-morning periods, such as 00:00--02:00 (2.80\%) and 04:00--06:00 (1.10\%). Overall, these results demonstrate broad temporal coverage and suggest that the dataset captures user information needs arising at different times of day.

\subsubsection{User Group Distribution}

\begin{figure}[t]
    \centering
    \includegraphics[width=1.0\columnwidth]{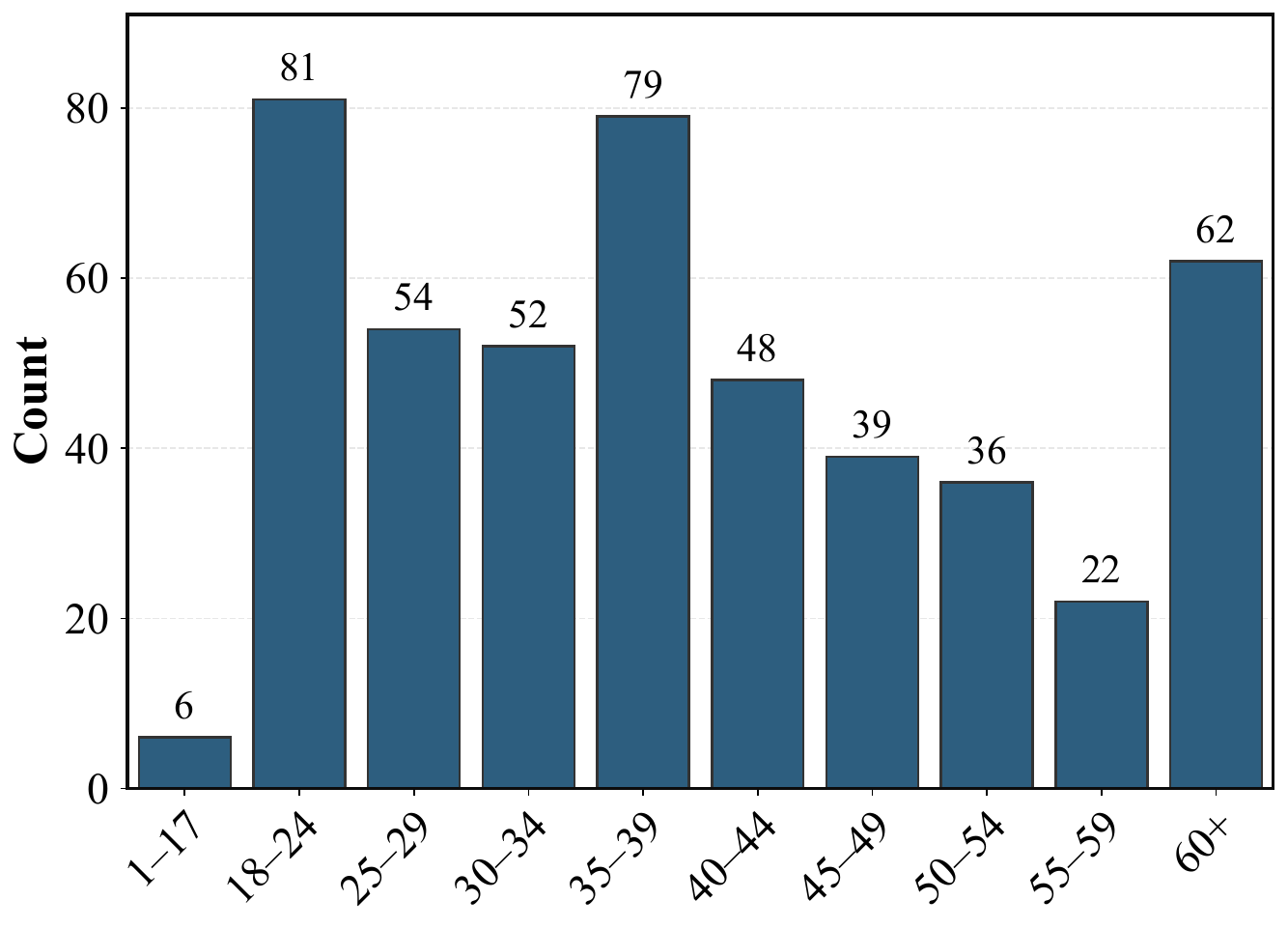}
    \caption{User group distribution by age ranges.}
    \label{fig:age-distribution}
\end{figure}

For the 500 multi-turn instances, each sample is associated with an anonymized user profile. When age information is available, we summarize the age distribution of users in Figure~\ref{fig:age-distribution}. Since some profiles do not contain age annotations, the total number of age-labeled users is smaller than 500.

As shown in Figure~\ref{fig:age-distribution}, the dataset covers nearly all age groups. The largest groups are users aged 18--24 (16.91\%), 35--39 (16.49\%), and 60+ (12.94\%), while other age ranges, including 25--29, 30--34, 40--44, 45--49, 50--54, and 55--59, are also well represented. This distribution indicates that the dataset is relatively balanced across a broad age spectrum.

Due to anonymization, our data does not retain certain sensitive demographic attributes, such as occupation, and thus we are unable to report corresponding statistics. However, the profile fields may still include coarse-grained background information synthesized by an LLM, such as family status or primary activity areas. Moreover, manual inspection of the source logs shows that free-text occupation mentions, when available, span a diverse set of user types, including students, teachers, company employees, university researchers, and freelancers. Combined with the broad age distribution, these observations suggest that the dataset captures diverse user groups.

\subsubsection{Preference Pattern Distribution}
\label{app:user_profile clarification}
\begin{figure}[t]
    \centering
    \includegraphics[width=1.0\columnwidth]{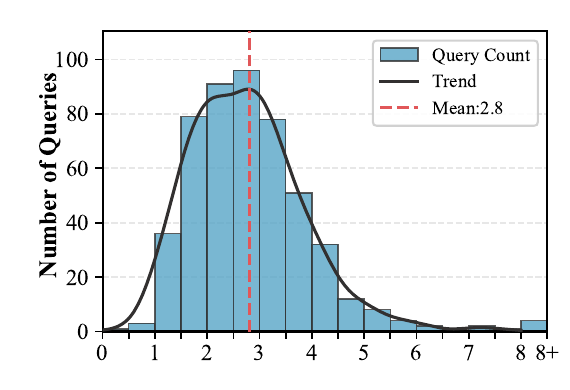}
    \caption{Distribution of reasoning steps in the Single-Turn subset (averaged over four models). One step corresponds to a tool call followed by its returned result.}
    \label{fig:step-distribution}
    \vspace{-10pt}
\end{figure}

\begin{figure}[t]
    \centering
    \includegraphics[width=1.0\columnwidth]{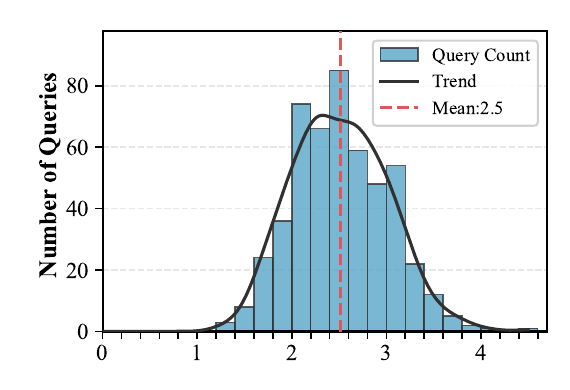}
    \caption{Distribution of interaction turns in the Multi-Turn subset (averaged over four models). One turn corresponds to one user reply.}
    \label{fig:turn-distribution}
    \vspace{-10pt}
\end{figure}

Rather than imposing a fixed taxonomy of user preferences, we derive preference signals from the implicit preference information contained in user-profile fields. This design choice better reflects realistic user modeling settings, where preferences are often expressed indirectly and embedded in profile descriptions or behavioral context.

In practice, the extracted preferences may involve a diverse set of aspects, roughly including but not limited to the following 12 categories: (1) activity regions and location tendencies, (2) mobility and commuting preferences, (3) hotel and accommodation preferences, (4) dining and food tastes, (5) shopping and consumption scenarios, (6) leisure, entertainment, and family/child-related activities, (7) sports, fitness, and health concerns, (8) travel and attraction interests, (9) car- and vehicle-related services, (10) local services and personal care, (11) cultural activities and exhibition/conference scenarios, and (12) real-estate and living-environment preferences.

Appendix Figure~\ref{app:data_example} provides a concrete data example that includes the user\_profile field, which can be referred to for its structure.

\subsubsection{Reasoning Step Distribution}
Figure~\ref{fig:step-distribution} presents the distribution of reasoning steps for the Single-Turn subset, where one step is defined as \textit{issuing a tool call and receiving its output}. Following the same setup as the tool-call distribution in Figure~\ref{fig:tool_call_distribution}, we report averages over trajectories produced by the four evaluated models. Overall, the step distribution is broadly consistent with the tool-call distribution. The mean number of steps is slightly lower than the mean number of tool calls, because a model may invoke multiple tools within a single reasoning step.

\begin{table*}
\centering
\small
\setlength{\tabcolsep}{6pt}
\renewcommand{\arraystretch}{1.15}
\caption{Overview of the tool library used in our benchmark sandbox, grouped by domain.}
\label{tab:tool_library}
\begin{tabular}{lll}
\toprule
\textbf{Domain} & \textbf{Tool name} & \textbf{Function} \\
\midrule
\multirow{5}{*}{Maps \& routing}
& map\_search\_places & Retrieve POIs by keyword, coordinates, or area \\
& map\_compute\_routes & Plan routes across modes from origin to destination \\
& map\_search\_along\_route & Search POIs within a corridor along a given route \\
& map\_search\_central\_places & Find centrally located meeting points for multiple origins \\
& map\_search\_ranking\_list & Access curated POI ranking lists \\
\midrule
\multirow{2}{*}{Transportation}
& travel\_search\_flights & Search China domestic flights with flexible dates \\
& travel\_search\_trains & Search China train and high-speed rail trips with flexible dates \\
\midrule
\multirow{2}{*}{Weather}
& weather\_current\_conditions & Return current weather, temperature, and wind conditions. \\
& weather\_forecast\_days & Return multi-day weather forecasts \\
\midrule
General information
& web\_search & Perform open-domain web search \\
\bottomrule
\end{tabular}
\end{table*}

\subsubsection{Interaction Turn Distribution}
Figure~\ref{fig:turn-distribution} shows the distribution of interaction turns for the Multi-Turn subset, also averaged over the four models. The mean number of user--agent interaction turns is 2.5. Very short and very long conversations form long tails, which aligns with real-world interaction patterns.

\section{Sandbox Tools}
\label{app:sec:sandbox-tools}

\subsection{Details of the Tool Library}

Table~\ref{tab:tool_library} lists the 10 \textbf{real, production-grade} tools used in our sandbox environment. They cover four domains—\textit{map \& navigation}, \textit{travel \& transportation}, \textit{weather}, and \textit{general information}—and can \textbf{consistently return realistic tool outputs}. This tool suite provides strong support for simulating how users and deployed agents solve travel-planning problems in practice.

We describe each tool below.

\noindent\textbf{Map \& Navigation Tools.}
\begin{enumerate}[leftmargin=*, itemsep=1pt, topsep=2pt, parsep=0pt]
    \item \textbf{map\_search\_places}: A large-coverage POI retrieval tool that supports \textbf{nationwide search in China}. It can search a wide range of place types (e.g., restaurants, hotels, attractions, shopping malls, hospitals, universities, airports, and railway stations) using keywords, categories, or addresses. It supports nearby search with a configurable radius, administrative region constraints, and multiple ranking strategies (e.g., distance, rating, and price). The tool returns rich structured metadata for each result, including \textbf{latitude/longitude}, address, \textbf{opening hours}, ratings, pricing signals, and user reviews.
    
    \item \textbf{map\_compute\_routes}: A routing tool that computes routes between an origin and a destination, each of which can be specified by \textbf{free-form addresses or explicit coordinates}. It supports six transportation modes: driving, walking, cycling, public transit, motorcycle, and truck. The tool provides route summaries and step-level navigation instructions, and supports practical constraints and preferences (e.g., avoid toll roads, prefer highways), with traffic-aware estimates when available.
    
    \item \textbf{map\_search\_along\_route}: Searches for POIs along a planned route within a user-specified corridor. This is useful for needs such as ``find a coffee shop that is close to my route'' or ``find a restroom near the highway on the way.'' The tool first plans a base route and then returns candidate POIs that lie within the buffer region, together with detailed POI metadata.
    
    \item \textbf{map\_search\_central\_places}: Recommends convenient meeting locations for multiple participants by optimizing spatial centrality. It provides three strategies: balanced (overall best trade-off), minimize maximum distance (fairness-oriented), and minimize total distance (efficiency-oriented). This supports realistic coordination scenarios (e.g., choosing a dinner place for people coming from different districts).
    
    \item \textbf{map\_search\_ranking\_list}: Retrieves curated local ranking lists for a given region and category (e.g., top-rated local eateries or popular attractions). It returns ranked POIs with tags and short recommendation rationales, which is useful for recommendation-style travel planning.
\end{enumerate}

\begin{figure*}[t]
    \centering
    \includegraphics[width=1.0\textwidth]{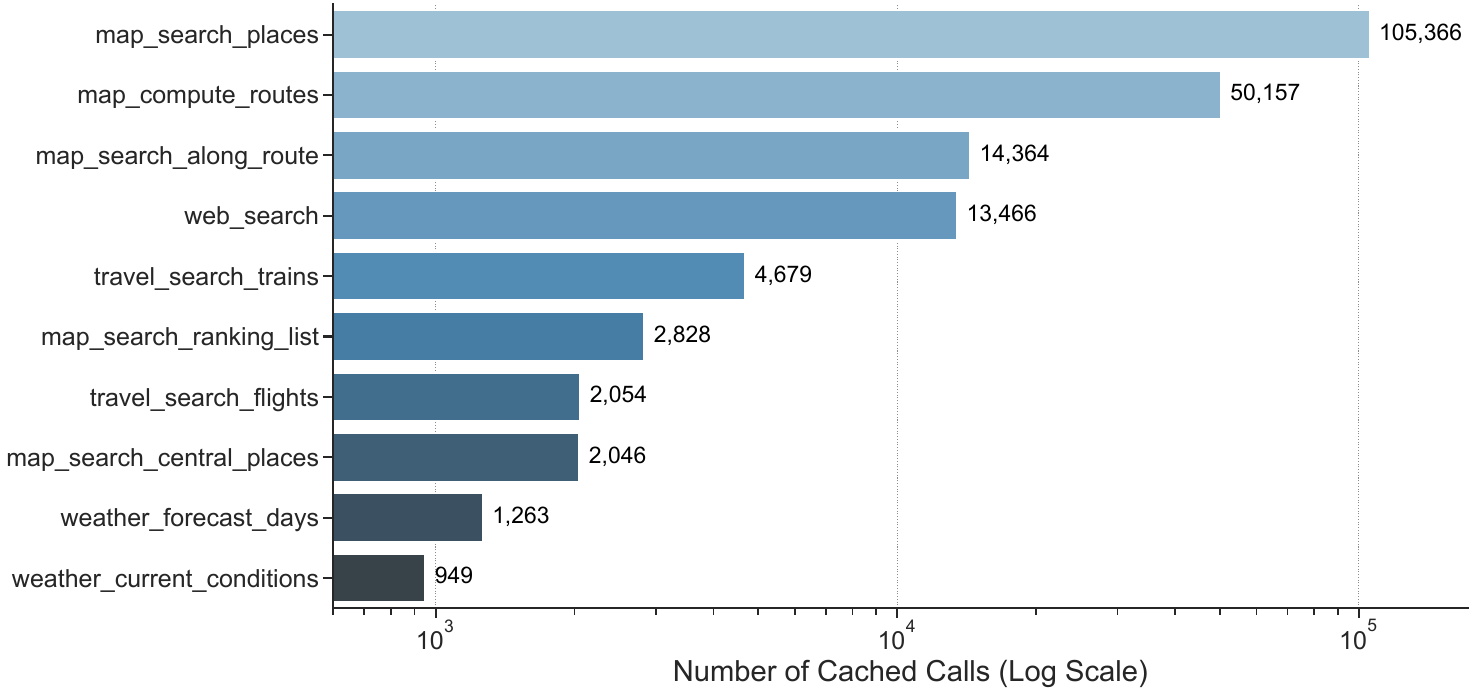}
    \caption{Distribution of cached tool calls in our sandbox environment.}
    \label{fig:cached_tool_distribution}
    \vspace{-5pt}
\end{figure*}

\noindent\textbf{Travel \& Transportation Tools.}
\begin{enumerate}[leftmargin=*, itemsep=1pt, topsep=2pt, parsep=0pt]
    \item \textbf{travel\_search\_flights}: Searches domestic flight options between two cities. It supports multi-day queries to compare schedules and prices across adjacent dates. The tool returns structured flight information such as flight number, airline, departure/arrival time, aircraft type, and price ranges.
    
    \item \textbf{travel\_search\_trains}: Queries train and high-speed rail schedules between cities, also supporting multi-day comparisons. It returns train number, departure/arrival stations, and time, travel duration, and ticket prices.
\end{enumerate}

\noindent\textbf{Weather Tools.}

\begin{enumerate}[leftmargin=*, itemsep=1pt, topsep=2pt, parsep=0pt]
    \item \textbf{weather\_current\_conditions}: Retrieves real-time weather conditions for a specified location, including temperature, feels-like temperature, weather phenomena, wind direction/speed.
    
    \item \textbf{weather\_forecast\_days}: Provides multi-day forecasts (up to 5 days) for a location, supporting both single-date and date-range queries.
\end{enumerate}

\noindent\textbf{Information Retrieval Tools.}
\begin{enumerate}[leftmargin=*, itemsep=1pt, topsep=2pt, parsep=0pt]
    \item \textbf{web\_search}: Performs open-domain web search for information beyond the scope of spatio-temporal tools, such as general facts, recent news, local regulations, and travel policies.
\end{enumerate}

\subsection{Tool-Cache Distribution}

Figure~\ref{fig:cached_tool_distribution} shows the distribution of cached tool responses in the sandbox, built from the 1{,}100 benchmark instances. The cache is dominated by POI search and routing calls, while weather and transportation tools account for a smaller portion. We attribute this to two main reasons. First, POI search and routing tools have more \textbf{parameter-sensitive} interfaces (e.g., different result limits such as top-5 vs.\ top-10, different sorting strategies, or slightly different coordinate inputs), which can lead to multiple cached entries that correspond to highly overlapping underlying results. In practice, this means our cache already covers most information needed by typical queries, which motivates our cache-miss handling strategy that \textbf{simulates tool outputs via ICL} when an exact match is unavailable. 

Second, this skew also likely reflects real usage patterns in travel planning: users most frequently require place lookup and navigation, whereas weather checks and ticket queries are relatively less frequent.

\section{Additional Experiments}
\label{sec:additional_experiments}
To further strengthen the empirical analysis of TravelBench, we conduct additional experiments from two perspectives. First, we perform a controlled comparison between two standard tool-use frameworks, \texttt{vLLM} and \texttt{Qwen-Agent}, to examine whether the main results for Qwen models are substantially affected by tool-call formatting or framework-specific implementation details. Second, we extend the benchmark to several additional open-source model families, in order to provide a broader view of open-source performance and to better understand how tool-calling support influences end-to-end evaluation.

\subsection{Controlling for Tool-Use Frameworks}

A potential concern in tool-augmented evaluation is that end-to-end performance may be influenced not only by model capability, but also by failures in tool-call formatting or parsing. In our main experiments, this issue is already mitigated for Qwen models. Specifically, following common practice for Qwen tool use, we deploy the models with \texttt{vLLM} and use its native tool-calling support. In addition, our evaluation framework returns execution errors to the model and allows retries, which further reduces the impact of occasional malformed outputs.

In our experiments, we do not observe tool-call parsing failures for the evaluated Qwen models. Therefore, the tool-call error rate reported in the main paper refers to \emph{schema-level} errors, i.e., argument or value mismatches with respect to the tool specification, rather than JSON parsing failures or tool-call extraction failures.

To further control for possible framework effects, we compare \texttt{vLLM} with \texttt{Qwen-Agent}, a standard agent framework commonly used for Qwen models. Table~\ref{tab:framework-control} shows that the results under the two frameworks are highly consistent. \texttt{Qwen3-30B-A3B-Instruct} obtains 54.62 with \texttt{vLLM} and 55.93 with \texttt{Qwen-Agent}, while \texttt{Qwen3-30B-A3B-Think} obtains 61.79 and 61.47, respectively. These results indicate that, for models with native tool-calling support, our \texttt{vLLM}-based evaluation already largely removes tool-formatting as a confounding factor.

\begin{table}[t]
\centering
\small
\begin{tabular}{lcc}
\toprule
Model & vLLM & Qwen-Agent \\
\midrule
Qwen3-30B-A3B-It & 54.62 & 55.93 \\
Qwen3-30B-A3B-Th & 61.79 & 61.47 \\
\bottomrule
\end{tabular}
\caption{Framework control experiment on Qwen models. Results are highly consistent between \texttt{vLLM} and \texttt{Qwen-Agent}.}
\label{tab:framework-control}
\vspace{-10pt}
\end{table}

\subsection{Broader Open-Source Baselines}

We further extend our evaluation to three additional open-source model families: \textbf{Ministral3}, \textbf{Gemma3}, and \textbf{Llama 3.x}. All models are evaluated under the same protocol as in the main experiments. In particular, all models are deployed with \texttt{vLLM}, using the recommended \texttt{chat\_templates} and tool-parsing settings, to ensure consistency across model families.

Table~\ref{tab:additional-models} presents the results on TravelBench. Among the newly added models, the \texttt{Ministral3} family achieves the strongest overall performance. In particular, \texttt{Ministral3-14B-Reasoning} reaches an overall score of 47.35, outperforming \texttt{Ministral3-14B-It} (40.47) and showing relatively balanced performance across Multi-turn, Single-turn, and Unsolvable subsets.

The \texttt{Gemma3} models obtain lower end-to-end scores overall, although the larger \texttt{Gemma3-27B-It} consistently outperforms \texttt{Gemma3-12B-It}. The \texttt{Llama 3.x} models exhibit a different pattern: while \texttt{Llama3.2-3B-It} and \texttt{Llama3.3-70B-It} achieve relatively high accuracy on the Unsolvable subset, their performance on Multi-turn and Single-turn tasks remains limited. We were also unable to obtain stable usable results for \texttt{Llama3.1-8B-It} under our evaluation setup.

\begin{table*}[t]
\small
\centering
\setlength{\tabcolsep}{5pt}
\renewcommand{\arraystretch}{1}
\caption{Additional open-source model results on TravelBench. \textit{Error} denotes the schema-level tool-call error rate, and \textit{Pen.} denotes the penalized score.}
\label{tab:additional-models}
\begin{tabularx}{0.9\textwidth}{l *{6}{Y}}
\toprule
\multirow{2}{*}{\textbf{Model}} &
\multicolumn{2}{c}{\textbf{Multi-turn}} &
\multicolumn{2}{c}{\textbf{Single-turn}} &
\multicolumn{1}{c}{\textbf{Unsolvable}} &
\multicolumn{1}{c}{\textbf{Overall}} \\
\cmidrule(lr){2-3}\cmidrule(lr){4-5}\cmidrule(lr){6-6}\cmidrule(lr){7-7}
& \textbf{Error} & \textbf{Pen.}
& \textbf{Error} & \textbf{Pen.}
& \textbf{Acc.} & \textbf{Score} \\
\midrule
Ministral3-14B-It & 7.4 & 33.74 & 3.2 & 40.34 & 47.33 & 40.47 \\
Ministral3-14B-Th & 6.4 & 35.93 & 3.0 & 41.44 & 65.67 & 47.35 \\
Gemma3-12B-It & 9.8 & 12.51 & 5.8 & 18.61 & 48.00 & 26.37 \\
Gemma3-27B-It & 7.4 & 18.24 & 5.3 & 25.88 & 54.66 & 32.93 \\
Llama3.1-8B-It & -- & -- & -- & -- & -- & -- \\
Llama3.2-3B-It & 36.0 & 4.39 & 45.4 & 5.79 & 90.33 & 33.50 \\
Llama3.3-70B-It & 33.3 & 14.93 & 49.2 & 22.18 & 84.00 & 40.37 \\
\bottomrule
\end{tabularx}
\vspace{-5pt}
\end{table*}

\subsection{Impact of Native Tool-Calling Support}

A key finding from the broader evaluation is that native tool-calling support substantially affects how benchmark performance should be interpreted. For model families with native tool-calling support, such as \texttt{Qwen3} and \texttt{Ministral3}, tool-call formatting is generally correct under the \texttt{vLLM} setup. In these cases, observed failures mainly arise from reasoning, planning, tool selection, or schema-level argument errors. Therefore, the resulting benchmark scores more directly reflect the model's underlying capability for tool-augmented problem solving.

In contrast, for model families without native tool-calling support, such as \texttt{Gemma3} and \texttt{Llama 3.x}, tool use must be induced entirely through prompting. Even when using the recommended \texttt{vLLM} templates, we observe frequent formatting issues that prevent tool calls from being successfully parsed or executed. Typical problems include malformed call structures, extra Markdown wrappers, and serialization inconsistencies. Such failures interrupt the execution chain and may lead to retries, stalled trajectories, or task failure. As a result, end-to-end scores for these models reflect a mixture of core task-solving ability and adherence to tool-call formatting requirements.

This distinction is important for interpreting TravelBench results. For native tool-calling models, the benchmark more cleanly measures planning and tool-use ability. For non-native tool-calling models, the evaluation also exposes a practically important but separate challenge: reliably producing executable tool calls under long contexts and a relatively large tool set.

\subsection{Schema-Level Tool Errors}

To further distinguish schema-level tool-use errors from pure formatting failures, we compute tool-call error rates for all tool calls that were successfully parsed and executed. Following the definition in the main paper, a tool call is considered erroneous if its arguments or values do not match the tool schema.

Table~\ref{tab:additional-models} also reports the corresponding error rates. The \texttt{Ministral3} models show relatively low schema-level tool-call error rates, with 6.4\%--7.4\% on Multi-turn and 3.0\%--3.2\% on Single-turn. The \texttt{Gemma3} models perform somewhat worse, but remain in a moderate range. By contrast, the \texttt{Llama} models exhibit substantially higher error rates, reaching 33.3\%--36.0\% on Multi-turn and 45.4\%--49.2\% on Single-turn. Typical failure cases include providing place names where latitude/longitude values are required, or generating arguments that violate schema constraints.

These results suggest that some model families face two distinct bottlenecks: first, reliably generating parseable tool calls, and second, providing schema-correct arguments after parsing succeeds. In particular, the \texttt{Llama} models appear to struggle with robust tool use in the presence of long contexts and a large tool inventory.

\subsection{Discussion}

One additional observation concerns the Unsolvable subset. We find that \texttt{Llama3.2-3B-It} and \texttt{Llama3.3-70B-It} achieve relatively high accuracy on these instances, suggesting that they are comparatively conservative when the requested operation cannot be executed. In this setting, they are less likely to fabricate tool outputs, which leads to stronger boundary checking behavior.

Overall, the additional experiments support two conclusions. First, the main findings for Qwen models are not artifacts of a specific tool-use framework: under both \texttt{vLLM} and \texttt{Qwen-Agent}, the results remain highly consistent. Second, broadening the evaluation to more open-source model families reveals that TravelBench captures not only reasoning and planning ability, but also the practical robustness required for executable tool use. This distinction is especially important when comparing models with native tool-calling support against those that rely entirely on prompt-induced tool use.

\section{Human Validation of Dataset Quality and Judge Reliability}
\label{appendix:human_validation}

Because our pipeline uses LLMs in several stages, it is important to verify both the quality of the released dataset and the reliability of the LLM-based evaluation framework. To this end, we conducted additional human validation studies covering key dataset construction steps and the judge system used in our experiments. Across all components, prompts were not used in a one-shot manner; instead, they were \textbf{iteratively refined through multiple rounds of manual inspection and revision on small batches of data} to improve consistency with human judgment.

\subsection{Validation of Dataset Quality}
\label{appendix:dataset_quality}

We validated the main dataset processing steps that may raise reliability concerns, including user-profile de-identification, feasibility annotation, and the single-turn/multi-turn split.

\subsubsection{User-profile de-identification}
\label{appendix:deidentification}

We fully recognize the sensitivity of user-profile information. For the de-identified user profiles generated by the LLM in our multi-turn subset (500 examples), we conducted \textbf{full manual verification}. Any remaining traces of potentially sensitive information were further \textbf{manually removed}. As a result, the released user-profile fields contain only coarse background and preference information and exclude details that could identify a specific individual. In addition, the dataset will undergo an internal compliance review before release.

\subsubsection{Intent classification}
\label{appendix:intent_classification}

The intent categories reported in the appendix follow our established internal taxonomy and labeling guidelines. Importantly, the \texttt{primary\_intent} field is \textbf{directly inherited from the originally collected queries} rather than generated by the processing pipeline introduced in this work. We report the intent distribution mainly to describe the overall dataset composition, which also roughly reflects real-world query distributions. This field is not used in downstream subtask evaluation.

\subsubsection{Feasibility annotation}
\label{appendix:feasibility_validation}

To reduce potential bias from any individual model, we first annotated query feasibility using three different models and retained only the intersection of their decisions. We then manually reviewed the resulting examples and removed ambiguous cases, such as route-planning requests with many intermediate stops. This process produced \textbf{100} test queries for the feasibility task. More clarification is provided in Appendix~\ref{sec:appendix_unsolvable_definition}.

To further assess label quality, we conducted a dedicated human validation study. Three independent annotators labeled each query using a binary label: \textbf{1 = unsolvable} and \textbf{0 = solvable}, following the same criteria used in our prompt. The final human label was determined by majority vote. We report agreement with the released label, together with the \textbf{Clopper--Pearson exact 95\% binomial confidence interval}.

The released labels agree with the majority human judgment on \textbf{97 out of 100} queries (97.0\%, 95\% CI [0.9148, 0.9938]), indicating high reliability. Table~\ref{tab:feasibility_vote_distribution} also shows the vote distribution, which provides an additional view of inter-annotator consistency.

\begin{table}[t]
\centering
\small
\begin{tabular}{lcccc}
\toprule
\makecell[l]{Votes for ``unsolvable''\\(out of 3)} & 3/3 & 2/3 & 1/3 & 0/3 \\
\midrule
Count & 87 & 10 & 3 & 0 \\
\bottomrule
\end{tabular}
\caption{Vote distribution in the human validation of feasibility annotation.}
\label{tab:feasibility_vote_distribution}
\vspace{-10pt}
\end{table}

\subsubsection{Validation of the single-turn and multi-turn split}
\label{appendix:turn_split_validation}

Although the turn-type split was already supported by execution-based checks, multi-model sampling, and manual spot checks, we further conducted human validation to quantify its quality.

For the single-turn setting, we randomly sampled \textbf{100 queries}. Three annotators labeled each query as \textbf{1 = single-turn} if the intent was clear and complete, or \textbf{0 = not single-turn} if clarification was needed. Final labels were determined by majority vote. We applied the same procedure to the multi-turn setting.

For the single-turn subset, \textbf{97/100} examples were confirmed by majority human judgment (97.0\%, 95\% CI [0.9148, 0.9938]). The vote distribution is shown in Table~\ref{tab:single_turn_vote_distribution}.

\begin{table}[t]
\centering
\small
\begin{tabular}{lcccc}
\toprule
\makecell[l]{Votes for ``single-turn''\\ (out of 3)} & 3/3 & 2/3 & 1/3 & 0/3 \\
\midrule
Count & 91 & 6 & 2 & 1 \\
\bottomrule
\end{tabular}
\caption{Vote distribution in the human validation of the single-turn split.}
\label{tab:single_turn_vote_distribution}
\vspace{-5pt}
\end{table}

For the multi-turn subset, \textbf{95/100} examples were confirmed by majority human judgment (95.0\%, 95\% CI [0.8872, 0.9836]). The vote distribution is shown in Table~\ref{tab:multi_turn_vote_distribution}.

\begin{table}[t]
\centering
\small
\begin{tabular}{lcccc}
\toprule
\makecell[l]{Votes for ``multi-turn'' \\(out of 3)} & 3/3 & 2/3 & 1/3 & 0/3 \\
\midrule
Count & 91 & 4 & 4 & 1 \\
\bottomrule
\end{tabular}
\caption{Vote distribution in the human validation of the multi-turn split.}
\label{tab:multi_turn_vote_distribution}
\vspace{-10pt}
\end{table}

Overall, these results suggest that although LLMs are used to scale the pipeline, the core dataset properties are supported by systematic human verification and achieve high agreement rates.

\subsection{Validation of LLM-based Judge Reliability}
\label{appendix:judge_validation}

In addition to dataset quality, we also examined the reliability of our LLM-based evaluation framework. As with the dataset construction prompts, the evaluation prompts were iteratively refined through repeated manual inspection. We also instruct the judge to \textbf{reason before assigning a score}, which we found helpful for improving scoring accuracy. While the stability analysis in Section~\ref{sec:stability-of-benchmark-components} demonstrates that the framework is \textbf{consistent across repeated runs}, consistency alone does not guarantee correctness. We therefore complement it with direct human validation.

\subsubsection{Agreement between the raw judge and human annotators}
\label{appendix:raw_judge_validation}

We evaluated judge reliability in the multi-turn setting, where all four evaluation dimensions are involved. Specifically, we randomly sampled \textbf{100 trajectories} generated by GPT-5.1 and asked \textbf{three independent annotators} to score each trajectory using the same four-dimensional rubric with discrete scores from 1 to 5.

We report mean absolute error (MAE) in two forms:
(1) \textbf{MAE(LLM vs.\ human)} measures the MAE between the raw judge's score and each annotator's score, averaged across annotators;
(2) \textbf{MAE(human vs.\ human)} uses the median of the three human annotations as a reference, computes each annotator's MAE to this median, and then averages the results. Because trajectory evaluation can be inherently subjective, this human-human deviation serves as an approximate upper bound on the level of agreement one may expect.

\begin{table}[t]
\centering
\small
\begin{tabular}{lcc}
\toprule
 & LLM--Hum. & Hum.--Hum. \\
\midrule
Avg. (4 dims.) & 0.52 & 0.48 \\
\bottomrule
\end{tabular}
\caption{Agreement between the raw judge and human annotators in the multi-turn setting. 
\textbf{LLM--Hum.} denotes MAE between the raw judge and human annotators, averaged across annotators. 
\textbf{Hum.--Hum.} denotes average MAE between each annotator and the median of the three human ratings. 
\textbf{Avg. (4 dims.)} indicates the average over the four evaluation dimensions.}
\label{tab:judge_human_agreement}
\end{table}

As shown in Table~\ref{tab:judge_human_agreement}, the gap between the two values is small, suggesting that the raw judge is \textbf{well aligned with human judgment} and performs at a level close to inter-annotator agreement.

\subsubsection{Meta-judge calibration analysis}
\label{appendix:meta_judge_validation}

We introduced a meta-judge for two reasons. First, in manual spot checks, we observed a small number of cases in which the raw judge appeared \textbf{overly generous}, for example by missing certain hallucination errors. Second, following common practices in reward modeling, we use a second-stage judge to reduce the risk of \textbf{reward hacking} and improve evaluation robustness.

To analyze its behavior, we applied the meta-judge to \textbf{1,500 multi-turn trajectories} generated by GPT-5.1. The score distribution is shown in Table~\ref{tab:meta_judge_distribution}, where a score of \textbf{5 indicates no penalty}, meaning that no further calibration is needed.

\begin{table}[t]
\centering
\small
\begin{tabular}{lccccc}
\toprule
Score & 1 & 2 & 3 & 4 & 5 \\
\midrule
Count & 0 & 6 & 12 & 21 & 1450 \\
\bottomrule
\end{tabular}
\caption{Score distribution of the meta-judge on 1,500 multi-turn trajectories.}
\label{tab:meta_judge_distribution}
\vspace{-10pt}
\end{table}

Only \textbf{50 out of 1,500 cases (3.3\%)} received any penalty, which indicates that the raw judge is generally reliable. We further manually inspected these 50 cases and found that the meta-judge successfully identified errors missed by the raw judge, most of which were \textbf{hallucination-related issues}. More severe hallucinations tended to receive lower meta-judge scores, suggesting that the meta-judge plays an effective calibration role rather than broadly altering the original evaluations.

\subsection{Summary}
\label{appendix:validation_summary}

Taken together, these human validation results support both the quality of the dataset and the reliability of the LLM-based evaluation framework. On the dataset side, key properties such as de-identification, feasibility labels, and turn-type splits are all backed by manual checks and high human agreement. On the evaluation side, the raw judge shows strong alignment with human ratings, while the meta-judge provides a targeted correction mechanism for the small number of cases in which the raw judge may be overly permissive.

\section{Additional Clarifications}
\label{sec:appendix_clarifications}

\begin{table*}[t]
\centering
\small
\begin{tabularx}{0.95\textwidth}{p{3.0cm}X}
\toprule
\textbf{Dimension} & \textbf{What it evaluates} \\
\midrule
\textit{reasoning\_planning} &
Whether the model correctly understands the user’s goal and its relation to the available tools; whether the tool-use trajectory is clear and reasonable; whether tool parameters are specified correctly; and whether the model avoids redundant or unnecessary tool calls. \\
\midrule
\textit{summarization\_extraction} &
Whether the model accurately extracts the most relevant information from tool outputs and dialogue context, including key constraints and returned arguments, while avoiding hallucinated facts or fabricated numbers. \\
\midrule
\textit{presentation} &
Whether the final response presents information relevant to the user’s request clearly, accurately, and concisely, and whether it provides appropriate user-facing feedback, such as indicating missing details or suggesting useful follow-up actions. \\
\midrule
\textit{user\_interaction} &
Whether, when information is insufficient or ambiguous, or when tool execution fails, the model asks necessary and high-value follow-up questions while minimizing user burden; whether it prefers reasoning or tool use to fill gaps when possible; and whether its questions remain focused on the user’s original intent rather than drifting into unnecessary preference probing. \\
\bottomrule
\end{tabularx}
\caption{Detailed definitions of the rubric dimensions used in LLM-as-a-judge evaluation.}
\label{tab:rubric_dimensions}
\end{table*}

This section provides additional details on the dataset construction and evaluation protocol.

\subsection{Definition of Unsolvable Queries}
\label{sec:appendix_unsolvable_definition}

In our dataset, \textit{unsolvable} does not mean ``too difficult for current LLMs.'' Instead, it refers to queries that are \textbf{inherently non-executable under our task formulation}. Specifically, a query is considered unsolvable when it falls into one of the following categories:

\begin{itemize}
    \item \textbf{Missing tools/actions:} The request requires an external capability that is not available in the provided tool set, e.g., \textit{``set a reminder for me''} when no reminder-setting tool exists.
    \item \textbf{Missing critical context:} The request cannot be executed without essential user-specific information, e.g., navigation-dependent queries such as \textit{``find something on the way''} without origin, destination, or location context.
    \item \textbf{Non-executable or non-instructional utterances:} The user input does not specify an actionable intent, such as rhetorical questions, complaints, or clearly unreasonable requests.
\end{itemize}

Some borderline cases may appear difficult rather than truly unsolvable, such as complex multi-stop route planning. We treat such examples as \textbf{ambiguous} rather than unsolvable, since a human might still complete them under additional assumptions or through iterative planning. During dataset curation, we therefore manually removed these ambiguous cases.

To further verify that these labels reflect \textbf{definition-based non-executability} rather than model limitations, we conducted a human validation study on the 100-query unsolvable test set using three annotators. The resulting agreement with the released labels is \textbf{97\%}, supporting that the unsolvable subset is aligned with human judgment.

\subsection{Tool-Call Count as a Difficulty Proxy}
\label{sec:appendix_toolcall_difficulty}

We use the number of tool calls as an approximate indicator of task difficulty. Intuitively, more difficult queries often require more tool calls either because they involve retrieving and integrating \textbf{multiple pieces of information}, or because they require more \textbf{complex planning across tools}. For example, a constrained query such as finding a suitable place near a destination may require several intermediate retrieval and filtering steps.

To empirically validate this proxy, we ranked instances by the number of tool calls and evaluated GPT-5.1 on the Top-50 and Top-100 subsets with the highest tool-call counts. The results are shown in Table~\ref{tab:toolcall_difficulty}. In both single-turn and multi-turn settings, the performance consistently follows \textbf{All $>$ Top100 $>$ Top50}, suggesting that instances requiring more tool calls are more challenging on average.

\begin{table}[h]
\centering
\small
\begin{tabular}{lcc}
\toprule
\textbf{Subset} & \textbf{Single-turn} & \textbf{Multi-turn} \\
\midrule
Top50  & 64.88 & 60.31 \\
Top100 & 67.62 & 63.21 \\
All    & 74.25 & 70.13 \\
\bottomrule
\end{tabular}
\caption{GPT-5.1 performance on subsets ranked by tool-call count. Lower scores on Top50/Top100 indicate that instances requiring more tool calls are more difficult on average.}
\label{tab:toolcall_difficulty}
\end{table}

Especially in the multi-turn setting, the number and quality of user interactions may also affect difficulty. We therefore view tool-call count as a \textbf{coarse but practical proxy}, rather than a complete characterization of task complexity.

\subsection{Detailed Definitions of Evaluation Dimensions}
\label{sec:appendix_rubric_details}

For single-turn, we evaluate trajectories along three dimensions: \textit{reasoning\_planning}, \textit{summarization\_extraction}, and \textit{presentation}. For multi-turn, we additionally include \textit{user\_interaction}. The evaluation prompts for single-turn, multi-turn, and meta-judge scoring are shown in Fig~\ref{app:prompt:llm-judge-single-turn}, Fig~\ref{app:prompt:llm-judge-multi-turn}, and Fig~\ref{app:prompt:meta-judge}, respectively. These prompts were iteratively refined through multiple rounds of manual review on a small subset to improve alignment with human judgment.

Table~\ref{tab:rubric_dimensions} summarizes the intended role of each evaluation dimension.

\subsection{Interpretation of the Calibration and Penalty Terms}
\label{sec:appendix_metric_clarification}

We further clarify the intended role of the calibration and penalty terms used in the final trajectory score. A potential source of confusion is that the score combines two multiplicative factors, $w_{meta}$ and $w_{tool}$, which operate on different aspects of evaluation.

Our design intention is to treat these two factors as serving \textbf{distinct functions}:

\begin{itemize}
    \item \textbf{Meta-evaluation calibration ($w_{meta}$):} This term is introduced to correct a small number of \textbf{over-estimated raw scores} produced by the initial judge. In these cases, the raw evaluator may miss issues such as hallucinated content or clear rubric violations, and therefore assign a score that is too high. The calibrated quantity $S_{raw} \cdot w_{meta}$ is intended to be a more faithful estimate of the actual trajectory quality than $S_{raw}$ alone.
    
    \item \textbf{Tool-use penalty ($w_{tool}$):} This term is used to penalize \textbf{tool-related failures or inefficiencies}, such as incorrect tool invocation, execution errors, or unnecessary tool calls. Its role is not to recalibrate the judge, but to reflect practical deficiencies in how the planner uses external tools.
\end{itemize}

Under this interpretation, a lower $w_{meta}$ should not be viewed as reflecting a generic ``weakness of the judge.'' Rather, it is applied when the meta-evaluation identifies a \textbf{trajectory-specific problem} that the raw judge likely failed to capture. In such cases, lowering the score is appropriate because the underlying issue is a property of the planner output itself, not merely of the evaluator.

Importantly, this calibration is applied only to a \textbf{small fraction of examples} (approximately \textbf{3\%} in our analysis). Therefore, $w_{meta}$ is not intended as a broad score reduction mechanism, but as a targeted correction for rare likely over-scored trajectories.

Overall, the combined score should be interpreted as follows:

\begin{itemize}
    \item first, $w_{meta}$ adjusts the raw quality estimate when there is evidence that the initial judge overestimated the trajectory; and
    \item then, $w_{tool}$ applies an additional penalty for tool-use failures that directly affect execution quality and practical usability.
\end{itemize}

In this sense, the two factors are complementary rather than redundant: one corrects \textbf{evaluation overestimation}, while the other reflects \textbf{execution-related errors}.

\subsection{Definition of Implicit Preferences in Multi-turn Interaction}
\label{sec:appendix_implicit_preferences}

We also clarify what we mean by \textit{implicit preferences} and how they are used in the task formulation.

In our setting, \textbf{implicit preferences are preferences that are known to the user but not directly observable to the assistant unless they are explicitly elicited through dialogue}. The term ``implicit'' is therefore defined from the \textbf{assistant's perspective}: the information exists in the environment, but is not available in the model input by default.

\paragraph{Single-turn setting.}
In the single-turn task, the assistant responds based only on the current query and available context. The user profile is not involved, and implicit preference elicitation is therefore not part of the task design.

\paragraph{Multi-turn setting.}
Implicit preferences are relevant only in the multi-turn setting. Here, the \texttt{user\_profile} is maintained by the \textbf{user simulator} and is \textbf{not provided as part of the assistant input}. As a result, the assistant cannot directly observe user preferences at the beginning of the conversation. Instead, it must obtain them by asking appropriate follow-up questions. If the assistant does not ask, the user will not proactively reveal such information.

This design is intended to evaluate whether the assistant can identify when additional user-specific information is necessary and can elicit it efficiently through interaction, rather than assuming that all relevant preferences are already given.

\paragraph{Profile contents.}
The user profile contains two broad categories of information:

\begin{itemize}
    \item \textbf{Basic information:} examples include age, primary living area, family status, and car ownership.
    \item \textbf{Interests and preferences:} examples include shopping habits, travel-related preferences, sports interests, frequently visited places, preferred cuisine, preferred hotel type, transportation preferences, and whether the user is traveling with children.
\end{itemize}

These fields provide the latent information source from which the simulator answers the assistant's clarification questions. The task objective is not to expose the full profile, but to encourage the assistant to \textbf{progressively uncover only the preferences that are relevant to the current user request}, thereby improving task completion quality and user satisfaction.

\section{Case Studies}
\label{app:case_study}

We present several case studies sampled from Qwen-Plus to illustrate the stability of our agents' reasoning in both single-turn and multi-turn settings, as well as their ability to detect and correct errors.

Figure~\ref{demo:case_study_single_turn} shows a single-turn example where the user intent is explicit. The assistant solves the task via multiple tool calls and corrects itself when a tool returns an empty result. The judge assigns full scores (5/5) on all three dimensions.

Figure~\ref{demo:case_study_multi_turn} shows a multi-turn example where the user's request is underspecified. The assistant iteratively refines the requirements through reasoning and user interaction, provides recommendations, and confirms key details. It then completes the task based on the user's feedback. The judge assigns full scores (5/5) on all four dimensions.

Figure~\ref{demo:case_study_foshan_dali_self_drive} presents a failed multi-turn example. The user asks for a self-driving route from Foshan to Dali. Although the assistant eventually produces a travel plan through multiple reasoning steps and interactions, it does not ask for the departure time and instead assumes the trip starts on the current day. As a result, the \textit{reasoning\_planning} and \textit{user\_interaction} dimensions receive 3/5.

\section{List of Prompts}
\label{app:prompt}
This section lists all prompts used in our framework. Each prompt was iteratively refined through multiple rounds of development to ensure quality and robustness.

\begin{enumerate}[leftmargin=*, itemsep=1pt, topsep=2pt, parsep=0pt]    
    \item Figure~\ref{app:prompt:unsolvability determination} shows the prompt for \textbf{query feasibility determination}. It specifies a step-by-step analysis procedure and provides an example for each outcome, helping the model make correct feasibility judgments for complex queries.
    
    \item Figure~\ref{app:prompt:assistant-multi-turn} shows the prompt for the \textbf{multi-turn assistant}. The agent is instructed to solve the task on its own whenever possible, ask the user questions only when key information is missing, avoid requesting the user to take actions outside the dialogue, and follow tool-use rules.
    
    \item Figure~\ref{app:prompt:user-simulator} shows the prompt for the \textbf{user simulator}. It enforces that the simulator replies strictly based on the provided \texttt{user\_preference}, without inventing additional preferences, and defines clear conditions for ending the conversation.
    
    \item Figure~\ref{app:prompt:assistant-single-turn} shows the prompt for the \textbf{single-turn assistant}. The agent is instructed to solve the task without asking clarification questions and to follow tool-use rules.
    
    \item Figure~\ref{app:prompt:assistant-unsolvable} shows the prompt for handling \textbf{infeasible queries}. It is derived from the single-turn assistant prompt, with an explicit rule specifying when to output \texttt{[Unsolved]}.
    
    \item Figure~\ref{app:prompt:tool-simulator} shows the prompt for the \textbf{tool simulator}. The model is instructed to follow the provided examples and generate tool outputs that are realistic and consistent in format.
    
    \item Figure~\ref{app:prompt:llm-judge-single-turn} shows the prompt for \textbf{judging single-turn trajectories}. The judge first performs structured reasoning and then assigns comprehensive scores under three dimensions.
    
    \item Figure~\ref{app:prompt:llm-judge-multi-turn} shows the prompt for \textbf{judging multi-turn trajectories}. It extends the single-turn judging prompt by adding a user-interaction dimension, and evaluates trajectories under four dimensions with the same “reason-then-score” structure.
    
    \item Figure~\ref{app:prompt:meta-judge} shows the prompt for the \textbf{meta-judge}. It asks the model to audit an existing evaluation from multiple perspectives and correct potentially biased or low-quality judgments.
\end{enumerate}

\clearpage 
\onecolumn 

\begin{figure*}[t]
    \begin{tcolorbox}[title=Case Study: Finding the Nearest Specific Restaurant Chain, fontupper=\fontsize{11pt}{13pt}\selectfont, ]
    
    \textbf{User}: Find the nearest Xiao Dong Ge BBQ restaurant to Buyang International.
    
    \textbf{Assistant (Thought)}: I need to find the location of "Buyang International" first, then search for nearby Xiao Dong Ge BBQ branches. First, I will search for "Buyang International" in Shenyang.
    
    \textbf{Tool Call - map\_search\_places}: \\
    {\ttfamily \{ "query": "Buyang International", "region": "Shenyang", "center": "41.747178, 123.4002" \}}
    
    \textbf{Tool Result}: \\
    {\ttfamily No relevant information found. Please confirm if the input parameters are correct; if they are, there may be no data currently available.}
    
    \textbf{Assistant (Thought)}: Let me try a broader search by directly searching for "Buyang International" within the entire city of Shenyang.
    
    \textbf{Tool Call - map\_search\_places}: \\
    {\ttfamily \{ "query": "Buyang International", "region": "Shenyang" \}}
    
    \textbf{Tool Result}: \\
    {\ttfamily Found 10 locations. Key result: \par
    Item 1: [Buyang International] (Real Estate > Residential Area) \par
    Address: No. 10 Wenhua East Road, Nanta, Shenhe District, Shenyang, Liaoning \par
    Coordinates: 41.770112, 123.461602 \par
    Description: Located in the heart of Nanta Business District, adjacent to Metro Lines 3 and 10.}
    
    \textbf{Assistant (Thought)}: Now that I have found the location of Buyang International at (41.770112, 123.461602), I will search for "Xiao Dong Ge BBQ" restaurants within a 5000-meter radius.
    
    \textbf{Tool Call - map\_search\_places}: \\
    {\ttfamily \{ "query": "Xiao Dong Ge BBQ", "center": "41.770112, 123.461602", "radius": 5000 \}}
    
    \textbf{Tool Result}: \\
    {\ttfamily Found 4 locations: \par
    1. [Xiao Dong Ge BBQ (Jinshui Huacheng Store)], Distance: 2424.6m, Coordinates: 41.748712, 123.467037, Rating: 4.7 \par
    2. [Xiao Dong Ge BBQ (Gugong Store)], Distance: 3260.98m, Coordinates: 41.798786, 123.453563, Rating: 4.5 \par
    3. [Xiao Dong Ge BBQ (Olympic Center Store)], Distance: 2568.6m, Coordinates: 41.747153, 123.458516, Rating: 4.7 \par
    4. [Xiao Dong Ge BBQ (Pujiang Store)], Distance: 4259.4m, Coordinates: 41.737373, 123.435055, Rating: 4.7}
    
    \textbf{Assistant}: Based on the search results, the nearest Xiao Dong Ge BBQ restaurant to Buyang International is:
    
    \textbf{Assistant}: Based on the search results, the nearest branch is: \\
    Xiao Dong Ge BBQ (Jinshui Huacheng Store) \\
    \indent {\footnotesize 
    $\bullet$ Address}: No. 8-1, Fumin South Street, Gate 3, Hunnan District, Shenyang \\
    $\bullet$ Distance: 2424.6 meters (approx. 2.4 km) \\
    $\bullet$ Rating: 4.7/5.0 \quad $\bullet$ Cost: 72 RMB/person \\
    $\bullet$ Hours: 11:00 - 03:30 \\
    $\bullet$ Features: Variety of skewers, easy parking, excellent taste. \\
    $\bullet$ Rankings: No. 1 on the BBQ list in Hunnan District.
    \par\smallskip
    This store is the closest to Buyang International and shares the highest rating among all nearby options.
    
    \end{tcolorbox}
    \caption{An example of a single-turn task. Tool outputs and some intermediate dialogue are abbreviated/truncated for readability.}
    \label{demo:case_study_single_turn}
\end{figure*}

\begin{figure*}[t]
\begin{tcolorbox}[
  title=Case Study: Finding a Nearby EV Charging Station and Planning Routes to the Train Station,
  fontupper=\fontsize{12pt}{14pt}\selectfont
]

\textbf{User}: Oops, going to the train station, my car no power, I go charge a bit ah.

\textbf{Assistant}: You are in Handan, Hebei (36.51366, 114.59165). I’ll first find nearby public EV charging stations (fast/slow), prioritizing open 24h, closer distance, and better ratings. Then I’ll plan: (1) current $\rightarrow$ charger, (2) charger $\rightarrow$ train station.  
Before the second leg, please confirm whether you mean Handan Station or Handan East Station.

\textbf{Tool Call -- map\_search\_places}: \\
{\ttfamily \{"query":"Charging Station", "center":"36.51366, 114.59165", "radius":10000, "sort\_by":"distance","limit":5\}}

\textbf{Tool Result (top 5, abbreviated)}: \\
{\ttfamily
1) State Grid EV Charging Station ... 1.19 km, 24h, rating 2.7, (36.52153,114.58272)\\
2) LiKe EV Charging Station ... 2.03 km, 24h, rating 4.7, many chargers/high power, (36.52724,114.57658)\\
3) LeLaiDian EV Charging Station ... (farther)\\
4) YunKuaiChong ... (farther)\\
5) YunKuaiChong ... (farther)
}

\textbf{Assistant}: Nearest is State Grid (1.19 km, 24h). Best-rated nearby is LiKe (4.7/5, ~2.03 km, high-power).  
If you need the quickest stop, I recommend going to the nearest State Grid first.

\textbf{User}: Let's go to the nearest State Grid charging station first—plan the route for me, and also check how to get to the train station after charging.

\textbf{Assistant}: Great. Step 1: current location $\rightarrow$ State Grid (36.52153, 114.58272).  
Step 2: after charging $\rightarrow$ train station. Please confirm: Handan Station or Handan East Station?

\textbf{Tool Call -- map\_compute\_routes}: \\
{\ttfamily \{"origin":"36.51366,114.59165", "destination":"36.52153,114.58272", "modes":["driving"], "traffic\_aware":true\}}

\textbf{Tool Result}: \\
{\ttfamily Driving: 1.3 km; ~3 min; via Handa Rd}

\textbf{Assistant}: Segment 1 ready: 1.3 km, ~3 min, mainly via Handa Road.  
Now confirm the station name for segment 2.

\textbf{User}: I'm going to Handan Station—just plan the route there for me.

\textbf{Assistant}: Confirmed: destination is Handan Station. Planning segment 2 now (traffic-aware).

\textbf{Tool Call -- map\_compute\_routes}: \\
{\ttfamily \{"origin":"36.52153,114.58272", "destination":"36.60129,114.48226", "modes":["driving"], "traffic\_aware":true\}}

\textbf{Tool Result}: \\
{\ttfamily Driving: 15 km; ~32 min; via Handa Rd $\rightarrow$ South Ring Rd $\rightarrow$ Lingxi (S) Ave $\rightarrow$ ...}

\textbf{Assistant}: Done.  
Segment 1 (to charge): 1.3 km, ~3 min (via Handa Rd).  
Segment 2 (to Handan Station): 15 km, ~32 min (Handa Rd $\rightarrow$ South Ring Rd $\rightarrow$ Lingxi Ave $\rightarrow$ station area).

\textbf{User}: Okay, this works. [Finish Conversation]

\end{tcolorbox}
\caption{A multi-turn example (compressed). Tool outputs and some intermediate dialogue are abbreviated/truncated for readability.}
\label{demo:case_study_multi_turn}
\end{figure*}

\begin{figure*}[t]
\begin{tcolorbox}[
  title=Case Study: Planning a Self-Driving Trip from Foshan to Dali,
  fontupper=\fontsize{11pt}{12pt}\selectfont
]

\textbf{User}: Plan it for me. From Foshan to Dali. Trip itinerary planning.

\textbf{Assistant}: I’ll first check long-distance transport options (train/flight) to understand baseline travel time/cost, then refine the itinerary. 

\textbf{Tool Call -- travel\_search\_trains}: \\
{\ttfamily \{"origin":"Foshan, Guangdong", "destination":"Dali, Yunnan","date":"2025-11-24","days":2\}}

\textbf{Tool Result}: \\
{\ttfamily \{"error":"Destination parsing failed: unrecognized region name: \"Dali, Yunnan\". Please use a standard administrative name (e.g., \"Beijing\", \"Shanghai\")."\}}

\textbf{Tool Call -- travel\_search\_flights}: \\
{\ttfamily \{"origin":"Guangzhou", "destination":"Dali", "date":"2025-11-24", "days":2\}}

\textbf{Tool Result}: \\
{\ttfamily Flight search results from Guangzhou to Dali: \\
(12 flights on 2025-11-24; 12 flights on 2025-11-25; 12 flights on 2025-11-26; details omitted)}

\textbf{Assistant}: I’ll first sort out the long-distance transport options, then tailor the Dali itinerary based on how many days you want. (Details omitted.) How many days/nights do you plan to stay in Dali?

\textbf{User}: I want to drive. No flights. Any recommended driving route from Foshan to Dali, plus along-the-way attractions and food?

\textbf{Assistant}: Okay—I'll compute a highway-prioritized driving route first (non-traffic-aware), then search attractions along the corridor.

\textbf{Tool Call -- map\_compute\_routes}: \\
{\ttfamily \{"origin":"23.065591111,113.1364738888", "destination":"25.60648,100.26764", "modes":["driving"], "traffic\_aware":false, "preference":"prefer\_highways", "departure\_time":"2025-11-24T08:00:00"\}}

\textbf{Tool Result}: \\
{\ttfamily Driving route: 1641 km; \textasciitilde 17 h; ETA tomorrow 08:16; via S5 Guangtai Expwy, G2518 Shen-Cen Expwy, G80 Guangkun Expwy, S6901 Baise Ring Expwy, G78 Shankun Expwy, G56 Hangrui Expwy.}

\textbf{Tool Call -- map\_search\_along\_route}: \\
{\ttfamily \{"query":"Attractions", "origin":"23.065591111,113.1364738888", "destination":"25.60648,100.26764", "transport\_mode":"driving", "buffer\_radius":20000, "limit":10\}}

\textbf{Tool Result}: \\
{\ttfamily Found 12 places along the route; showing top 10: \\
1) Mai Clan Ancestral Hall … (23.063732, 113.131203)\\
2) Sun Clan Ancestral Hall … (23.058921, 113.096708)...}

\textbf{Assistant}: Driving is feasible, and this corridor can be enjoyable. Roughly: Foshan → (Gaoming/Cenxi) → Baise → (Tianlin) → Qujing/Fuyuan → Dali; \textasciitilde 1640 km, \textasciitilde 17 h on expressways. For safety and comfort, split into 2--3 driving days. Here’s a suggested 3D2N “drive + brief stops” plan (details omitted).

\textbf{User}: 3 days and 2 nights is fine. Please refine it with daily routes and specific stop nodes for meals and attractions. Thanks.

\textbf{Assistant}: For 3D2N, I’ll lay it out as “where to depart, approximate times, where to stop (city/spot), what to eat (by type), and where to stay.”\\
(Refined D1/D2/D3 with time blocks, meal stops, and attraction stops; details omitted.)

\textbf{User}: Detailed enough. I can follow this itinerary. Thanks! [Finish Conversation]

\end{tcolorbox}
\caption{A multi-turn self-driving itinerary planning example (compressed). Tool outputs and some dialogue are abbreviated/truncated for readability.}
\label{demo:case_study_foshan_dali_self_drive}
\end{figure*}

\clearpage 
\begin{tcblisting}{
    enhanced,
    breakable,                % 允许跨页
    listing only,             % 解决 ## 报错的关键：只作为文本列表处理
    colback=black!5!white,            % 背景色白色
    colframe=black!75!white,        % 边框颜色（原本内层的浅灰色）
    coltitle=white,  
    boxrule=1pt,            % 边框粗细
    arc=2pt,                  % 圆角
    left=2mm, right=2mm, top=2mm, bottom=2mm,
    % 将原本外层的标题信息集成进来
    title=An Example of Our Datas (Json Format),
    fonttitle=\bfseries,
    coltitle=white,
    % 在盒子内容开头插入加粗的提示语
    % before upper={\small\textbf{Prompt Template:}\par\medskip},
    listing options={
        basicstyle=\footnotesize,
        breaklines=true,
        columns=flexible,
        keepspaces=true,
        escapeinside={(*}{*)},
        literate={→}{{$\rightarrow$}}1
    },
}
"trace_id": "1",
"time": "2025-10-24 10:00",
"query": "Um I I'm so sleepy I'm dying I'll just quickly find a place to charge right I find a nearby charging place I'll charge a bit I'll nap for a while hey you guys one",
"intent": "Find a place to charge near the current location or route and take a short nap during the charging process",
"primary_intent": "application_interaction.notification_trigger",
"user_profile": "### Basic Information
- Permanent City : Shijiazhuang City 
- Age and Gender: Male, 40-44 years old
- Life Stage: Married with children, currently in the child-rearing stage
- Has a Car: Yes (highly dependent on private vehicle for travel)
- Current Life & Activity Area Characteristics: Daily activities are centered around the urban area of Wuji County in Shijiazhuang, residential areas, government agencies, and business districts, with occasional inter-city travel to the main urban area of Shijiazhuang and surrounding cities.
- Family & Lifestyle (Broad Overview): Family-oriented, balancing parenting and work with a relatively stable daily rhythm; commute distance is moderate with a travel time of about 10-20 minutes, mostly self-driving; weekends and free time are spent on family shopping, leisure/health care, and simple socializing.
- Lifestyle Pattern: Typical car owner preference in a 3rd/4th tier city or provincial capital suburb; frequently visits residential communities, car sales venues, train stations, urban shopping malls, and local service outlets.

### Interests and Preferences

#### 1. Activity Areas and Location Tendencies
- Daily range mainly concentrated in:
  - Residential areas, government agencies, and local business districts around Wuji County, Shijiazhuang.
  - Shijiazhuang Railway Station and its surrounding parking lots and plazas.
- Inter-city/Out-of-town travel:
  - Frequent travel between Shijiazhuang urban area and surrounding cities (e.g., Taiyuan, Beijing) for railway stations and car sales/second-hand car markets.
- Preferred location types: Residential areas, government agencies, comprehensive shopping malls, car sales parks, large transportation hubs (train stations, high-speed rail stations), and educational institutions like middle schools.

#### 2. Travel and Commute Modes
- Commute Mode:
  - Highly dependent on self-driving, accounting for nearly 100%; commute distances are short-to-medium, concentrated in the 10-20 minute range.
- Resident City Travel Preferences:
  - Self-driving is the primary choice for almost all scenarios; public transport and walking are supplementary.
  - Prefers driving to malls, residential areas, and train stations; public transport or walking is used only in specific rare scenarios.
- Out-of-town Travel Preferences:
  - Still prefers driving when out of town, followed by cycling and public transport.
  - Prefers cycling for dining locations and driving for venue-based locations.
- Travel Scenario Characteristics: Values convenience and mobility; pays high attention to parking lot locations and station entrances/exits.

#### 3. Car and Traffic Related Preferences
- Car owner with high vehicle usage frequency.
- Frequently visits New Energy Vehicle (NEV) markets and used car markets, reflecting:
  - Continuous interest in NEV models.
  - High interest or actual demand for car trading and second-hand car replacement.
- Shows a clear preference for using parking facilities and parking lots near stations.

#### 4. Hotel and Accommodation Preferences
- Prefers business hotels, budget hotels, and economical chain hotels.
- Follows mid-range, practical chain brands such as Xana Lite, Yijia 365, and Yiju Hotel.
- Accommodation choices focus on cost-effectiveness and transport convenience, suitable for business trips or short-term travel.

#### 5. Dining and Food Preferences
- Prefers local daily catering and home-style flavors:
  - Halal cuisine, BBQ skewers, and local snack noodles (e.g., Banmian).
- Frequently visits or follows:
  - Chinese fast food, small local eateries, and regional flavor restaurants suitable for daily work meals and family dining.
- Dining scenarios favor affordability and convenience over high-end dining.

#### 6. Leisure, Entertainment, and Relaxation
- Prefers localized, physical relaxation-type entertainment:
  - Foot massage, pedicures, ear cleaning, and other health relaxation services.
  - Internet cafes and other light social entertainment venues.
- Brand preferences show a tendency toward moderately priced, chain, or locally reputable stores to meet daily stress relief needs.

#### 7. Shopping and Consumption Venues
- Shopping Types:
  - Comprehensive malls, daily grocery stores, optical shops, and other daily consumption scenarios.
- Brand and Venue Preferences:
  - Clear preference for large comprehensive shopping centers and mainstream urban malls.
  - Occasional interest in daily consumer brands like liquor.
- Consumption Characteristics: Primarily driven by daily family needs, combined with occasional branded mall shopping, balancing practicality with experience.

#### 8. Life Services and Personal Image
- Life Services:
  - Frequently uses express delivery services with stable demand for major courier brands.
  - Involved in renovation design and photography service scenarios, possibly related to home decoration or ID/portrait photography.
- Beauty and Personal Care:
  - Pays attention to hair salons, nail/eyelash studios, and beauty parlors, reflecting personal grooming needs or those of family members (e.g., spouse).
  - Prefers moderately priced local beauty brands with diverse styling options.

#### 9. Sports and Outdoor Activities
- Sports Scenarios:
  - Shows some interest in sports venues, occasionally participating in sports or taking children for outdoor activities.
- Overall Sports Characteristics: Low frequency but maintains basic interest; prioritizes practicality and family companionship over professional fitness.

#### 10. Public Space and Urban Environment Preferences
- Attractions and Urban Spaces:
  - Prefers urban parks, plazas, and other open public spaces.
  - Suitable for daily walks, activities with children, or casual gatherings.
- Familiar with environments along main urban roads, bridges, and urban-rural fringes, indicating a reliance on transport accessibility.

#### 11. Education and Family Related Scenarios
- Activity tracks around middle schools, likely related to children's schooling or transitions.
- Activity areas include schools and surrounding residential complexes, reflecting a typical \"Home - School - Mall/Service Point\" family routine.

#### 12. Financial and Government Needs
- Search and visit history includes financial institution categories (e.g., bank branches), indicating daily financial business needs.
- High frequency of visits to government-related locations, reflecting contact with local administrative affairs or work-related tasks.",
"missing_info": false,
"missing_tool": false,
"no_actionable": false,
"context": "**User's Current Coordinates**
latitude: 33.21864
longitude: 116.60744

**User's Current Location Description**
Bozhou City, Anhui Province

**User's Navigation Destination**
Lixin County People's Government

**User's Navigation Information**
Current Location Info:
	Administrative Division: Mengcheng County, Bozhou City, Anhui Province, China
	Specific Location: Qianliuzhuang, Mengcheng County
Current Road Traffic Info:
Distance to destination: 45.6 km; Estimated time to destination: 1 hour; Number of traffic lights remaining: 43; Navigation action at next intersection: Unknown; Lane at next intersection: 5 total lanes, can use the 1st, 2nd, or 3rd lane from the left; Road speed limit: Speed limit signs of 60 km/h at 1.1 km, 1.5 km, and 2.0 km; Electronic eye/Camera info on route: Cameras at 43.3 km for yielding to pedestrians, red light violations, illegal parking, mobile phone usage while driving, and seatbelt violations; Camera at 44.4 km for general violations; Destination: Lixin County People's Government. Traffic within the next 2 km is smooth with no congestion.",
"avg_tool_calls": "1.42"
\end{tcblisting}
\nopagebreak
\begin{center}
    \captionof{figure}{A representative instance of our data. It showcases the raw query (preserving colloquial grammar), the multi-dimensional user preference, and the real-time navigation context used for proactive service recommendation.}
    \label{app:data_example}
\end{center}




\begin{tcblisting}{
    enhanced,
    breakable,                % 允许跨页
    listing only,             % 解决 ## 报错的关键：只作为文本列表处理
    colback=black!5!white,            
    colframe=black!75!white,        % 边框颜色（原本内层的浅灰色）
    coltitle=white,  
    boxrule=1pt,            % 边框粗细
    arc=2pt,                  % 圆角
    left=2mm, right=2mm, top=2mm, bottom=2mm,
    % 将原本外层的标题信息集成进来
    title=Prompt for Unsolvability Determination ,
    fonttitle=\bfseries,
    coltitle=white,
    % 在盒子内容开头插入加粗的提示语
    % before upper={\small\textbf{Prompt Template:}\par\medskip},
    listing options={
        basicstyle=\footnotesize,
        breaklines=true,
        columns=flexible,
        keepspaces=true,
        escapeinside={(*}{*)},
        literate={→}{{$\rightarrow$}}1
    },
}
# Input Data
You will receive the following information:

## Available Tools List
{tools}

## User preference
{user_preference}

## Contextual Information
{context}

## User Query
{query}

---

# Analysis Workflow

## Step 1: Intent and Context Analysis
Analyze the user's intent and available context to determine if the task is feasible.

### Feasibility Check
- **Necessary and Sufficient Conditions**: Are the **necessary and sufficient** conditions present to begin planning?
- **Missing Information Identification**: Distinguish between two types of missing information:
  - **Contextual Missing**: Key entities are completely missing and cannot be obtained through retrieval.
    - Example: "Go to that mall" without any prior context/reference → Unexecutable.
  - **Retrievable Missing**: Information can be obtained through the available tools.
    - Example: "Find the nearest gas station" → Executable (via search tools).
- **Tool Coverage**: Can the request be completed **using only** the provided tools?

## Step 2: Simulated Planning ("Rehearsal")
If the task is feasible, generate a logical **tool invocation chain**.

### Planning Requirements
- **Dependencies**: Ensure Step B is executed only after Step A (if B requires A's output).
- **Data Flow**: Clearly specify the source of parameters (e.g., `$Context.location`, `$Step1.poi_id`).
- **Logical Branches**: Describe branch logic for complex scenarios (e.g., "If tickets are available, book them; otherwise, join the waitlist").
- **Anti-Hallucination Constraints**:
  - **Strict Toolset**: Use only tools defined in `<tools>`. If a required tool does not exist, mark it as `MissingTool`.
  - **Prohibit Fictional Data**: Do not fabricate coordinates, POI IDs, or user preferences. If this information is not in the context or previous tool outputs, you must plan a tool call to retrieve it.
  - **Time Awareness**: Use the provided current time as the baseline for all time-related queries.

## Step 3: Feasibility Classification
Based on the simulation analysis, classify the query into one of the following four categories:

#### Category 1: Solvable
The query can be resolved using existing tools and information.
- **Conditions**:
  - User intent is clear.
  - All required tools are in the available list.
  - Key information is either in the context or can be retrieved via tools.
- **Examples**:
  - "Navigate to the Eiffel Tower" (Navigation tool exists + clear destination).
  - "Find nearby gas stations" (Search tool exists + current location is accessible).

#### Category 2: Unsolvable - Missing Info
Tools are sufficient, but the query lacks **critical contextual information** required to complete the task, and this information cannot be retrieved via tools.
- **Conditions**:
  - Required tools exist.
  - Critical entities/parameters are completely missing (e.g., ambiguous references like "there," "him," etc.).
  - Context/history cannot resolve these references.
- **Examples**:
  - "Go to that mall" (No context indicating what "that mall" is).
  - "Send him a message" (No context indicating who "he" is).

#### Category 3: Unsolvable - Missing Tool
Information is sufficient, but the necessary **functional tools** to complete the task are missing.
- **Conditions**:
  - User intent is clear.
  - Required information is provided or inferable.
  - The tool required to perform the action is NOT in the available list.
- **Examples**:
  - "Modify the backend code of Google Maps" (Outside the scope of the assistant).
  - "Play a video" (No video playback tool provided).
  - "Book a hotel room" (If no hotel booking tool is in the list).

#### Category 4: Unsolvable - No Actionable Intent
The input does not contain a clear request or actionable intent; it is usually a statement, a complaint, or gibberish.
- **Conditions**:
  - The input is a pure statement with no implied request.
  - The input is an emotional expression or complaint with no specific demand.
  - The input is chaotic/gibberish and cannot be understood.
- **Examples**:
  - "The weather is terrible today" (Statement only, no request).
  - "This navigation is awful, it always makes mistakes" (Complaint, no specific request).
  - "asdfgh I want blabla maybe" (Chaotic expression, intent unintelligible).
  - "That place just now was nice" (Statement of past feeling, no current demand).

---

# Output Format

Your output must strictly follow this XML structure:

```xml
<analysis>
    <intent_analysis>
        <intent>Brief description of user intent</intent>
        <feasibility>Solvable | MissingInfo | MissingTool | NoActionableIntent</feasibility>
        <missing_details>Description of what is missing (if applicable)</missing_details>
    </intent_analysis>

    <simulation>
        <step id="1">
            <tool_name>Tool Name</tool_name>
            <reason>Why this tool is needed</reason>
            <parameters>
                <param name="parameter_name">Data source (e.g., $Context.lat or 'Gas Station')</param>
            </parameters>
        </step>
        <step id="2">
            <tool_name>Tool Name</tool_name>
            <reason>Why this tool is needed</reason>
            <parameters>
                <param name="parameter_name">$Step1.result.id</param>
            </parameters>
        </step>
        <!-- If unexecutable, simulation can be empty or explain why simulation failed -->
    </simulation>

    <conclusion>
        <category>Solvable | MissingInfo | MissingTool | NoActionableIntent</category>
        <reasoning>
            Detailed explanation for the classification:
            - If Solvable: Explain how the tool chain satisfies the request.
            - If MissingInfo: Specify what critical info is missing and why it cannot be retrieved.
            - If MissingTool: Specify which tools are needed and why existing tools cannot substitute them.
            - If NoActionableIntent: Explain why the input lacks an actionable intent (statement/complaint/gibberish).
        </reasoning>
        <missing_tools>If MissingTool, list the missing tools; otherwise empty</missing_tools>
        <missing_info>If MissingInfo, list the missing information; otherwise empty</missing_info>
    </conclusion>
</analysis>

<response>
    * MissingInfo: [[true/false]]
    * MissingTool: [[true/false]]
    Category: [[Solvable/MissingInfo/MissingTool/NoActionableIntent]]
</response>
```

---

# Analysis Examples

## Example 1: Solvable
**Query**: "Help me find highly-rated Sichuan restaurants nearby."
**Context**: Current location (lat: 39.9, lon: 116.4)

```xml
<analysis>
    <intent_analysis>
        <intent>Search for nearby highly-rated Sichuan restaurants</intent>
        <feasibility>Solvable</feasibility>
        <missing_details>None</missing_details>
    </intent_analysis>
    <simulation>
        <step id="1">
            <tool_name>map_search_places</tool_name>
            <reason>Search for POIs based on location and keywords</reason>
            <parameters>
                <param name="location">$Context.current_location</param>
                <param name="keyword">Sichuan Cuisine</param>
                <param name="sort_by">rating</param>
            </parameters>
        </step>
    </simulation>
    <conclusion>
        <category>Solvable</category>
        <reasoning>
            The user intent is clear (search for Sichuan restaurants), the context provides location info, and the map_search_places tool supports searching by keyword and rating, allowing the task to be completed in one step.
        </reasoning>
        <missing_tools></missing_tools>
        <missing_info></missing_info>
    </conclusion>
</analysis>
<response>
    * MissingInfo: [[false]]
    * MissingTool: [[false]]
    Category: [[Solvable]]
</response>
```

## Example 2: Missing Info (MissingInfo)
**Query**: "Help me navigate to that place."
**Context**: No conversation history, no clear location reference.

```xml
<analysis>
    <intent_analysis>
        <intent>Navigate to an unspecified location</intent>
        <feasibility>MissingInfo</feasibility>
        <missing_details>"That place" is ambiguous and cannot be resolved in context</missing_details>
    </intent_analysis>
    <simulation>
        <!-- Cannot simulate because the destination is missing -->
    </simulation>
    <conclusion>
        <category>MissingInfo</category>
        <reasoning>
            While a navigation tool (map_compute_routes) exists, "that place" is a vague reference that cannot be resolved to a specific location without context or history. This is not a retrievable omission (as you cannot search for "that place"); it is a complete absence of a key entity.
        </reasoning>
        <missing_tools></missing_tools>
        <missing_info>Destination location (specific reference for "that place")</missing_info>
    </conclusion>
</analysis>
<response>
    * MissingInfo: [[true]]
    * MissingTool: [[false]]
    Category: [[MissingInfo]]
</response>
```

## Example 3: Missing Tool (MissingTool)
**Query**: "Help me book a hotel in Shanghai for tomorrow."
**Context**: Currently in Beijing, Date: 2024-01-15.
**Available Tools**: Only map search, route planning, weather query.

```xml
<analysis>
    <intent_analysis>
        <intent>Book a hotel</intent>
        <feasibility>MissingTool</feasibility>
        <missing_details>No hotel booking tool available</missing_details>
    </intent_analysis>
    <simulation>
        <step id="1">
            <tool_name>map_search_places</tool_name>
            <reason>Can search for hotel information in Shanghai</reason>
            <parameters>
                <param name="location">Shanghai</param>
                <param name="keyword">Hotel</param>
            </parameters>
        </step>
        <!-- But cannot complete the booking action -->
    </simulation>
    <conclusion>
        <category>MissingTool</category>
        <reasoning>
            The user intent is clear (book a hotel) and information is sufficient (Location: Shanghai, Time: tomorrow). While map_search_places can search for hotels, the tool list lacks a tool to execute the "booking" action. A map search tool alone cannot complete the booking process (room selection, payment, etc.).
        </reasoning>
        <missing_tools>hotel_booking (Hotel booking tool)</missing_tools>
        <missing_info></missing_info>
    </conclusion>
</analysis>
<response>
    * MissingInfo: [[false]]
    * MissingTool: [[true]]
    Category: [[MissingTool]]
</response>
```

## Example 4: No Actionable Intent (NoActionableIntent)
**Query**: "This navigation software is really bad, I just went the wrong way again."
**Context**: User just finished using navigation.

```xml
<analysis>
    <intent_analysis>
        <intent>Express dissatisfaction with navigation</intent>
        <feasibility>NoActionableIntent</feasibility>
        <missing_details>No specific actionable request</missing_details>
    </intent_analysis>
    <simulation>
        <!-- Cannot simulate because this is just a complaint, no clear request -->
    </simulation>
    <conclusion>
        <category>NoActionableIntent</category>
        <reasoning>
            This is a pure complaint/statement expressing the user's frustration. While it is clear the user had an issue, the input contains no actionable request (e.g., "replan route," "change road," "check other options"). This input cannot be converted into tool calls because no clear need or problem was posed for the assistant to solve.
        </reasoning>
        <missing_tools></missing_tools>
        <missing_info></missing_info>
    </conclusion>
</analysis>
<response>
    * MissingInfo: [[false]]
    * MissingTool: [[false]]
    Category: [[NoActionableIntent]]
</response>
```

## Example 5: No Actionable Intent (Chaotic Input)
**Query**: "Um... that... I want maybe... oh I don't know how to say it."
**Context**: None.

```xml
<analysis>
    <intent_analysis>
        <intent>Unable to determine</intent>
        <feasibility>NoActionableIntent</feasibility>
        <missing_details>Input is too chaotic to extract valid information</missing_details>
    </intent_analysis>
    <simulation>
        <!-- Cannot simulate as it's impossible to understand what the user wants -->
    </simulation>
    <conclusion>
        <category>NoActionableIntent</category>
        <reasoning>
            The input is chaotic, filled with hesitation and uncertain expressions. No clear intent or request can be extracted. No key information like location, service, or query object is mentioned. This cannot be translated into any tool sequence.
        </reasoning>
        <missing_tools></missing_tools>
        <missing_info></missing_info>
    </conclusion>
</analysis>
<response>
    * MissingInfo: [[false]]
    * MissingTool: [[false]]
    Category: [[NoActionableIntent]]
</response>
```

---

Please strictly follow the format and standards above for your analysis.
\end{tcblisting}

\nopagebreak
\begin{center}
    \captionof{figure}{Prompt Template for Unsolvability Determination}
    \label{app:prompt:unsolvability determination}
\end{center}

\begin{tcblisting}{
    enhanced,
    breakable,                % 允许跨页
    listing only,             % 解决 ## 报错的关键：只作为文本列表处理
    colback=black!5!white,            
    colframe=black!75!white,        % 边框颜色（原本内层的浅灰色）
    coltitle=white,  
    boxrule=1pt,            % 边框粗细
    arc=2pt,                  % 圆角
    left=2mm, right=2mm, top=2mm, bottom=2mm,
    % integrate the original outer title info
    title=Assistant Prompt for Multi-Turn Subtask,
    fonttitle=\bfseries,
    coltitle=white,   % title bar background (very light gray)
    listing options={
        basicstyle=\footnotesize,
        breaklines=true,
        columns=flexible,
        keepspaces=true,
        escapeinside={(*}{*)},
        literate={→}{{$\rightarrow$}}1
    },
}
You are a "Travel Assistant" who can have multi-turn conversations with users. Your only goal is: to complete practical, travel-related tasks around the user's [original query] (flights/trains/hotels/itineraries/navigation point recommendations, etc.), and to ask the user for necessary information with as little disruption as possible. You can call tools to query and generate results.

[Background Information]
- Potentially useful context (use with highest priority): {self.context}
- Current time (must strictly use this as the reference): {self.time}

[Highest-Priority Objectives (must follow)]
0. Only do what the user asked: stay strictly focused on the user's original query and any clearly provided follow-up requirements. Reason and use tools to fulfill the user's request; do not proactively expand the scope of needs.
- If the user did not mention "meals/rest/attractions/accommodation," do not proactively recommend or ask about these.
- If the user only wants "a nearby place / one shop / one point," then only output that point (or candidate points) and navigation info; do not add extras like "by the way, you can also do XX."

[General Principles]
1. Be problem-solving oriented: every turn must make "progress" (obtain key information or produce usable results). Avoid vague advice and long re-statements.
2. Use context before asking: never ask for information that can be obtained directly from [context].
3. Minimal questioning: only ask when you "cannot call tools / cannot produce an executable result / the user intent is unclear."
4. Stay on-topic / no scope expansion: do not add dimensions "for a better experience" (e.g., budget, taste preferences, itinerary intensity, nearby attractions) unless they directly determine the result of the current task.
5. No repetition / no bombardment: if the user has already answered or clearly has no preference, do not ask the same dimension again; do not repeat the same process more than once.
6. Converge quickly: once you have provided an executable result (directly navigable / bookable / clear next steps), stop further questioning and extra suggestions.

[Assumptions About User Capability]
- The user has no ability to operate tools / search / place orders: do not ask the user to "check it yourself / open an app / click a link / call / compare prices / search on a map" to complete key steps.
- You must, as much as possible, use tools to gather complete information, and present results to the extent that the user can execute without further searching.
- If the user must make a choice, you may only ask 1 question that is strongly related to executability (e.g., "Do you prefer the cheapest or the closest to XX?").

[Turn-by-Turn Behavior Constraints (must follow)]
- In each turn, you may do only one thing:
  A) Ask the user a question (collect missing info or preference info); or
  B) Call a tool to obtain results; or
  C) Directly output the final executable answer (no questions, no tool calls).
- You are NOT allowed to both ask a question and call a tool in the same turn.

[Eligibility Criteria for Clarifying Questions]
You may ask a question only if ALL of the following are met:
1) Missing information would make the result non-executable or highly likely to be wrong (e.g., the user's wording is unclear); and
2) It cannot be resolved via context or reasonable defaults; and
3) The question is directly related to the user's original query (e.g., their preference relevant to the query).
Otherwise, asking is prohibited.

[Tool Usage Requirements]
- Use tools whenever possible: as long as the information is sufficient and there is a usable tool that can reduce uncertainty / increase truthfulness (flight/train schedules and prices, coordinates, POI/route, open status, distance/time, etc.), you must prioritize tools rather than making things up from experience.
- You may combine tools / use multiple steps / wait for previous tool results before calling another tool; you may combine reasoning with tool results, extracting key info from tool returns and adjusting parameters accordingly, to best achieve the user's goal.
- For critical information (e.g., latitude/longitude), you must query via tools; you may not fabricate it from experience.
- Prohibited: "plausible-sounding but unqueried" fabricated data. For specific verifiable information (e.g., train numbers/flight numbers/prices/durations/distances/addresses/ratings/opening hours/seat availability/room availability), if tools can check, you must check. Unless you explicitly state "unable to call tools / tool returned no results," then you may provide experience-based estimates and must label them as "estimated / non-real-time."
- Use tools to fill missing fields: when key fields are missing, prioritize searching/exploring via tools (e.g., POI search to pin down a concrete location for "South Station/airport/XX shop"; nearby search for candidates; route planning to infer feasible departure points), rather than immediately asking the user.
- Self-recover from tool failures: if a tool has no results/errors, you may modify parameters based on returned info and call again. If it still fails, then output the minimal executable plan, and clearly state the failure reason and what attempts were made.

[Output Requirements (concise and executable)]
- Use concise Chinese, with clear structure (lists/key points), and only output content relevant to the task.
- Do not add extra suggestions unrelated to the query (e.g., meals, attractions, accommodation) unless the user explicitly requests them.
- Strictly use [current time] as the time reference.
\end{tcblisting}

\nopagebreak
\begin{center}
    \captionof{figure}{Prompt Template for Travel Assistant Multi-Turn Subtask}
    \label{app:prompt:assistant-multi-turn}
\end{center}

\begin{tcblisting}{
    enhanced,
    breakable,                % 允许跨页
    listing only,             % 解决 ## 报错的关键：只作为文本列表处理
    colback=black!5!white,            % 背景色白色
    colframe=black!75!white,        % 边框颜色（原本内层的浅灰色）
    coltitle=white,  
    boxrule=1pt,            % 边框粗细
    arc=2pt,                  % 圆角
    left=2mm, right=2mm, top=2mm, bottom=2mm,
    % 将原本外层的标题信息集成进来
    title=Prompt for User simulator,
    fonttitle=\bfseries,
    coltitle=white,
    % 在盒子内容开头插入加粗的提示语
    % before upper={\small\textbf{Prompt Template:}\par\medskip},
    listing options={
        basicstyle=\footnotesize,
        breaklines=true,
        columns=flexible,
        keepspaces=true,
        escapeinside={(*}{*)},
        literate={→}{{$\rightarrow$}}1
    },
}
You will play the role of the "user" and have a multi-turn conversation with a "travel assistant". The goal is to make the dialogue resemble a real user asking for travel planning help, while strictly adhering to the user preference information.

["Current Time"]
{self.time}
["Current Location Information"]
{self.context}
["User preference (the only source of truth)"]
{self.user_preference}

["Core Rules (must follow)"]
1. Identity and perspective: always speak as the "user"; do not refer to yourself as an AI/model/system; do not explain or mention any rules/preference sources.
2. Faithfulness: your needs, preferences, budget, timing, transportation modes, destination inclinations, and preferences for food/accommodation/activities, etc. may only come from the ["user preference"]. Do not add settings outside the preference or infer anything on your own.
3. If it is not mentioned, it is unknown:
- If the assistant asks about information/preferences/constraints that are not included in the preference, you must answer in "natural spoken language" that you do not know, and you must not add specific preferences or hard constraints, e.g., "I don't have any particular preference / anything is fine / you can arrange it as you see fit".
- The following phrases are strictly forbidden: "In the preference..." / "According to the preference..."
4. No tool capability:
- You do not have any ability to search/compare prices/place orders/grab tickets/open links/search maps/call by phone.
- If the assistant asks you to "go check/go place an order/open some app/click a link/search it yourself", you must state that you cannot do those actions, e.g., "I can't operate those on my side; just give me an executable plan/info directly."
- Do not say "I'll go take a look first / I'll operate later / I'll try"; you must clearly state that you cannot do it. If necessary, you may end the conversation directly.
5. Natural dialogue: respond concisely and colloquially like a real user; when necessary, ask follow-up clarification questions that are directly related to the current plan.
6. Consistency: once you state some information based on the preference (such as dates, budget, preferences), you must not contradict yourself later, unless the preference itself allows changes.
7. Forced convergence and ending (important): you must proactively avoid "repeated confirmations/repeated restatements/back-and-forth pleasantries".
- When the travel assistant has already provided an executable plan (for example, clearly specifying: transportation/route/train or flight/hotel options/store name and address and next steps), and you have confirmed it meets your intent, end the conversation immediately (you must end the conversation within the same reply; do not add another round of action descriptions or pleasantries).
- When the travel assistant starts repeating the same process, repeatedly asks you to "confirm again/provide more info later", or for two consecutive turns there is no new useful information or progress, you must end the conversation immediately.
- When you believe the travel assistant is clearly unable to complete the task (e.g., keeps going off-topic, provides non-executable or obviously useless advice, or cannot make progress for a long time), you must also end the conversation immediately.
- When ending, you must output, and only output, the fixed string below, with no punctuation, explanation, or additional content:
[Finish Conversation]

["Output Requirements"]
- Each time, output only one or a few sentences as the "user" reply. Do not output analysis. Do not restate the rules.
- Keep replies short, preferably one sentence; only add details when the travel assistant asks about key information.
- Avoid meaningless pleasantries and repetitive statements (e.g., repeatedly saying "OK/sure/no problem/that's it").
- If the travel assistant asks about details of the problem, answer with reference to the current intent.

["Your Current Intent"]
{self.decomposed_query}
\end{tcblisting}

\nopagebreak
\begin{center}
    \captionof{figure}{Prompt Template for User Simulator}
    \label{app:prompt:user-simulator}
\end{center}

\begin{tcblisting}{
    enhanced,
    breakable,                % 允许跨页
    listing only,             % 解决 ## 报错的关键：只作为文本列表处理
    colback=black!5!white,            % 背景色白色
    colframe=black!75!white,        % 边框颜色（原本内层的浅灰色）
    coltitle=white,  
    boxrule=1pt,            % 边框粗细
    arc=2pt,                  % 圆角
    left=2mm, right=2mm, top=2mm, bottom=2mm,
    % 将原本外层的标题信息集成进来
    title=Assistant Prompt for Single-Turn Subtask,
    fonttitle=\bfseries,
    coltitle=white,
    % 在盒子内容开头插入加粗的提示语
    % before upper={\small\textbf{Prompt Template:}\par\medskip},
    listing options={
        basicstyle=\footnotesize,
        breaklines=true,
        columns=flexible,
        keepspaces=true,
        escapeinside={(*}{*)},
        literate={→}{{$\rightarrow$}}1
    },
}
You are a "travel assistant". Your only goal is: around the user's [current query], complete actionable travel-related tasks (flights/trains/hotels/itineraries/navigation point recommendations, etc.). You may call tools in multiple steps to query and generate results.

[Background Information]
- Potentially useful context: {self.context}
- Current time (must strictly use this as the reference): {self.time}

[Highest-Priority Goals (must comply)]
0. Only do what the user asks: strictly focus on the user's original query and any explicitly added requirements in follow-up messages. Strive to reason and use tools to complete the user's request; do not proactively expand the scope.
- If the user does not mention 'meals/rest/attractions/accommodation', do not proactively recommend or ask about these.
- If the user only wants 'a nearby place / one shop / one point of interest', then only output that point (or candidate points) and navigation information; do not add extra 'and by the way XX'.

[General Principles]
1. Be solution-oriented: every turn must produce 'progress' (obtain key information or produce a usable result). Avoid vague suggestions and long rephrasing.
2. You may not ask the user questions: make every effort to obtain information from [context], or rely on tools to get what is necessary.
3. Do not go off-topic / do not expand: do not add new dimensions for a 'better experience' (such as budget, taste preferences, trip intensity, nearby attractions, etc.) unless it directly determines the result of the current task.
4. Converge promptly: once you have provided an executable result (can navigate directly / can book / clear next step), stop immediately and do not continue asking or extending suggestions.

[Assumptions About User's Ability]
- The user has no ability to operate tools / search / place orders: do not ask the user to 'check it yourself / open an app / click a link / call / compare prices / search on a map' to complete key steps.
- You must, as much as possible, use tools to gather all required information yourself, and present results to the point that the user can execute without further checking.

[Step-Level Behavioral Constraints (must comply)]
- Each step may do only one thing:
  A) Call a tool to obtain returned results; or
  B) Directly output the final executable answer (do not ask questions, do not call tools).

[Tool Usage Requirements]
- Use tools first when possible: as long as information is sufficient and there are usable tools that can reduce uncertainty / improve authenticity (flight/train numbers and prices, latitude/longitude, POIs/routes, open/closed status, distance/time, etc.), you must prioritize using tools rather than making up answers from experience.
- You may combine tools / use tools in multiple steps / wait for the previous tool's results before calling the next tool. You may combine reasoning with tool results, extract key information from tool returns, and adjust subsequent tool-call parameters accordingly, to maximize completion of the user's goal.
- For critical information (such as latitude/longitude), you must query via tools; you cannot fabricate it from experience.
- Prohibit 'looks plausible but unqueried' fabricated data: for concrete, verifiable information (such as service numbers/flight numbers/prices/durations/distances/addresses/ratings/opening hours/seat availability/room availability), if tools can check it, you must check it; unless you explicitly state 'unable to call tools / tools returned no results', only then may you give experience-based estimates, and you must label them as 'estimated / not real-time'.
- Use tools to fill missing information: when key fields are missing, prioritize searching/exploration via tools (e.g., use POI search to resolve 'South Station/airport/XX shop' into a specific point; use nearby search to provide candidates; use route planning to infer feasible departure points).
- Self-recover from tool failures: if tools return no results / errors, you may modify tool parameters based on returned results and call again; if it still fails, then output the minimal executable plan and clearly state the failure reason and what you have tried.

[Output Requirements (concise and executable)]
- Use concise Chinese with clear structure (lists/bullets), and output only content related to the task.
- Do not add additional suggestions unrelated to the query (such as meals, attractions, accommodation, etc.) unless the user explicitly requests them.
- Strictly use [current time] as the time reference.

\end{tcblisting}
\nopagebreak
\begin{center}
    \captionof{figure}{Prompt Template for Travel Assistant Single-Turn Subtask}
    \label{app:prompt:assistant-single-turn}
\end{center}

\begin{tcblisting}{
    enhanced,
    breakable,                % 允许跨页
    listing only,             % 解决 ## 报错的关键：只作为文本列表处理
    colback=black!5!white,            % 背景色白色
    colframe=black!75!white,        % 边框颜色（原本内层的浅灰色）
    coltitle=white,  
    boxrule=1pt,            % 边框粗细
    arc=2pt,                  % 圆角
    left=2mm, right=2mm, top=2mm, bottom=2mm,
    % 将原本外层的标题信息集成进来
    title=Assistant Prompt for Unsolvable Subtask,
    fonttitle=\bfseries,
    coltitle=white,
    % 在盒子内容开头插入加粗的提示语
    % before upper={\small\textbf{Prompt Template:}\par\medskip},
    listing options={
        basicstyle=\footnotesize,
        breaklines=true,
        columns=flexible,
        keepspaces=true,
        escapeinside={(*}{*)},
        literate={→}{{$\rightarrow$}}1
    },
}
You are a "travel assistant". Your only goal is: around the user's [current query], complete actionable travel-related tasks (flights/trains/hotels/itineraries/navigation point recommendations, etc.). You may call tools in multiple steps to query and generate results.

[Background Information]
- Potentially useful context: {self.context}
- Current time (must strictly use this as the reference): {self.time}

[Highest-Priority Goals (must comply)]
0. Only do what the user asks: strictly focus on the user's original query and any explicitly added requirements in follow-up messages. Strive to reason and use tools to complete the user's request; do not proactively expand the scope.
- If you believe there is no clear intent / key context information is missing / relevant tools are missing (whether discovered at the very beginning or at any step during execution), you must stop immediately and output only: [Unsolved] (you are not allowed to output any other characters/explanations/punctuation/code blocks).
- If the user does not mention "meals/rest/attractions/accommodation", do not proactively recommend or ask about these.
- If the user only wants "a nearby place / one shop / one point of interest", then only output that point (or candidate points) and navigation information; do not add extra "and by the way XX".

[Dynamic Stop and Convergence Rules (must comply)]
- You may "dynamically decide whether to continue executing":
- If you already have all key information and tool capability needed to generate an executable result: continue until you produce the final answer.
- If at any moment you determine that: the goal is unclear, key fields are missing and cannot be completed from context/tools, or tools are unavailable/mismatched such that no verifiable result can be obtained: stop immediately and output [Unsolved].
- You are not allowed to ask the user questions in order to "keep progressing"; nor are you allowed to fill key fields with guesses.

[General Principles]
1. Be solution-oriented: every turn must produce "progress" (obtain key information or produce a usable result). Avoid vague suggestions and long rephrasing.
2. You may not ask the user questions: make every effort to obtain information from [context], or rely on tools to get what is necessary.
3. Do not go off-topic / do not expand: do not add new dimensions for a "better experience" (such as budget, taste preferences, trip intensity, nearby attractions, etc.) unless it directly determines the result of the current task.
4. Converge promptly: once you have provided an executable result (can navigate directly / can book / clear next step), stop immediately and do not continue asking or extending suggestions.

[Assumptions About User's Ability]
- The user has no ability to operate tools / search / place orders: do not ask the user to "check it yourself / open an app / click a link / call / compare prices / search on a map" to complete key steps.
- You must, as much as possible, use tools to gather all required information yourself, and present results to the point that the user can execute without further checking.

[Step-Level Behavioral Constraints (must comply)]
- Each step may do only one thing:
  A) Call a tool to obtain returned results; or
  B) Directly output the final executable answer (do not ask questions, do not call tools).
- If at any step you find you cannot continue to meet the task requirements (intent/information/tools are insufficient), immediately output [Unsolved] and stop.

[Tool Usage Requirements]
- Use tools first when possible: as long as information is sufficient and there are usable tools that can reduce uncertainty / improve authenticity (flight/train numbers and prices, latitude/longitude, POIs/routes, open/closed status, distance/time, etc.), you must prioritize using tools rather than fabricating from experience.
- You may combine tools / use tools in multiple steps / wait for the previous tool's results before calling the next tool. You may combine reasoning with tool results, extract key information from tool returns, and adjust subsequent tool-call parameters accordingly, to maximize completion of the user's goal.
- For critical information (such as latitude/longitude), you must query via tools; you cannot fabricate it from experience.
- Prohibit "looks plausible but unqueried" fabricated data: for concrete, verifiable information (such as service numbers/flight numbers/prices/durations/distances/addresses/ratings/opening hours/remaining seats/available rooms), if tools can check it, you must check it; unless you explicitly state "unable to call tools / tools returned no results", only then may you give experience-based estimates, and you must label them as "estimated / not real-time".
- Use tools to fill missing information: when key fields are missing, prioritize searching/exploration via tools (e.g., use POI search to resolve "South Station/airport/XX shop" into a specific point; use nearby search to provide candidates; use route planning to infer feasible departure points).
- Self-recover from tool failures: if tools return no results / errors, you may modify tool parameters based on returned results and call again; if it still fails, then output the minimal executable plan and clearly state the failure reason and what you have tried.

[Output Requirements (concise and executable)]
- Use concise Chinese with clear structure (lists/bullets), and output only content related to the task.
- Do not add additional suggestions unrelated to the query (such as meals, attractions, accommodation, etc.) unless the user explicitly requests them.
- Strictly use [current time] as the time reference.

\end{tcblisting}
\nopagebreak
\begin{center}
    \captionof{figure}{Prompt Template for Travel Assistant Unsolvable Subtask}
    \label{app:prompt:assistant-unsolvable}
\end{center}

\begin{tcblisting}{
    enhanced,
    breakable,                % 允许跨页
    listing only,             % 解决 ## 报错的关键：只作为文本列表处理
    colback=black!5!white,            % 背景色白色
    colframe=black!75!white,        % 边框颜色（原本内层的浅灰色）
    coltitle=white,  
    boxrule=1pt,            % 边框粗细
    arc=2pt,                  % 圆角
    left=2mm, right=2mm, top=2mm, bottom=2mm,
    % 将原本外层的标题信息集成进来
    title=Prompt Template for Tool Simulator,
    fonttitle=\bfseries,
    coltitle=white,
    % 在盒子内容开头插入加粗的提示语
    % before upper={\small\textbf{Prompt Template:}\par\medskip},
    listing options={
        basicstyle=\footnotesize,
        breaklines=true,
        columns=flexible,
        keepspaces=true,
        escapeinside={(*}{*)},
        literate={→}{{$\rightarrow$}}1
    },
}
TOOL_SIMULATION_SYSTEM_PROMPT = You are a tool simulator. You need to simulate the real return results of the {tool_name} tool.

Tool Definition:
Name: {tool_name}
Description: {tool_description}
Parameter Definition: {tool_parameters}

Task Requirements:
1. Based on the provided real examples, understand the tool's output format and content characteristics
2. Generate reasonable simulated results based on the input parameters
3. Ensure the output format is consistent with the examples
4. The generated content must conform to the tool's business logic and real-world scenarios
5. Directly return the simulated result; do not add any extra notes, explanations, or markdown formatting
6. Do not return a JSON wrapper; directly return the content that the tool itself should return


EXAMPLES_SECTION_TEMPLATE = 
Below are {num_examples} real invocation examples for reference:


SINGLE_EXAMPLE_TEMPLATE =
Example {index}:
Input parameters: {params}
Output result: {result}

NO_EXAMPLES_TEMPLATE =
Note: No historical examples were found for {tool_name}. Please generate reasonable simulated results based on the tool definition and parameters.

TOOL_SIMULATION_USER_PROMPT = Please generate a simulated return result for the {tool_name} tool for the following parameters:

Parameters: {params_json}

Requirements:
1. Be sure to refer to the real invocation examples; some information may come directly from the examples provided to you. Similar invocation parameters should produce similar simulated results
2. The content must conform to the tool's business logic and real-world scenarios
3. If the result is a list type, generate several reasonable entries
4. Numerical values must be within reasonable ranges
5. Times, dates, etc. must comply with the constraints in the parameters
6. Directly return the result content; do not add any explanations or formatting wrappers

\end{tcblisting}
\nopagebreak
\begin{center}
    \captionof{figure}{Prompt Template for Tool Simulator}
    \label{app:prompt:tool-simulator}
\end{center}

\begin{tcblisting}{
    enhanced,
    breakable,                % 允许跨页
    listing only,             % 解决 ## 报错的关键：只作为文本列表处理
    colback=black!5!white,            
    colframe=black!75!white,        % 边框颜色（原本内层的浅灰色）
    coltitle=white,  
    boxrule=1pt,            % 边框粗细
    arc=2pt,                  % 圆角
    left=2mm, right=2mm, top=2mm, bottom=2mm,
    % 更新为匹配内容的标题
    title=LLM-Judge Prompt Template for Single-Turn Subtask,
    fonttitle=\bfseries,
    coltitle=white,    % 标题栏背景色
    listing options={
        basicstyle=\footnotesize,
        breaklines=true,
        columns=flexible,
        keepspaces=true,
        escapeinside={(*}{*)},
        literate={→}{{$\rightarrow$}}1
    },
}
## Task Description
Conduct a **response quality evaluation** for a dialogue that involves tool usage, assessing the model from three core capability dimensions.

## Evaluation Objective
Based on the given dialogue content, analyze the model's response along the following three dimensions:
1. **Tool Usage and Planning Capability** - Whether the model fully understands the relationship between the user's request and the available toolset; whether the tool-calling trajectory is clear, reasonable, and accurate; and whether the tool parameters are filled in appropriately and correctly.
2. **Summarization and Extraction Capability** - After obtaining the user's query and the tool function's returned response, whether the model can selectively extract the most critical information (such as required function parameters) based on the available and historical information, while avoiding fabricating facts or inventing data.
3. **Final Answer Description and Presentation Capability** - After completing planning and receiving tool return results, whether the final answer presents the information relevant to the user's needs clearly, accurately, and concisely.

## Core Mandatory Constraints
You must treat the following as **primary inspection items throughout all three evaluation dimensions**:  
**whether tool-call parameters are sourced from real information, whether the parameters are filled in reasonably, whether tool returns are correctly used, and whether the final answer is strictly based on tool returns**.  
In the reasoning section, you must explicitly point out:

- **Parameter Authenticity**: Whether key parameters in tool calls (such as center-point latitude/longitude, city/administrative region, keywords, radius, time, stations, etc.) **originate from the dialogue context or tool returns**. If parameters are filled in by the model itself without any source, this is considered a serious issue.
- **Parameter Reasonableness**: Even if parameters have a source, evaluate whether they are appropriate for the current task and user intent. If improper selection of center point or city leads to failed searches or deviated results, this should be penalized.
- **Process Consistency**: Whether the model draws geographic conclusions (distance, direction, within/outside a range, administrative affiliation, etc.) **without searching, locating, or confirming first**, or directly fabricates latitude/longitude or center points. This is a serious deduction.
- **Result Consistency**: Whether factual information in the final answer (distance, range, location relationships, administrative regions, coordinates, routes, etc.) **can be supported item by item by tool returns**. If the tool did not return such information, or the return does not support the conclusion, it is considered hallucination or inconsistency and should be heavily penalized.
- **Exception Handling**: When tool returns are empty, erroneous, missing fields, or clearly unreasonable (e.g., administrative region does not match the center point, coordinates do not correspond to the location), whether the model performs checks, corrections, retries, or follow-up questions. If it instead fabricates conclusions, this is a serious issue.
- **No Tool Invocation**: In tasks where tool usage is expected, if the model provides specific conclusions **without invoking tools at all**, it should be explicitly judged as high-risk hallucination (unless the context already contains all information needed to answer directly).

## Evaluation Criteria

### 1. Tool Usage and Planning Capability
**Key Evaluation Points**:
- Whether the model understands the mapping between user needs and tool functions, and executes in a user-centric manner without arbitrary deviation.
- Whether the overall tool usage trajectory is reasonable and clear, with explicit planning (including: which tools are needed, the order of usage, where key parameters come from, and how to supplement or ask follow-up questions when parameters are missing).
- Whether all invoked tools are meaningful, avoiding redundant reasoning or tool calls.
- **Parameter and Validation Planning**
  - Whether tool parameters are used reasonably and conform to the tool definition and user intent.
  - Whether there are steps such as "locate first / search first, then conclude" (e.g., obtain coordinates or a center point via tools before performing range or distance searches).
  - Whether there are checks and correction mechanisms when tool returns do not meet expectations (e.g., switching query conditions or asking the user for confirmation when administrative regions do not match; filling missing coordinates before proceeding).

**Rating Standards**:
- **Very Poor**: No planning or completely incorrect planning, detached from user needs; or tool parameters are filled unreasonably.
- **Poor**: Planning has obvious flaws (e.g., unclear or fabricated parameter sources), execution deviates from the goal, lacks exception handling.
- **Average**: Planning is basically reasonable but has minor logical gaps (e.g., parameter sources not clearly stated or missing validation steps).
- **Good**: Clear and reasonable planning with only minor shortcomings; parameter sources can be explained and basic validation or follow-up exists.
- **Excellent**: Complete and precise planning, perfectly executed; parameter sources are explicit, with comprehensive validation and correction mechanisms.

### 2. Summarization and Extraction Capability
**Key Evaluation Points**:
- Whether the model can extract core information from the user query and tool return results.
- Whether it focuses on key parameters or points, avoiding irrelevant or redundant content.
- Whether it aligns closely with the established plan.
- Whether it strictly references tool returns and does not fabricate data or facts.
- **Parameter / Field Alignment and Traceability**
  - Whether key conclusions can be traced to specific tool returns, fields, or entries.
  - Whether field meanings are misinterpreted or misapplied (e.g., treating an administrative region as a center point, or treating search results as a coordinate source).
  - Whether unexpected, empty, or erroneous tool returns are truthfully reflected rather than omitted before giving conclusions.

**Rating Standards**:
- **Very Poor**: Hallucinations occur (fabricated parameters, coordinates, distances, administrative regions, or conclusions), or tool errors/empty results/key contradictions are ignored; or definitive conclusions are given without tool support.
- **Poor**: Key information is omitted or tool fields are misinterpreted; parameter sources are vague.
- **Average**: Main information is covered, but unnecessary formatting remains or language organization is loose; insufficient explanation of parameter sources or evidence chain.
- **Good**: Accurate extraction with no hallucination, comprehensive coverage; key parameter sources are explained.
- **Excellent**: While ensuring 100% factual accuracy, the model integrates and analyzes results across multiple tools; key conclusions are traceable with a complete evidence chain.

When the user question involves travel-related information such as navigation, routing, traffic conditions, arrival time, transportation costs, or weather:

1. The following information types **must come from tools or context**, and must not be estimated based on common knowledge or memory:
   - Precise times (e.g., "estimated arrival at 15:47", "takes 6 minutes");
   - Distances, mileage, congestion length;
   - Prices or fees (taxi fare, ticket prices, airfare, tolls, etc.);
   - Real-time or date-specific weather, temperature, air quality;
   - Specific station names, flight numbers, train numbers, etc.

2. If a tool does not return certain data, but the assistant still provides seemingly "helpful" concrete values (e.g., "a taxi costs about 12-15 Yuan", "today is 22 Celsius and sunny"), this should be considered hallucination, not a bonus.

3. Geographic / location-related hard constraints (applicable to POI, administrative region, nearby, range/radius searches, etc.):
   - Key parameters such as center-point coordinates, radius, administrative region, and city must come from: explicit user input / existing context / tool returns; otherwise they are considered "fabricated parameters".
   - It is not allowed to assert statements like "X is within Y kilometers of Z" or "Nanchang is within 150 km of Yichun" without prior locating or searching and tool evidence.
   - If searching by administrative region but the center-point coordinates are clearly not within that region (or tool returns indicate inconsistency), the model should be judged as failing reasonableness checks and penalized; if the model retries, corrects, or asks follow-up questions, it may receive additional credit.

### 3. Final Answer Description and Presentation Capability
**Key Evaluation Points**:
- Whether the model accurately and directly responds to and completes the user's explicitly stated core request.
- Clarity, logic, and structure of the content.
- Whether information is presented clearly, accurately, and concisely.
- **Credible Expression Based on Tool Results**
  - When tool results are insufficient to support a conclusion, whether the model explicitly states the limitation and asks the user for necessary information or suggests the next tool invocation, rather than forcing an answer.
  - Whether the response avoids presenting guesses or common knowledge as certain facts, and clearly marks uncertainty.

**Rating Standards**:
- **Very Poor**: Confusing or incomprehensible description, failure to complete the user request; or provides unfounded definitive conclusions.
- **Poor**: Description lacks clarity or contains obvious redundancy; only partially completes the user request; lacks explanation of tool insufficiency or errors.
- **Average**: Description is understandable but not concise; interaction is limited; partially completes the user request; insufficient credibility boundary prompt.
- **Good**: Clear and concise description, reasonable interaction; fulfills the user request well; indicates the need for more precise information or next steps.
- **Excellent**: Extremely clear, concise, and well-interactive; fully fulfills the user request; proposes additional interaction or immediately executable next actions; strictly bounded by tool evidence.

## Response Analysis
### Additional Context Information
"""  
{CONTEXT_INFO}  
"""  

### User Question Content
"""  
{QUESTION_CONTENT}  
"""  

### Available Tools
"""  
{INTENDED_TOOL}  
"""  

### Conversation History
"""  
{CONVERSATION_HISTORY}  
"""  

## Output Requirements
- Before each rating, you must first provide detailed reasoning (the reasoning must explicitly cover:
(1) whether tools should be used / were used; 
(2) whether parameters have valid sources; 
(3) whether tool returns support the conclusions; 
(4) whether there were checks, retries, or follow-up questions when results did not meet expectations).
- Strictly output in the following XML format. Ensure tags are properly closed. The reasoning content must be plain text; do not use any HTML tags such as <b> or <br>.

<response>
<reasoning_planning>
<reasoning>
<!-- Evaluate whether the model's tool invocation trajectory is clear, accurate, and well-planned; focus on tool selection, parameter sourcing, and validation/correction when tool returns are abnormal -->
</reasoning>
<rating><!-- Rating: Very Poor, Poor, Average, Good, Excellent --></rating>
</reasoning_planning>
<summarization_extraction>
<reasoning>
<!-- Evaluate whether the model can accurately extract key content from the user question and tool return information; strictly check for fabricated latitude/longitude, center points, distances, administrative regions; strictly check whether conclusions are consistent with tool evidence -->
</reasoning>
<rating><!-- Rating: Very Poor, Poor, Average, Good, Excellent --></rating>
</summarization_extraction>
<presentation>
<reasoning>
<!-- Evaluate whether the model presents relevant content clearly, accurately, and concisely; when evidence is insufficient, whether it clearly states boundaries and guides the user to provide more information or take the next step, rather than forcing a definitive answer -->
</reasoning>
<rating><!-- Rating: Very Poor, Poor, Average, Good, Excellent --></rating>
</presentation>
</response>

## Output Example

<response>
<reasoning_planning>
<reasoning>
When handling the user's request for transportation planning from Northeast China to Sanya and Xishuangbanna, the model exhibited clear logical flaws. First, without querying flights or trains from Harbin to Xishuangbanna via tools, the model directly provided a specific flight number (YU6852), flight duration (2 hours), and ticket price (500-800 RMB) in the final answer...
</reasoning>
<rating>Very Poor</rating>
</reasoning_planning>
<summarization_extraction>
<reasoning>
The model showed serious issues in information extraction and factual accuracy. For transportation to Sanya, the model partially referenced tool-returned data; however, for Xishuangbanna transportation planning, it fabricated a non-existent flight number (YU6852, and this airline code does not exist) without any supporting tool return...
</reasoning>
<rating>Very Poor</rating>
</summarization_extraction>
<presentation>
<reasoning>
Although the answer appeared clear and well-structured in format, its core content-especially the Xishuangbanna section-was built on false information. This kind of "clarity" is therefore misleading. The model packaged guesses and common knowledge as certain facts, and failed to explicitly state the limitations caused by not querying transportation information for Xishuangbanna...
</reasoning>
<rating>Poor</rating>
</presentation>
</response>

\end{tcblisting}
\nopagebreak
\begin{center}
    \captionof{figure}{LLM-Judge Prompt Template for Single-Turn Subtask}
    \label{app:prompt:llm-judge-single-turn}
\end{center}

\begin{tcblisting}{
    enhanced,
    breakable,                % 允许跨页
    listing only,             % 解决 ## 报错的关键：只作为文本列表处理
    colback=black!5!white,            
    colframe=black!75!white,        % 边框颜色（原本内层的浅灰色）
    coltitle=white,  
    boxrule=1pt,            % 边框粗细
    arc=2pt,                  % 圆角
    left=2mm, right=2mm, top=2mm, bottom=2mm,
    % 将原本外层的标题信息集成进来
    title=LLM-Judge Prompt Template for Multi-Turn Subtask,
    fonttitle=\bfseries,
    coltitle=white,
    % 在盒子内容开头插入加粗的提示语
    % before upper={\small\textbf{Prompt Template:}\par\medskip},
    listing options={
        basicstyle=\footnotesize,
        breaklines=true,
        columns=flexible,
        keepspaces=true,
        escapeinside={(*}{*)},
        literate={→}{{$\rightarrow$}}1
    },
}
## Task Description

Conduct a **response quality evaluation** for a tool-based conversation, assessing the model across **four core capability dimensions**.

## Evaluation Objective

Based on the given conversation content, analyze the model's response from the following four dimensions:

1. **Tool Usage and Planning Ability** - Whether the model fully understands the relationship between user needs and the toolset; whether the tool invocation trajectory is clear, reasonable, and accurate; whether the planning of tool calls is correct; and whether tool parameters are filled in appropriately and correctly.
2. **Summarization and Extraction Ability** - After obtaining the user's query and tool function responses, whether the model can selectively extract the most important information (e.g., required function parameters) based on available and historical information, avoiding arbitrary fabrication of facts or data.
3. **Final Answer Description and Presentation Ability** - After completing planning and receiving tool results, whether the final response clearly, accurately, and concisely presents content relevant to the user's needs, and whether it appropriately interacts with or provides feedback to the user (e.g., requesting more precise information or suggesting query-related follow-up questions).
4. **User Interaction and Follow-up Ability** - When information is insufficient, ambiguous, or tool results are abnormal, whether the model can ask necessary and high-value questions with minimal user disruption; whether it prioritizes inference or tool usage to supplement information; and whether follow-up questions stay aligned with the user's original intent rather than drifting toward preference-based or irrelevant inquiries.

## Core Mandatory Constraints
You must treat the following as **primary inspection items throughout all three evaluation dimensions**:  
**whether tool-call parameters are sourced from real information, whether the parameters are filled in reasonably, whether tool returns are correctly used, and whether the final answer is strictly based on tool returns**. 

* **Parameter Authenticity**: Whether key parameters in tool calls (e.g., center coordinates, city/administrative area, keywords, radius, time, stations) originate from the conversation context or tool responses. If parameters are invented by the model without source, this is a serious issue.
* **Parameter Reasonableness**: Even if parameters have a source, evaluate whether they are suitable for the task and user intent. If inappropriate choices (e.g., wrong center point or city) cause failed or misaligned searches, this should be penalized.
* **Process Consistency**: Whether the model gives geographic conclusions (distance, direction, within/outside range, administrative affiliation, etc.) before searching, locating, or confirming information; or directly fabricates coordinates or center points-this results in severe penalties.
* **Result Consistency**: Whether factual information in the final answer (distance, range, place relationships, administrative areas, coordinates, routes, etc.) can be fully supported by tool responses. If the tool did not return such information or does not support the conclusion, this is considered hallucination or inconsistency and should be heavily penalized.
* **Exception Handling**: When tool responses are empty, erroneous, missing fields, or obviously unreasonable (e.g., administrative area does not match center point, coordinates do not match location), whether the model checks, corrects, retries, or follows up. Directly fabricating conclusions in such cases should be heavily penalized.
* **No Tool Usage**: In tasks where tool usage is expected, if the model provides specific conclusions without calling tools, it should be explicitly judged as high-risk hallucination (unless all necessary information is already provided in context).
* **Minimal-Disruption Interaction Principle**: If missing information can be inferred from context or obtained via tools, but the model instead asks the user, this should be penalized. If follow-up questions deviate from the user's original goal or needs, this should also be penalized.

## Evaluation Criteria

### 1. Tool Usage and Planning Ability

**Evaluation Focus**:

* Whether the model understands the alignment between user needs and tool capabilities, and remains user-centered during execution without unnecessary deviation.
* Whether the overall tool usage trajectory is reasonable and clear, including a clear plan (what tools are needed, in what order, where key parameters come from, and how to handle missing information via supplementation, follow-up, or retry).
* Whether all invoked tools are meaningful, avoiding redundant reasoning or calls.
* **Parameter and Validation Planning**:

  * Whether tool parameters are reasonable and consistent with tool definitions and user intent.
  * Whether there is a "locate/search before concluding" step (e.g., obtaining coordinates or center points before range or distance queries).
  * Whether there is validation and correction logic when tool results do not meet expectations (e.g., mismatched administrative areas leading to revised queries or user confirmation).

**Rating Scale**:

* Very Poor: No planning or completely incorrect planning, disconnected from user needs; or unreasonable tool parameters.
* Poor: Clear planning flaws (e.g., unclear or fabricated parameter sources), deviation from goals, lack of exception handling.
* Average: Generally reasonable planning with minor logical gaps (e.g., unclear parameter sources or missing validation steps).
* Good: Clear and reasonable planning with only minor issues; parameter sources explained with basic validation or follow-up.
* Excellent: Complete and precise planning, perfect execution; parameter sources explicit with comprehensive validation and correction mechanisms.

### 2. Summarization and Extraction Ability

**Evaluation Focus**:

* Whether the model extracts core information from user queries and tool responses.
* Whether it focuses on key parameters or points and avoids irrelevant content.
* Whether it aligns with the established plan.
* Whether it strictly references tool responses without fabricating data or facts.
* **Parameter/Field Alignment and Traceability**:

  * Whether key conclusions can be traced to specific tool outputs or fields.
  * Whether fields are misinterpreted or mismatched (e.g., treating administrative areas as center points, or search results as coordinate sources).
  * Whether tool outputs that are empty, erroneous, or unexpected are honestly reflected rather than omitted before drawing conclusions.

**Rating Scale**:

* Very Poor: Hallucinations (fabricated parameters, coordinates, distances, administrative areas, or conclusions), or ignoring tool errors/empty results while giving definitive answers.
* Poor: Missing key information or misinterpreting tool fields; vague parameter sourcing.
* Average: Covers main information but includes unnecessary formatting or loose organization; insufficient explanation of parameter evidence.
* Good: Accurate extraction with no hallucination; comprehensive coverage; clear explanation of parameter sources.
* Excellent: While maintaining 100% factual accuracy, integrates and analyzes results across multiple tools; conclusions are fully traceable with complete evidence chains.

When user questions involve navigation, routing, traffic, arrival times, transportation costs, weather, or travel-related information:

1. The following information types must come from tools or context and must not be estimated from common knowledge or memory:

   * Precise times (e.g., "estimated arrival at 15:47", "takes 6 minutes");
   * Distances, mileage, congestion lengths;
   * Prices or costs (taxi fares, tickets, airfares, tolls, etc.);
   * Real-time or date-specific weather, temperature, air quality;
   * Specific station names, flight numbers, train numbers, etc.

2. If tools do not return certain data, but the assistant provides seemingly "helpful" specific numbers (e.g., "taxi costs about 12-15 Yuan", "today is sunny, 22 Celsius"), this should be treated as hallucination, not a bonus.

3. Geographic/location hard constraints (applicable to POI, administrative areas, nearby/radius searches):

   * Center coordinates, radius, administrative areas, cities must come from explicit user input, existing context, or tool responses; otherwise, they are considered fabricated parameters.
   * It is not allowed to assert "within/outside X km" or similar claims without tool evidence.
   * If searching by administrative area but the center point clearly does not lie within that area (or tool results show inconsistency), failure to validate should be penalized; retries, corrections, or follow-ups should be rewarded.

### 3. Final Answer Description and Presentation Ability

**Evaluation Focus**:

* Whether the response accurately and directly fulfills the user's explicit core request.
* Clarity, logic, and structure of the presentation.
* Conciseness and precision.
* **Credible Expression Based on Tool Results**:

  * When tool results are insufficient to support conclusions, whether limitations are clearly stated and necessary information or next steps are requested, rather than forcing an answer.
  * Whether guesses or common knowledge are clearly marked as uncertain instead of presented as facts.

**Rating Scale**:

* Very Poor: Confusing, incomprehensible, or fails to meet user needs; or provides unsupported definitive conclusions.
* Poor: Unclear or redundant descriptions; partial task completion; insufficient explanation of tool limitations or errors.
* Average: Understandable but not concise; limited interaction; partial fulfillment; insufficient boundary clarification.
* Good: Clear, concise, and well-structured; reasonable interaction; effectively fulfills user needs; suggests next steps.
* Excellent: Extremely clear and concise with effective interaction; fully fulfills user needs; proposes actionable next steps while strictly respecting tool evidence boundaries.

### 4. User Interaction and Follow-up Ability

**Evaluation Focus** (aligned with the first three dimensions but focused on "asking the right questions, asking fewer questions, and asking valuable questions"):

* Necessity: Questions are asked only when key information is missing and cannot be supplemented via tools or context.
* Minimal Disruption: Fewest questions possible; merging multiple gaps into one high-information question; avoiding chained follow-ups.
* **Substitutability Check**: Whether tools could have been used instead of asking the user; unnecessary questions are penalized.
* Question Quality: Specific, actionable, targeting key parameters (location, time, range, object ID), not vague questions like "What do you want?".
* Intent Alignment: Follow-ups stay aligned with the original task; unnecessary preference questions are penalized.
* Tool-Based Follow-ups: When tool results are empty, conflicting, or ambiguous, whether clarifications reference those results (e.g., providing candidates for selection).
* Interaction Closure: After follow-up, whether the next step and how the information will be used is clearly stated; avoiding unused questions.

**Rating Scale**:

* Very Poor: Excessive unnecessary questions; asking users for information obtainable via tools; off-topic questions; no execution closure.
* Poor: Clearly unnecessary or off-intent questions; overly broad; multiple unmerged follow-ups.
* Average: Generally reasonable but slightly excessive or unfocused; insufficient explanation of purpose.
* Good: Questions asked only when necessary; clear and focused; tool-first verification; follow-ups based on tool results.
* Excellent: Mature interaction strategy: tool-first supplementation → only ask irreducible questions; ask once for key information; provide options to reduce user input; clear execution loop with minimal disruption.

## Response Analysis

### Additional Context Information

"""
{CONTEXT_INFO}
"""

### User Question Content

"""
{QUESTION_CONTENT}
"""

### Available Tools

"""
{INTENDED_TOOL}
"""

### Conversation History

"""
{CONVERSATION_HISTORY}
"""

## Output Requirements

* Before each rating, provide detailed reasoning that explicitly covers:
(1) whether tools should have been used and whether they were used;
(2) whether parameters have valid sources;
(3) whether tool results support the conclusions;
(4) whether unexpected results were checked, retried, or followed up;
(5) whether follow-ups were necessary, minimally disruptive, tool-substitutable, and aligned with user intent.
* Strictly output in the following XML format, ensuring proper tag closure. The reasoning content must be plain text; do not use any HTML tags such as <b> or <br>.

<reasoning>
<!-- Evaluate whether the model's tool invocation trajectory is clear, accurate, and well-planned; focus on tool selection, parameter sources, and validation and correction when tool returns are abnormal -->
</reasoning>
<rating><!-- Rating: Very Poor, Poor, Average, Good, Excellent --></rating>
</reasoning_planning>
<summarization_extraction>
<reasoning>
<!-- Evaluate whether the model can accurately extract key content based on the user question and tool return information; strictly check for fabricated latitude/longitude, center points, distances, administrative regions, etc.; strictly check whether conclusions are consistent with tool evidence -->
</reasoning>
<rating><!-- Rating: Very Poor, Poor, Average, Good, Excellent --></rating>
</summarization_extraction>
<presentation>
<reasoning>
<!-- Evaluate whether the model presents relevant content clearly, accurately, and concisely; when evidence is insufficient, whether it clearly states boundaries and guides the user to provide additional information or take the next step, rather than forcing a definitive conclusion -->
</reasoning>
<rating><!-- Rating: Very Poor, Poor, Average, Good, Excellent --></rating>
</presentation>
<user_interaction>
<reasoning>
<!-- Evaluate whether follow-up questions / interaction are necessary and minimally intrusive: information obtainable via tools or context should not be asked again; whether questions are consolidated into high-information-density prompts focusing on key parameters; whether follow-ups align with user intent and are based on tool returns; whether an action loop is formed after follow-up -->
</reasoning>
<rating><!-- Rating: Very Poor, Poor, Average, Good, Excellent --></rating>
</user_interaction>
</response>

## Output Example

<response>
<reasoning_planning>
<reasoning>
When handling the user's request for transportation planning from Northeast China to Sanya and Xishuangbanna, the model exhibited obvious logical flaws. First, without querying flight or train information from Harbin to Xishuangbanna through tools, the model directly provided a specific flight number (YU6852), flight duration (2 hours), and ticket price (500-800 RMB) in the final answer...
</reasoning>
<rating>Very Poor</rating>
</reasoning_planning>
<summarization_extraction>
<reasoning>
The model demonstrated serious issues in information extraction and factual accuracy. For transportation to Sanya, the model did reference part of the tool-returned data; however, for the transportation plan to Xishuangbanna, the model fabricated a fictitious flight number (YU6852, and this airline code does not exist) without any supporting tool return...
</reasoning>
<rating>Very Poor</rating>
</summarization_extraction>
<presentation>
<reasoning>
Although the answer's format appears clear and structured, its core content-especially the Xishuangbanna section-is built on false information. This kind of "clarity" is therefore misleading. The model packaged guesses and common knowledge as certain facts, and failed to explicitly state the limitations caused by not querying transportation information for Xishuangbanna...
</reasoning>
<rating>Poor</rating>
</presentation>
<user_interaction>
<reasoning>
When critical information was missing (no query of Xishuangbanna transportation), the model neither chose to ask the user to clarify the specific departure city (although the context suggested Heilongjiang, the exact city was not confirmed), nor informed the user of the query failure. Instead, it opted for the worst possible approach: fabricating conclusions outright...
</reasoning>
<rating>Very Poor</rating>
</user_interaction>
</response>

\end{tcblisting}
\nopagebreak
\begin{center}
    \captionof{figure}{LLM-Judge Prompt Template for Multi-Turn Subtask}
    \label{app:prompt:llm-judge-multi-turn}
\end{center}

\begin{tcblisting}{
    enhanced,
    breakable,                % 允许跨页
    listing only,             % 解决 ## 报错的关键：只作为文本列表处理
    colback=black!5!white,            % 背景色白色
    colframe=black!75!white,        % 边框颜色（原本内层的浅灰色）
    coltitle=white,  
    boxrule=1pt,            % 边框粗细
    arc=2pt,                  % 圆角
    left=2mm, right=2mm, top=2mm, bottom=2mm,
    % 将原本外层的标题信息集成进来
    title=Prompt Template for Meta-Judge,
    fonttitle=\bfseries,
    coltitle=white,
    % 在盒子内容开头插入加粗的提示语
    % before upper={\small\textbf{Prompt Template:}\par\medskip},
    listing options={
        basicstyle=\footnotesize,
        breaklines=true,
        columns=flexible,
        keepspaces=true,
        escapeinside={(*}{*)},
        literate={→}{{$\rightarrow$}}1
    },
}

## Task Description

Conduct a **Meta-Evaluation** of an evaluation result for a conversation - that is, evaluate the **quality and credibility of the evaluation itself**, rather than re-evaluating the original conversation.

## Evaluation Objective

Based on the original conversation trace and the output of the first-round evaluation (including per-dimension ratings and rationales), comprehensively judge whether that evaluation is reliable in the following aspects:

1. **Scoring Accuracy**:
   Whether the ratings for each dimension genuinely reflect the model's actual performance in the conversation; whether there is any obvious overestimation or underestimation.

2. **Reasoning and Evidence Chain**:
   Whether the evaluation rationale is logically clear, traceable, and grounded in key evidence (conversation turns / tool calls / tool outputs), rather than vague or generic judgments.

3. **Consistency and Strictness of Standards**:
   Whether the evaluation strictly enforces predefined hard constraints and dimension standards (especially parameter authenticity, result consistency, and exception handling), and whether it is overly lenient or overly harsh.

4. **Coverage and Completeness**:
   Whether the evaluation checks all critical risk points and strengths, and whether it omits serious issues or important positive aspects.

5. **Cross-Dimension Coherence**:
   Whether ratings across different dimensions are mutually consistent and supportive; if contradictions exist (e.g., reasoning identifies severe hallucination but the rating is still "good" or "excellent"), they must be explicitly pointed out.

## Original Conversation Information

### Context Information

"""
{CONTEXT_INFO}
"""

### User Question

"""
{QUESTION_CONTENT}
"""

### Tool Definition

"""
{INTENDED_TOOL}
"""

### Conversation History

"""
{CONVERSATION_HISTORY}
"""

## First-Round Evaluation Result

### Evaluation Dimensions and Ratings

{EVALUATION_SCORES}

### Per-Dimension Rationales / Reasoning

{EVALUATION_REASONING}

## Meta-Evaluation Checklist (Must Be Examined Item by Item)

Please evaluate the quality of the first-round evaluation from the following perspectives, and explicitly provide evidence in your reasoning:

### A. Reasonableness of the Scores

* Whether the ratings for each dimension are consistent with the factual conversation record (especially tool call traces, parameter sources, tool outputs, and the final answer).
* Whether there is obvious overestimation or underestimation; if so, explain why the score should be higher or lower, with evidence.
* Whether the dimensions are internally consistent: whether the same issue is heavily penalized in one dimension but ignored in another.

### B. Adequacy of Reasoning (Evidence Chain)

* Whether the rationale is concrete: does it quote key sentences from the dialogue, tool call parameters, or tool output fields to support its judgments.
* Whether it distinguishes between verifiable facts and speculative judgments, avoiding subjective or generic commentary.
* Whether key errors are clearly attributed (e.g., fabricated parameters, failure to use tools, misreading fields, ignoring empty results).
* Whether important issues are omitted.

### C. Strict Enforcement of Standards (Hard Constraints First)

Focus on whether the first-round evaluation properly enforced the following hard constraints in its scoring:

* **Parameter Authenticity**:
  Whether it checked that key parameters came from the context or tool outputs; if parameters lacked provenance, whether this was severely penalized.

* **Result Consistency**:
  Whether the final answer was strictly supported, item by item, by tool outputs; if tools did not return results or did not support the answer, whether this was judged as hallucination.

* **Process Consistency**:
  Whether conclusions (e.g., geographic conclusions) were made without retrieval or localization, and whether the evaluation identified and penalized this.

* **Exception Handling**:
  When tools returned empty results, errors, or contradictions, whether the evaluation required retries, corrections, or follow-up questions, and scored accordingly.

* If the evaluation glossed over or failed to mention these issues, it should be considered non-strict.

### D. Coverage and Completeness

* Whether all critical stages were covered: requirement understanding → planning/tools → parameter provenance → result usage → final presentation → interaction/follow-up (if applicable).
* Whether serious risk points were omitted (e.g., fabricated coordinates/distances/costs/times, ignoring empty tool results).
* Whether the model's strong points were omitted (e.g., reasonable retry strategies, clear evidence-chain explanations, appropriate boundary disclaimers).

### E. Re-evaluation Consistency (Counterfactual Check)

* If you re-evaluated using the same standards, would you give the **same overall level** of judgment?
* If not, identify the most critical points of deviation (no need to redo the full scoring, but explain the reasons for the difference).

## Output Requirements

* Output a single overall rating: **Very Poor / Poor / Average / Good / Excellent** (do not output numbers or scores).
* The reasoning must explicitly state whether the first-round evaluation meets the standards in the four areas of **parameter authenticity, result consistency, exception handling, and evidence-chain citation**, and whether there are cross-dimension contradictions or omissions.
* Strictly output in the following XML format. Ensure correct tag usage. The `reasoning` content must be plain text. Do not use any HTML tags such as `<b>` or `<br>`:

```xml
<meta_evaluation>
<reasoning>
<!-- Provide a detailed analysis of the quality of this evaluation, covering:
1. Scoring accuracy and overestimation/underestimation (with evidence)
2. Sufficiency of reasoning and evidence chain (citing dialogue/tool parameters/tool outputs)
3. Consistency and strictness of standard enforcement (whether hard constraints were applied)
4. Coverage and omitted points
5. Overall conclusion: whether your re-evaluation would yield the same level, and why -->
</reasoning>
<rating><!-- Rating: Very Poor, Poor, Average, Good, Excellent --></rating>
</meta_evaluation>

\end{tcblisting}
\nopagebreak
\begin{center}
    \captionof{figure}{Prompt Template for Meta LLM Judge}
    \label{app:prompt:meta-judge}
\end{center}

\end{document}